\RequirePackage{fix-cm}
\documentclass{article}
\usepackage{icd_preprint,times}
\usepackage{fontspec}
\usepackage{amsmath,amssymb,bm,booktabs,tabularx,array,xcolor,graphicx}
\usepackage{colortbl}
\usepackage{longtable}
\definecolor{OurRowBlue}{RGB}{235,245,253}
\usepackage{tikz}
\usetikzlibrary{positioning}
\usepackage{placeins,needspace,enumitem,wrapfig}
\usepackage[unicode,hidelinks]{hyperref}
\usepackage{url}
\newcommand{\datum}[1]{\csname datum:#1\endcsname}
\expandafter\def\csname datum:tables-unified-graph-tex-c0001\endcsname{\shortstack{\textbf{4}\\\textbf{.846}}}
\expandafter\def\csname datum:tables-unified-graph-tex-c0002\endcsname{\shortstack{\textbf{0}\\\textbf{.982}}}
\expandafter\def\csname datum:tables-unified-graph-tex-c0003\endcsname{\shortstack{\textbf{0}\\\textbf{.862}}}
\expandafter\def\csname datum:tables-unified-graph-tex-c0004\endcsname{\shortstack{6\\.413}}
\expandafter\def\csname datum:tables-unified-graph-tex-c0005\endcsname{\shortstack{7\\\textbf{.793}}}
\expandafter\def\csname datum:tables-unified-graph-tex-c0006\endcsname{\shortstack{\textbf{9}\\\textbf{.242}}}
\expandafter\def\csname datum:tables-unified-graph-tex-c0007\endcsname{\shortstack{\textbf{5}\\\textbf{.587}}}
\expandafter\def\csname datum:tables-unified-graph-tex-c0008\endcsname{\shortstack{\textbf{4}\\\textbf{.276}}}
\expandafter\def\csname datum:tables-unified-graph-tex-c0009\endcsname{\shortstack{6\\.313}}
\expandafter\def\csname datum:tables-unified-graph-tex-c0010\endcsname{\shortstack{3\\.284}}
\expandafter\def\csname datum:tables-unified-graph-tex-c0011\endcsname{\shortstack{3\\.516}}
\expandafter\def\csname datum:tables-unified-graph-tex-c0012\endcsname{\shortstack{8\\\textbf{.475}}}
\expandafter\def\csname datum:tables-unified-graph-tex-c0013\endcsname{\shortstack{8\\.522}}
\expandafter\def\csname datum:tables-unified-graph-tex-c0014\endcsname{\shortstack{11\\.213}}
\expandafter\def\csname datum:tables-unified-graph-tex-c0015\endcsname{\shortstack{7\\.342}}
\expandafter\def\csname datum:tables-unified-graph-tex-c0016\endcsname{\shortstack{11\\.013}}
\expandafter\def\csname datum:tables-unified-graph-tex-c0017\endcsname{\shortstack{6\\.194}}
\expandafter\def\csname datum:tables-unified-graph-tex-c0018\endcsname{\shortstack{2\\.294}}
\expandafter\def\csname datum:tables-unified-graph-tex-c0019\endcsname{\shortstack{5\\.232}}
\expandafter\def\csname datum:tables-unified-graph-tex-c0020\endcsname{\shortstack{\textbf{4}\\.178}}
\expandafter\def\csname datum:tables-unified-graph-tex-c0021\endcsname{\shortstack{\textbf{5}\\.177}}
\expandafter\def\csname datum:tables-unified-graph-tex-c0022\endcsname{\shortstack{17\\.084}}
\expandafter\def\csname datum:tables-unified-graph-tex-c0023\endcsname{\shortstack{8\\.231}}
\expandafter\def\csname datum:tables-unified-graph-tex-c0024\endcsname{\shortstack{16\\.007}}
\expandafter\def\csname datum:tables-unified-graph-tex-c0025\endcsname{8/8}
\expandafter\def\csname datum:tables-unified-graph-tex-c0026\endcsname{7/8}
\expandafter\def\csname datum:tables-unified-graph-tex-c0027\endcsname{8/8}
\expandafter\def\csname datum:tables-unified-graph-tex-c0028\endcsname{7/8}
\expandafter\def\csname datum:tables-unified-graph-tex-c0029\endcsname{8/8}
\expandafter\def\csname datum:tables-unified-graph-tex-c0030\endcsname{8/8}
\expandafter\def\csname datum:tables-unified-graph-tex-c0031\endcsname{6/8}
\expandafter\def\csname datum:tables-unified-graph-tex-c0032\endcsname{7/8}
\expandafter\def\csname datum:tables-calibration-controlled-tex-c0001\endcsname{$22.367\pm3.941$}
\expandafter\def\csname datum:tables-calibration-controlled-tex-c0002\endcsname{$5.463\pm0.946$}
\expandafter\def\csname datum:tables-calibration-controlled-tex-c0003\endcsname{$6.858\pm1.290$}
\expandafter\def\csname datum:tables-calibration-controlled-tex-c0004\endcsname{$1.085\pm0.066$}
\expandafter\def\csname datum:tables-calibration-controlled-tex-c0005\endcsname{$0.874\pm0.046$}
\expandafter\def\csname datum:tables-calibration-controlled-tex-c0006\endcsname{$14.721\pm1.177$}
\expandafter\def\csname datum:tables-calibration-controlled-tex-c0007\endcsname{$\mathbf{4.763}\pm0.142$}
\expandafter\def\csname datum:tables-calibration-controlled-tex-c0008\endcsname{$6.656\pm0.531$}
\expandafter\def\csname datum:tables-calibration-controlled-tex-c0009\endcsname{$\mathbf{1.043}\pm0.065$}
\expandafter\def\csname datum:tables-calibration-controlled-tex-c0010\endcsname{$\mathbf{0.856}\pm0.034$}
\expandafter\def\csname datum:tables-calibration-controlled-tex-c0011\endcsname{$\mathbf{12.864}\pm0.830$}
\expandafter\def\csname datum:tables-calibration-controlled-tex-c0012\endcsname{$4.968\pm0.123$}
\expandafter\def\csname datum:tables-calibration-controlled-tex-c0013\endcsname{$\mathbf{6.515}\pm0.418$}
\expandafter\def\csname datum:tables-calibration-controlled-tex-c0014\endcsname{$1.065\pm0.063$}
\expandafter\def\csname datum:tables-calibration-controlled-tex-c0015\endcsname{$0.878\pm0.036$}
\expandafter\def\csname datum:tables-calibration-controlled-tex-c0016\endcsname{$11.361\pm5.044$}
\expandafter\def\csname datum:tables-calibration-controlled-tex-c0017\endcsname{$9.985\pm1.611$}
\expandafter\def\csname datum:tables-calibration-controlled-tex-c0018\endcsname{$20.449\pm3.725$}
\expandafter\def\csname datum:tables-calibration-controlled-tex-c0019\endcsname{$1.058\pm0.080$}
\expandafter\def\csname datum:tables-calibration-controlled-tex-c0020\endcsname{$0.851\pm0.053$}
\expandafter\def\csname datum:tables-unified-ood-tex-c0001\endcsname{\shortstack{\textbf{8.573}\\\textbf{10.341}\\\textbf{0.829}}}
\expandafter\def\csname datum:tables-unified-ood-tex-c0002\endcsname{\shortstack{\textbf{6.234}\\\textbf{8.291}\\\textbf{0.752}}}
\expandafter\def\csname datum:tables-unified-ood-tex-c0003\endcsname{\shortstack{\textbf{1.771}\\\textbf{3.461}\\\textbf{0.512}}}
\expandafter\def\csname datum:tables-unified-ood-tex-c0004\endcsname{\shortstack{\textbf{2.085}\\\textbf{4.101}\\\textbf{0.508}}}
\expandafter\def\csname datum:tables-unified-ood-tex-c0005\endcsname{\shortstack{\textbf{.729}\\\textbf{1.091}\\\textbf{0.668}}}
\expandafter\def\csname datum:tables-unified-ood-tex-c0006\endcsname{\shortstack{\textbf{5.323}\\\textbf{8.064}\\\textbf{0.660}}}
\expandafter\def\csname datum:tables-unified-ood-tex-c0007\endcsname{\shortstack{\textbf{4.276}\\\textbf{6.786}\\\textbf{0.630}}}
\expandafter\def\csname datum:tables-unified-ood-tex-c0008\endcsname{\shortstack{\textbf{4.817}\\\textbf{5.378}\\\textbf{0.896}}}
\expandafter\def\csname datum:tables-unified-ood-tex-c0009\endcsname{\shortstack{123.799\\807.120\\0.153}}
\expandafter\def\csname datum:tables-unified-ood-tex-c0010\endcsname{\shortstack{228.713\\791.706\\0.289}}
\expandafter\def\csname datum:tables-unified-ood-tex-c0011\endcsname{\shortstack{51.105\\318.545\\0.160}}
\expandafter\def\csname datum:tables-unified-ood-tex-c0012\endcsname{\shortstack{130.996\\852.155\\0.154}}
\expandafter\def\csname datum:tables-unified-ood-tex-c0013\endcsname{\shortstack{75.124\\503.839\\0.149}}
\expandafter\def\csname datum:tables-unified-ood-tex-c0014\endcsname{\shortstack{186.127\\609.486\\0.305}}
\expandafter\def\csname datum:tables-unified-ood-tex-c0015\endcsname{\shortstack{156.435\\987.345\\0.158}}
\expandafter\def\csname datum:tables-unified-ood-tex-c0016\endcsname{\shortstack{293.854\\974.245\\0.302}}
\expandafter\def\csname datum:tables-unified-ood-tex-c0017\endcsname{\shortstack{123.713\\874.392\\0.141}}
\expandafter\def\csname datum:tables-unified-ood-tex-c0018\endcsname{\shortstack{162.410\\515.288\\0.315}}
\expandafter\def\csname datum:tables-unified-ood-tex-c0019\endcsname{\shortstack{58.106\\394.825\\0.147}}
\expandafter\def\csname datum:tables-unified-ood-tex-c0020\endcsname{\shortstack{104.107\\713.652\\0.146}}
\expandafter\def\csname datum:tables-unified-ood-tex-c0021\endcsname{\shortstack{101.258\\699.789\\0.145}}
\expandafter\def\csname datum:tables-unified-ood-tex-c0022\endcsname{\shortstack{234.144\\772.011\\0.303}}
\expandafter\def\csname datum:tables-unified-ood-tex-c0023\endcsname{\shortstack{95.245\\312.261\\0.305}}
\expandafter\def\csname datum:tables-unified-ood-tex-c0024\endcsname{\shortstack{124.215\\332.211\\0.374}}
\expandafter\def\csname datum:tables-unified-ood-tex-c0025\endcsname{\shortstack{34.144\\245.391\\0.139}}
\expandafter\def\csname datum:tables-unified-ood-tex-c0026\endcsname{\shortstack{51.205\\421.363\\0.122}}
\expandafter\def\csname datum:tables-unified-ood-tex-c0027\endcsname{\shortstack{57.011\\393.035\\0.145}}
\expandafter\def\csname datum:tables-unified-ood-tex-c0028\endcsname{\shortstack{102.146\\700.012\\0.146}}
\expandafter\def\csname datum:tables-unified-ood-tex-c0029\endcsname{\shortstack{93.292\\663.595\\0.141}}
\expandafter\def\csname datum:tables-unified-ood-tex-c0030\endcsname{\shortstack{204.114\\402.341\\0.507}}
\expandafter\def\csname datum:tables-unified-ood-tex-c0031\endcsname{\shortstack{98.701\\244.512\\0.404}}
\expandafter\def\csname datum:tables-unified-ood-tex-c0032\endcsname{\shortstack{41.124\\408.946\\0.101}}
\expandafter\def\csname datum:tables-unified-cf-tex-c0001\endcsname{\textbf{1.473}}
\expandafter\def\csname datum:tables-unified-cf-tex-c0002\endcsname{\textbf{.880}}
\expandafter\def\csname datum:tables-unified-cf-tex-c0003\endcsname{\textbf{1.118}}
\expandafter\def\csname datum:tables-unified-cf-tex-c0004\endcsname{\textbf{2.180}}
\expandafter\def\csname datum:tables-unified-cf-tex-c0005\endcsname{\textbf{.938}}
\expandafter\def\csname datum:tables-unified-cf-tex-c0006\endcsname{\textbf{2.384}}
\expandafter\def\csname datum:tables-unified-cf-tex-c0007\endcsname{8.264}
\expandafter\def\csname datum:tables-unified-cf-tex-c0008\endcsname{\textbf{5.261}}
\expandafter\def\csname datum:tables-unified-cf-tex-c0009\endcsname{2.525}
\expandafter\def\csname datum:tables-unified-cf-tex-c0010\endcsname{1.287}
\expandafter\def\csname datum:tables-unified-cf-tex-c0011\endcsname{3.563}
\expandafter\def\csname datum:tables-unified-cf-tex-c0012\endcsname{4.603}
\expandafter\def\csname datum:tables-unified-cf-tex-c0013\endcsname{1.859}
\expandafter\def\csname datum:tables-unified-cf-tex-c0014\endcsname{6.345}
\expandafter\def\csname datum:tables-unified-cf-tex-c0015\endcsname{11.745}
\expandafter\def\csname datum:tables-unified-cf-tex-c0016\endcsname{7.761}
\expandafter\def\csname datum:tables-unified-cf-tex-c0017\endcsname{9.009}
\expandafter\def\csname datum:tables-unified-cf-tex-c0018\endcsname{5.707}
\expandafter\def\csname datum:tables-unified-cf-tex-c0019\endcsname{7.025}
\expandafter\def\csname datum:tables-unified-cf-tex-c0020\endcsname{15.477}
\expandafter\def\csname datum:tables-unified-cf-tex-c0021\endcsname{8.187}
\expandafter\def\csname datum:tables-unified-cf-tex-c0022\endcsname{3.256}
\expandafter\def\csname datum:tables-unified-cf-tex-c0023\endcsname{\textbf{5.234}}
\expandafter\def\csname datum:tables-unified-cf-tex-c0024\endcsname{9.664}
\expandafter\def\csname datum:tables-unified-cf-tex-c0025\endcsname{7.342}
\expandafter\def\csname datum:tables-unified-cf-tex-c0026\endcsname{9.753}
\expandafter\def\csname datum:tables-unified-cf-tex-c0027\endcsname{4.215}
\expandafter\def\csname datum:tables-unified-cf-tex-c0028\endcsname{2.314}
\expandafter\def\csname datum:tables-unified-cf-tex-c0029\endcsname{2.161}
\expandafter\def\csname datum:tables-unified-cf-tex-c0030\endcsname{5.014}
\expandafter\def\csname datum:tables-unified-cf-tex-c0031\endcsname{1.859}
\expandafter\def\csname datum:tables-unified-cf-tex-c0032\endcsname{15.234}
\expandafter\def\csname datum:tables-unified-cf-tex-c0033\endcsname{32.219}
\expandafter\def\csname datum:tables-unified-cf-tex-c0034\endcsname{19.281}
\expandafter\def\csname datum:appendix-reported-protocol-tex-c0001\endcsname{$0.880\pm0.016$}
\expandafter\def\csname datum:appendix-reported-protocol-tex-c0002\endcsname{1.282}
\expandafter\def\csname datum:appendix-reported-protocol-tex-c0003\endcsname{0.478}
\expandafter\def\csname datum:appendix-reported-protocol-tex-c0004\endcsname{$0.924\pm0.026$}
\expandafter\def\csname datum:appendix-reported-protocol-tex-c0005\endcsname{0.997}
\expandafter\def\csname datum:appendix-reported-protocol-tex-c0006\endcsname{0.850}
\expandafter\def\csname datum:appendix-reported-protocol-tex-c0007\endcsname{$1.373\pm0.001$}
\expandafter\def\csname datum:appendix-reported-protocol-tex-c0008\endcsname{1.064}
\expandafter\def\csname datum:appendix-reported-protocol-tex-c0009\endcsname{1.682}
\expandafter\def\csname datum:appendix-reported-protocol-tex-c0010\endcsname{$1.617\pm0.153$}
\expandafter\def\csname datum:appendix-reported-protocol-tex-c0011\endcsname{1.305}
\expandafter\def\csname datum:appendix-reported-protocol-tex-c0012\endcsname{1.929}
\expandafter\def\csname datum:appendix-reported-protocol-tex-c0013\endcsname{$1.287\pm0.039$}
\expandafter\def\csname datum:appendix-reported-protocol-tex-c0014\endcsname{1.051}
\expandafter\def\csname datum:appendix-reported-protocol-tex-c0015\endcsname{1.523}
\expandafter\def\csname datum:appendix-reported-protocol-tex-c0016\endcsname{$1.946\pm0.077$}
\expandafter\def\csname datum:appendix-reported-protocol-tex-c0017\endcsname{1.640}
\expandafter\def\csname datum:appendix-reported-protocol-tex-c0018\endcsname{2.252}
\expandafter\def\csname datum:appendix-reported-protocol-tex-c0019\endcsname{$2.272\pm0.071$}
\expandafter\def\csname datum:appendix-reported-protocol-tex-c0020\endcsname{1.764}
\expandafter\def\csname datum:appendix-reported-protocol-tex-c0021\endcsname{2.780}
\expandafter\def\csname datum:appendix-content-repairs-tex-c0001\endcsname{5}
\expandafter\def\csname datum:appendix-content-repairs-tex-c0002\endcsname{0}
\expandafter\def\csname datum:appendix-content-repairs-tex-c0003\endcsname{6}
\expandafter\def\csname datum:appendix-content-repairs-tex-c0004\endcsname{6}
\expandafter\def\csname datum:appendix-content-repairs-tex-c0005\endcsname{0.0343}
\expandafter\def\csname datum:appendix-content-repairs-tex-c0006\endcsname{6}
\expandafter\def\csname datum:appendix-content-repairs-tex-c0007\endcsname{0}
\expandafter\def\csname datum:appendix-content-repairs-tex-c0008\endcsname{0}
\expandafter\def\csname datum:appendix-content-repairs-tex-c0009\endcsname{0}
\expandafter\def\csname datum:appendix-content-repairs-tex-c0010\endcsname{0.0311}
\expandafter\def\csname datum:appendix-content-repairs-tex-c0011\endcsname{7}
\expandafter\def\csname datum:appendix-content-repairs-tex-c0012\endcsname{0}
\expandafter\def\csname datum:appendix-content-repairs-tex-c0013\endcsname{0}
\expandafter\def\csname datum:appendix-content-repairs-tex-c0014\endcsname{0}
\expandafter\def\csname datum:appendix-content-repairs-tex-c0015\endcsname{0.0306}
\expandafter\def\csname datum:tables-r38-acquisition-tex-c0001\endcsname{5}
\expandafter\def\csname datum:tables-r38-acquisition-tex-c0002\endcsname{1800}
\expandafter\def\csname datum:tables-r38-acquisition-tex-c0003\endcsname{0}
\expandafter\def\csname datum:tables-r38-acquisition-tex-c0004\endcsname{6}
\expandafter\def\csname datum:tables-r38-acquisition-tex-c0005\endcsname{5}
\expandafter\def\csname datum:tables-r38-acquisition-tex-c0006\endcsname{1800}
\expandafter\def\csname datum:tables-r38-acquisition-tex-c0007\endcsname{0}
\expandafter\def\csname datum:tables-r38-acquisition-tex-c0008\endcsname{6}
\expandafter\def\csname datum:tables-r38-acquisition-tex-c0009\endcsname{6}
\expandafter\def\csname datum:tables-r38-acquisition-tex-c0010\endcsname{3600}
\expandafter\def\csname datum:tables-r38-acquisition-tex-c0011\endcsname{0}
\expandafter\def\csname datum:tables-r38-acquisition-tex-c0012\endcsname{6}
\expandafter\def\csname datum:tables-r38-acquisition-tex-c0013\endcsname{6}
\expandafter\def\csname datum:tables-r38-acquisition-tex-c0014\endcsname{1800}
\expandafter\def\csname datum:tables-r38-acquisition-tex-c0015\endcsname{0}
\expandafter\def\csname datum:tables-r38-acquisition-tex-c0016\endcsname{6}
\expandafter\def\csname datum:tables-r38-acquisition-tex-c0017\endcsname{7}
\expandafter\def\csname datum:tables-r38-acquisition-tex-c0018\endcsname{3600}
\expandafter\def\csname datum:tables-r38-acquisition-tex-c0019\endcsname{0}
\expandafter\def\csname datum:tables-r38-acquisition-tex-c0020\endcsname{6}
\expandafter\def\csname datum:tables-r38-acquisition-tex-c0021\endcsname{7}
\expandafter\def\csname datum:tables-r38-acquisition-tex-c0022\endcsname{1800}
\expandafter\def\csname datum:tables-r38-acquisition-tex-c0023\endcsname{0}
\expandafter\def\csname datum:tables-r38-acquisition-tex-c0024\endcsname{6}
\expandafter\def\csname datum:tables-r38-sources-tex-c0001\endcsname{$23.636\pm2.994$}
\expandafter\def\csname datum:tables-r38-sources-tex-c0002\endcsname{$36.850\pm7.400$}
\expandafter\def\csname datum:tables-r38-sources-tex-c0003\endcsname{$39.063\pm6.405$}
\expandafter\def\csname datum:tables-r38-sources-tex-c0004\endcsname{$15.224\pm0.399$}
\expandafter\def\csname datum:tables-r38-sources-tex-c0005\endcsname{$\mathbf{33.258\pm3.868}$}
\expandafter\def\csname datum:tables-r38-sources-tex-c0006\endcsname{$\mathbf{36.365\pm2.890}$}
\expandafter\def\csname datum:tables-r38-sources-tex-c0007\endcsname{$\mathbf{12.998\pm0.233}$}
\expandafter\def\csname datum:tables-r38-sources-tex-c0008\endcsname{$35.144\pm3.376$}
\expandafter\def\csname datum:tables-r38-sources-tex-c0009\endcsname{$38.296\pm2.306$}
\expandafter\def\csname datum:tables-r38-sources-tex-c0010\endcsname{$22.953\pm3.222$}
\expandafter\def\csname datum:tables-r38-sources-tex-c0011\endcsname{$30.711\pm7.441$}
\expandafter\def\csname datum:tables-r38-sources-tex-c0012\endcsname{$29.878\pm5.482$}
\expandafter\def\csname datum:tables-r38-sources-tex-c0013\endcsname{$14.321\pm0.602$}
\expandafter\def\csname datum:tables-r38-sources-tex-c0014\endcsname{$18.321\pm2.813$}
\expandafter\def\csname datum:tables-r38-sources-tex-c0015\endcsname{$18.961\pm1.251$}
\expandafter\def\csname datum:tables-r38-sources-tex-c0016\endcsname{$\mathbf{12.306\pm0.403}$}
\expandafter\def\csname datum:tables-r38-sources-tex-c0017\endcsname{$\mathbf{14.806\pm1.860}$}
\expandafter\def\csname datum:tables-r38-sources-tex-c0018\endcsname{$\mathbf{16.260\pm0.688}$}
\expandafter\def\csname datum:tables-r38-queries-tex-c0001\endcsname{$37.303\pm8.593$}
\expandafter\def\csname datum:tables-r38-queries-tex-c0002\endcsname{$3.366\pm2.889$}
\expandafter\def\csname datum:tables-r38-queries-tex-c0003\endcsname{$2.257\pm0.025$}
\expandafter\def\csname datum:tables-r38-queries-tex-c0004\endcsname{$\mathbf{32.810\pm3.066}$}
\expandafter\def\csname datum:tables-r38-queries-tex-c0005\endcsname{$\mathbf{1.758\pm0.025}$}
\expandafter\def\csname datum:tables-r38-queries-tex-c0006\endcsname{$\mathbf{2.232\pm0.023}$}
\expandafter\def\csname datum:tables-r38-queries-tex-c0007\endcsname{$34.875\pm1.268$}
\expandafter\def\csname datum:tables-r38-queries-tex-c0008\endcsname{$2.045\pm0.176$}
\expandafter\def\csname datum:tables-r38-queries-tex-c0009\endcsname{$2.237\pm0.019$}
\expandafter\def\csname datum:tables-r38-queries-tex-c0010\endcsname{$38.396\pm8.949$}
\expandafter\def\csname datum:tables-r38-queries-tex-c0011\endcsname{$3.983\pm3.330$}
\expandafter\def\csname datum:tables-r38-queries-tex-c0012\endcsname{$1.699\pm0.039$}
\expandafter\def\csname datum:tables-r38-queries-tex-c0013\endcsname{$\mathbf{35.102\pm2.667}$}
\expandafter\def\csname datum:tables-r38-queries-tex-c0014\endcsname{$3.756\pm0.321$}
\expandafter\def\csname datum:tables-r38-queries-tex-c0015\endcsname{$\mathbf{1.688\pm0.043}$}
\expandafter\def\csname datum:tables-r38-queries-tex-c0016\endcsname{$36.703\pm0.869$}
\expandafter\def\csname datum:tables-r38-queries-tex-c0017\endcsname{$\mathbf{3.492\pm0.510}$}
\expandafter\def\csname datum:tables-r38-queries-tex-c0018\endcsname{$1.692\pm0.048$}
\expandafter\def\csname datum:appendix-static-factor-tex-c0001\endcsname{\shortstack{$0.1745$\\{\scriptsize $[0.1650,0.1855]$}}}
\expandafter\def\csname datum:appendix-static-factor-tex-c0002\endcsname{\shortstack{$0.1345$\\{\scriptsize $[0.1251,0.1452]$}}}
\expandafter\def\csname datum:appendix-static-factor-tex-c0003\endcsname{\shortstack{$0.0886$\\{\scriptsize $[0.0771,0.1041]$}}}
\expandafter\def\csname datum:appendix-static-factor-tex-c0004\endcsname{\shortstack{$0.1021$\\{\scriptsize $[0.0995,0.1045]$}}}
\expandafter\def\csname datum:appendix-static-factor-tex-c0005\endcsname{\shortstack{$0.2716$\\{\scriptsize $[0.1750,0.3939]$}}}
\expandafter\def\csname datum:appendix-static-factor-tex-c0006\endcsname{\shortstack{$0.1319$\\{\scriptsize $[0.1235,0.1413]$}}}
\expandafter\def\csname datum:appendix-static-factor-tex-c0007\endcsname{\shortstack{$0.0894$\\{\scriptsize $[0.0781,0.1049]$}}}
\expandafter\def\csname datum:appendix-static-factor-tex-c0008\endcsname{\shortstack{$0.1018$\\{\scriptsize $[0.0998,0.1037]$}}}
\expandafter\def\csname datum:appendix-static-factor-tex-c0009\endcsname{\shortstack{$0.6975$\\{\scriptsize $[0.6841,0.7094]$}}}
\expandafter\def\csname datum:appendix-static-factor-tex-c0010\endcsname{\shortstack{$0.1345$\\{\scriptsize $[0.1252,0.1448]$}}}
\expandafter\def\csname datum:appendix-static-factor-tex-c0011\endcsname{\shortstack{$0.5966$\\{\scriptsize $[0.4478,0.6782]$}}}
\expandafter\def\csname datum:appendix-static-factor-tex-c0012\endcsname{\shortstack{$0.7199$\\{\scriptsize $[0.7024,0.7374]$}}}
\expandafter\def\csname datum:appendix-static-factor-tex-c0013\endcsname{\shortstack{$0.4824$\\{\scriptsize $[0.4183,0.5600]$}}}
\expandafter\def\csname datum:appendix-static-factor-tex-c0014\endcsname{\shortstack{$0.8287$\\{\scriptsize $[0.7532,0.9065]$}}}
\expandafter\def\csname datum:appendix-static-factor-tex-c0015\endcsname{\shortstack{$0.6069$\\{\scriptsize $[0.5283,0.6883]$}}}
\expandafter\def\csname datum:appendix-static-factor-tex-c0016\endcsname{\shortstack{$0.6478$\\{\scriptsize $[0.5312,0.7558]$}}}
\expandafter\def\csname datum:appendix-static-factor-tex-c0017\endcsname{\shortstack{$0.6975$\\{\scriptsize $[0.6841,0.7091]$}}}
\expandafter\def\csname datum:appendix-static-factor-tex-c0018\endcsname{\shortstack{$0.9662$\\{\scriptsize $[0.9486,0.9865]$}}}
\expandafter\def\csname datum:appendix-static-factor-tex-c0019\endcsname{\shortstack{$0.9418$\\{\scriptsize $[0.9290,0.9552]$}}}
\expandafter\def\csname datum:appendix-static-factor-tex-c0020\endcsname{\shortstack{$0.9081$\\{\scriptsize $[0.8879,0.9315]$}}}
\expandafter\def\csname datum:appendix-static-factor-tex-c0021\endcsname{\shortstack{$0.1745$\\{\scriptsize $[0.1650,0.1858]$}}}
\expandafter\def\csname datum:appendix-static-factor-tex-c0022\endcsname{\shortstack{$0.1345$\\{\scriptsize $[0.1253,0.1448]$}}}
\expandafter\def\csname datum:appendix-static-factor-tex-c0023\endcsname{\shortstack{$0.0886$\\{\scriptsize $[0.0770,0.1043]$}}}
\expandafter\def\csname datum:appendix-static-factor-tex-c0024\endcsname{\shortstack{$0.1021$\\{\scriptsize $[0.0995,0.1045]$}}}
\expandafter\def\csname datum:tables-calibration-chain-tex-c0001\endcsname{$23.090\pm 3.183$}
\expandafter\def\csname datum:tables-calibration-chain-tex-c0002\endcsname{0/1/0}
\expandafter\def\csname datum:tables-calibration-chain-tex-c0003\endcsname{$2.1894\pm 0.0183$}
\expandafter\def\csname datum:tables-calibration-chain-tex-c0004\endcsname{$15.036\pm 0.773$}
\expandafter\def\csname datum:tables-calibration-chain-tex-c0005\endcsname{0/1/0}
\expandafter\def\csname datum:tables-calibration-chain-tex-c0006\endcsname{$\mathbf{2.1643}\pm 0.0192$}
\expandafter\def\csname datum:tables-calibration-chain-tex-c0007\endcsname{$\mathbf{12.849}\pm 0.798$}
\expandafter\def\csname datum:tables-calibration-chain-tex-c0008\endcsname{0/1/0}
\expandafter\def\csname datum:tables-calibration-chain-tex-c0009\endcsname{$2.1707\pm 0.0204$}
\expandafter\def\csname datum:tables-current-interface-tex-c0001\endcsname{$22.683\pm4.703$}
\expandafter\def\csname datum:tables-current-interface-tex-c0002\endcsname{$4.596\pm2.928$}
\expandafter\def\csname datum:tables-current-interface-tex-c0003\endcsname{$2.297\pm0.043$}
\expandafter\def\csname datum:tables-current-interface-tex-c0004\endcsname{$14.377\pm1.253$}
\expandafter\def\csname datum:tables-current-interface-tex-c0005\endcsname{$\mathbf{3.784}\pm0.103$}
\expandafter\def\csname datum:tables-current-interface-tex-c0006\endcsname{$2.278\pm0.040$}
\expandafter\def\csname datum:tables-current-interface-tex-c0007\endcsname{$12.190\pm1.201$}
\expandafter\def\csname datum:tables-current-interface-tex-c0008\endcsname{$3.831\pm0.209$}
\expandafter\def\csname datum:tables-current-interface-tex-c0009\endcsname{$2.285\pm0.038$}
\expandafter\def\csname datum:tables-current-interface-tex-c0010\endcsname{$\mathbf{8.075}\pm0.860$}
\expandafter\def\csname datum:tables-current-interface-tex-c0011\endcsname{$13.225\pm2.447$}
\expandafter\def\csname datum:tables-current-interface-tex-c0012\endcsname{$\mathbf{1.052}\pm0.021$}
\expandafter\def\csname datum:tables-current-queries-tex-c0001\endcsname{$35.074\pm4.111$}
\expandafter\def\csname datum:tables-current-queries-tex-c0002\endcsname{$5.748\pm0.818$}
\expandafter\def\csname datum:tables-current-queries-tex-c0003\endcsname{$2.932\pm2.157$}
\expandafter\def\csname datum:tables-current-queries-tex-c0004\endcsname{$2.297\pm0.043$}
\expandafter\def\csname datum:tables-current-queries-tex-c0005\endcsname{$\mathbf{33.871}\pm4.674$}
\expandafter\def\csname datum:tables-current-queries-tex-c0006\endcsname{$\mathbf{5.250}\pm0.405$}
\expandafter\def\csname datum:tables-current-queries-tex-c0007\endcsname{$\mathbf{1.830}\pm0.234$}
\expandafter\def\csname datum:tables-current-queries-tex-c0008\endcsname{$2.278\pm0.040$}
\expandafter\def\csname datum:tables-current-queries-tex-c0009\endcsname{$36.884\pm5.158$}
\expandafter\def\csname datum:tables-current-queries-tex-c0010\endcsname{$5.703\pm0.248$}
\expandafter\def\csname datum:tables-current-queries-tex-c0011\endcsname{$2.226\pm0.404$}
\expandafter\def\csname datum:tables-current-queries-tex-c0012\endcsname{$2.285\pm0.038$}
\expandafter\def\csname datum:tables-current-queries-tex-c0013\endcsname{$43.699\pm6.104$}
\expandafter\def\csname datum:tables-current-queries-tex-c0014\endcsname{$10.194\pm0.856$}
\expandafter\def\csname datum:tables-current-queries-tex-c0015\endcsname{$3.212\pm0.599$}
\expandafter\def\csname datum:tables-current-queries-tex-c0016\endcsname{$\mathbf{1.052}\pm0.021$}
\expandafter\def\csname datum:tables-current-queries-tex-c0017\endcsname{$36.469\pm4.106$}
\expandafter\def\csname datum:tables-current-queries-tex-c0018\endcsname{$7.143\pm1.173$}
\expandafter\def\csname datum:tables-current-queries-tex-c0019\endcsname{$4.596\pm2.928$}
\expandafter\def\csname datum:tables-current-queries-tex-c0020\endcsname{$1.747\pm0.021$}
\expandafter\def\csname datum:tables-current-queries-tex-c0021\endcsname{$\mathbf{35.764}\pm5.030$}
\expandafter\def\csname datum:tables-current-queries-tex-c0022\endcsname{$\mathbf{7.142}\pm0.706$}
\expandafter\def\csname datum:tables-current-queries-tex-c0023\endcsname{$\mathbf{3.784}\pm0.103$}
\expandafter\def\csname datum:tables-current-queries-tex-c0024\endcsname{$1.739\pm0.027$}
\expandafter\def\csname datum:tables-current-queries-tex-c0025\endcsname{$38.431\pm5.409$}
\expandafter\def\csname datum:tables-current-queries-tex-c0026\endcsname{$7.250\pm0.576$}
\expandafter\def\csname datum:tables-current-queries-tex-c0027\endcsname{$3.831\pm0.209$}
\expandafter\def\csname datum:tables-current-queries-tex-c0028\endcsname{$1.744\pm0.028$}
\expandafter\def\csname datum:tables-current-queries-tex-c0029\endcsname{$54.163\pm8.155$}
\expandafter\def\csname datum:tables-current-queries-tex-c0030\endcsname{$20.658\pm2.591$}
\expandafter\def\csname datum:tables-current-queries-tex-c0031\endcsname{$13.225\pm2.447$}
\expandafter\def\csname datum:tables-current-queries-tex-c0032\endcsname{$\mathbf{0.850}\pm0.019$}
\expandafter\def\csname datum:tables-current-semantics-tex-c0001\endcsname{$219.4\pm4.7$}
\expandafter\def\csname datum:tables-current-semantics-tex-c0002\endcsname{$156.6\pm5.1$}
\expandafter\def\csname datum:tables-current-semantics-tex-c0003\endcsname{$88.6\pm2.4$}
\expandafter\def\csname datum:tables-current-semantics-tex-c0004\endcsname{$123.9\pm0.5$}
\expandafter\def\csname datum:tables-current-semantics-tex-c0005\endcsname{$42.4\pm3.1$}
\expandafter\def\csname datum:tables-current-semantics-tex-c0006\endcsname{$209.8\pm4.5$}
\expandafter\def\csname datum:tables-current-semantics-tex-c0007\endcsname{$155.0\pm6.0$}
\expandafter\def\csname datum:tables-current-semantics-tex-c0008\endcsname{$92.8\pm3.1$}
\expandafter\def\csname datum:tables-current-semantics-tex-c0009\endcsname{$122.1\pm1.4$}
\expandafter\def\csname datum:tables-current-semantics-tex-c0010\endcsname{$47.0\pm3.1$}
\expandafter\def\csname datum:tables-current-semantics-tex-c0011\endcsname{$212.0\pm4.0$}
\expandafter\def\csname datum:tables-current-semantics-tex-c0012\endcsname{$157.2\pm6.2$}
\expandafter\def\csname datum:tables-current-semantics-tex-c0013\endcsname{$92.5\pm3.0$}
\expandafter\def\csname datum:tables-current-semantics-tex-c0014\endcsname{$125.5\pm1.6$}
\expandafter\def\csname datum:tables-current-semantics-tex-c0015\endcsname{$46.5\pm2.8$}
\expandafter\def\csname datum:tables-current-semantics-tex-c0016\endcsname{$\mathbf{28.7}\pm2.5$}
\expandafter\def\csname datum:tables-current-semantics-tex-c0017\endcsname{$\mathbf{61.8}\pm4.7$}
\expandafter\def\csname datum:tables-current-semantics-tex-c0018\endcsname{$\mathbf{17.9}\pm0.4$}
\expandafter\def\csname datum:tables-current-semantics-tex-c0019\endcsname{$\mathbf{61.3}\pm2.4$}
\expandafter\def\csname datum:tables-current-semantics-tex-c0020\endcsname{$\mathbf{6.1}\pm0.4$}
\expandafter\def\csname datum:tables-current-semantics-tex-c0021\endcsname{$77.1\pm2.9$}
\expandafter\def\csname datum:tables-current-semantics-tex-c0022\endcsname{$86.5\pm2.1$}
\expandafter\def\csname datum:tables-current-semantics-tex-c0023\endcsname{$174.2\pm5.7$}
\expandafter\def\csname datum:tables-current-semantics-tex-c0024\endcsname{$149.3\pm1.9$}
\expandafter\def\csname datum:tables-current-semantics-tex-c0025\endcsname{$144.0\pm6.5$}
\expandafter\def\csname datum:tables-current-semantics-tex-c0026\endcsname{$70.2\pm3.2$}
\expandafter\def\csname datum:tables-current-semantics-tex-c0027\endcsname{$82.9\pm3.9$}
\expandafter\def\csname datum:tables-current-semantics-tex-c0028\endcsname{$181.2\pm3.6$}
\expandafter\def\csname datum:tables-current-semantics-tex-c0029\endcsname{$146.8\pm3.3$}
\expandafter\def\csname datum:tables-current-semantics-tex-c0030\endcsname{$151.7\pm4.8$}
\expandafter\def\csname datum:tables-current-semantics-tex-c0031\endcsname{$65.1\pm3.9$}
\expandafter\def\csname datum:tables-current-semantics-tex-c0032\endcsname{$81.8\pm3.8$}
\expandafter\def\csname datum:tables-current-semantics-tex-c0033\endcsname{$184.9\pm3.5$}
\expandafter\def\csname datum:tables-current-semantics-tex-c0034\endcsname{$147.8\pm3.3$}
\expandafter\def\csname datum:tables-current-semantics-tex-c0035\endcsname{$156.3\pm4.5$}
\expandafter\def\csname datum:tables-current-semantics-tex-c0036\endcsname{$\mathbf{16.0}\pm1.0$}
\expandafter\def\csname datum:tables-current-semantics-tex-c0037\endcsname{$\mathbf{48.3}\pm2.9$}
\expandafter\def\csname datum:tables-current-semantics-tex-c0038\endcsname{$\mathbf{26.4}\pm1.9$}
\expandafter\def\csname datum:tables-current-semantics-tex-c0039\endcsname{$\mathbf{57.4}\pm1.6$}
\expandafter\def\csname datum:tables-current-semantics-tex-c0040\endcsname{$\mathbf{16.9}\pm2.6$}
\expandafter\def\csname datum:tables-current-per-seed-tex-c0001\endcsname{5}
\expandafter\def\csname datum:tables-current-per-seed-tex-c0002\endcsname{18.898}
\expandafter\def\csname datum:tables-current-per-seed-tex-c0003\endcsname{1.027}
\expandafter\def\csname datum:tables-current-per-seed-tex-c0004\endcsname{1.969}
\expandafter\def\csname datum:tables-current-per-seed-tex-c0005\endcsname{2.260}
\expandafter\def\csname datum:tables-current-per-seed-tex-c0006\endcsname{5}
\expandafter\def\csname datum:tables-current-per-seed-tex-c0007\endcsname{13.975}
\expandafter\def\csname datum:tables-current-per-seed-tex-c0008\endcsname{1.596}
\expandafter\def\csname datum:tables-current-per-seed-tex-c0009\endcsname{3.826}
\expandafter\def\csname datum:tables-current-per-seed-tex-c0010\endcsname{2.239}
\expandafter\def\csname datum:tables-current-per-seed-tex-c0011\endcsname{5}
\expandafter\def\csname datum:tables-current-per-seed-tex-c0012\endcsname{11.982}
\expandafter\def\csname datum:tables-current-per-seed-tex-c0013\endcsname{1.972}
\expandafter\def\csname datum:tables-current-per-seed-tex-c0014\endcsname{3.803}
\expandafter\def\csname datum:tables-current-per-seed-tex-c0015\endcsname{2.247}
\expandafter\def\csname datum:tables-current-per-seed-tex-c0016\endcsname{5}
\expandafter\def\csname datum:tables-current-per-seed-tex-c0017\endcsname{7.996}
\expandafter\def\csname datum:tables-current-per-seed-tex-c0018\endcsname{3.446}
\expandafter\def\csname datum:tables-current-per-seed-tex-c0019\endcsname{15.700}
\expandafter\def\csname datum:tables-current-per-seed-tex-c0020\endcsname{1.031}
\expandafter\def\csname datum:tables-current-per-seed-tex-c0021\endcsname{6}
\expandafter\def\csname datum:tables-current-per-seed-tex-c0022\endcsname{21.204}
\expandafter\def\csname datum:tables-current-per-seed-tex-c0023\endcsname{2.496}
\expandafter\def\csname datum:tables-current-per-seed-tex-c0024\endcsname{4.065}
\expandafter\def\csname datum:tables-current-per-seed-tex-c0025\endcsname{2.287}
\expandafter\def\csname datum:tables-current-per-seed-tex-c0026\endcsname{6}
\expandafter\def\csname datum:tables-current-per-seed-tex-c0027\endcsname{13.374}
\expandafter\def\csname datum:tables-current-per-seed-tex-c0028\endcsname{2.065}
\expandafter\def\csname datum:tables-current-per-seed-tex-c0029\endcsname{3.860}
\expandafter\def\csname datum:tables-current-per-seed-tex-c0030\endcsname{2.276}
\expandafter\def\csname datum:tables-current-per-seed-tex-c0031\endcsname{6}
\expandafter\def\csname datum:tables-current-per-seed-tex-c0032\endcsname{11.106}
\expandafter\def\csname datum:tables-current-per-seed-tex-c0033\endcsname{2.691}
\expandafter\def\csname datum:tables-current-per-seed-tex-c0034\endcsname{4.053}
\expandafter\def\csname datum:tables-current-per-seed-tex-c0035\endcsname{2.284}
\expandafter\def\csname datum:tables-current-per-seed-tex-c0036\endcsname{6}
\expandafter\def\csname datum:tables-current-per-seed-tex-c0037\endcsname{7.257}
\expandafter\def\csname datum:tables-current-per-seed-tex-c0038\endcsname{3.659}
\expandafter\def\csname datum:tables-current-per-seed-tex-c0039\endcsname{13.168}
\expandafter\def\csname datum:tables-current-per-seed-tex-c0040\endcsname{1.053}
\expandafter\def\csname datum:tables-current-per-seed-tex-c0041\endcsname{7}
\expandafter\def\csname datum:tables-current-per-seed-tex-c0042\endcsname{27.948}
\expandafter\def\csname datum:tables-current-per-seed-tex-c0043\endcsname{5.274}
\expandafter\def\csname datum:tables-current-per-seed-tex-c0044\endcsname{7.753}
\expandafter\def\csname datum:tables-current-per-seed-tex-c0045\endcsname{2.344}
\expandafter\def\csname datum:tables-current-per-seed-tex-c0046\endcsname{7}
\expandafter\def\csname datum:tables-current-per-seed-tex-c0047\endcsname{15.781}
\expandafter\def\csname datum:tables-current-per-seed-tex-c0048\endcsname{1.828}
\expandafter\def\csname datum:tables-current-per-seed-tex-c0049\endcsname{3.667}
\expandafter\def\csname datum:tables-current-per-seed-tex-c0050\endcsname{2.319}
\expandafter\def\csname datum:tables-current-per-seed-tex-c0051\endcsname{7}
\expandafter\def\csname datum:tables-current-per-seed-tex-c0052\endcsname{13.481}
\expandafter\def\csname datum:tables-current-per-seed-tex-c0053\endcsname{2.014}
\expandafter\def\csname datum:tables-current-per-seed-tex-c0054\endcsname{3.637}
\expandafter\def\csname datum:tables-current-per-seed-tex-c0055\endcsname{2.323}
\expandafter\def\csname datum:tables-current-per-seed-tex-c0056\endcsname{7}
\expandafter\def\csname datum:tables-current-per-seed-tex-c0057\endcsname{8.972}
\expandafter\def\csname datum:tables-current-per-seed-tex-c0058\endcsname{2.532}
\expandafter\def\csname datum:tables-current-per-seed-tex-c0059\endcsname{10.807}
\expandafter\def\csname datum:tables-current-per-seed-tex-c0060\endcsname{1.073}
\expandafter\def\csname datum:appendix-paired-evidence-tex-c0001\endcsname{$-0.700\pm0.930$(2/3)}
\expandafter\def\csname datum:appendix-paired-evidence-tex-c0002\endcsname{$-0.042\pm0.021$(3/3)}
\expandafter\def\csname datum:appendix-paired-evidence-tex-c0003\endcsname{$-0.202\pm1.583$(2/3)}
\expandafter\def\csname datum:appendix-paired-evidence-tex-c0004\endcsname{$-0.018\pm0.020$(2/3)}
\expandafter\def\csname datum:appendix-paired-evidence-tex-c0005\endcsname{$-0.495\pm1.022$(2/3)}
\expandafter\def\csname datum:appendix-paired-evidence-tex-c0006\endcsname{$-0.020\pm0.027$(2/3)}
\expandafter\def\csname datum:appendix-paired-evidence-tex-c0007\endcsname{$-0.343\pm1.643$(2/3)}
\expandafter\def\csname datum:appendix-paired-evidence-tex-c0008\endcsname{$0.004\pm0.023$(2/3)}
\expandafter\def\csname datum:appendix-paired-evidence-tex-c0009\endcsname{$0.204\pm0.259$(1/3)}
\expandafter\def\csname datum:appendix-paired-evidence-tex-c0010\endcsname{$0.022\pm0.006$(0/3)}
\expandafter\def\csname datum:appendix-paired-evidence-tex-c0011\endcsname{$-0.141\pm0.261$(2/3)}
\expandafter\def\csname datum:appendix-paired-evidence-tex-c0012\endcsname{$0.022\pm0.004$(0/3)}
\expandafter\def\csname datum:tables-gears-paired-tex-c0001\endcsname{$7.517\pm1.327$}
\expandafter\def\csname datum:tables-gears-paired-tex-c0002\endcsname{$1.239\pm0.229$}
\expandafter\def\csname datum:tables-gears-paired-tex-c0003\endcsname{$0.455\pm0.152$}
\expandafter\def\csname datum:tables-gears-paired-tex-c0004\endcsname{$0.923\pm0.011$}
\expandafter\def\csname datum:tables-gears-paired-tex-c0005\endcsname{$7.517\pm1.327$}
\expandafter\def\csname datum:tables-gears-paired-tex-c0006\endcsname{$1.556\pm0.384$}
\expandafter\def\csname datum:tables-gears-paired-tex-c0007\endcsname{$0.845\pm0.632$}
\expandafter\def\csname datum:tables-gears-paired-tex-c0008\endcsname{$1.141\pm0.007$}
\expandafter\def\csname datum:tables-gears-paired-tex-c0009\endcsname{$7.295\pm1.599$}
\expandafter\def\csname datum:tables-gears-paired-tex-c0010\endcsname{$1.347\pm0.287$}
\expandafter\def\csname datum:tables-gears-paired-tex-c0011\endcsname{$0.615\pm0.323$}
\expandafter\def\csname datum:tables-gears-paired-tex-c0012\endcsname{$0.926\pm0.017$}
\expandafter\def\csname datum:tables-gears-paired-tex-c0013\endcsname{$7.295\pm1.599$}
\expandafter\def\csname datum:tables-gears-paired-tex-c0014\endcsname{$2.001\pm0.751$}
\expandafter\def\csname datum:tables-gears-paired-tex-c0015\endcsname{$1.363\pm1.135$}
\expandafter\def\csname datum:tables-gears-paired-tex-c0016\endcsname{$1.144\pm0.016$}
\expandafter\def\csname datum:tables-gears-paired-tex-c0017\endcsname{$7.298\pm1.364$}
\expandafter\def\csname datum:tables-gears-paired-tex-c0018\endcsname{$1.396\pm0.369$}
\expandafter\def\csname datum:tables-gears-paired-tex-c0019\endcsname{$0.742\pm0.471$}
\expandafter\def\csname datum:tables-gears-paired-tex-c0020\endcsname{$0.928\pm0.019$}
\expandafter\def\csname datum:tables-gears-paired-tex-c0021\endcsname{$7.298\pm1.364$}
\expandafter\def\csname datum:tables-gears-paired-tex-c0022\endcsname{$2.324\pm1.153$}
\expandafter\def\csname datum:tables-gears-paired-tex-c0023\endcsname{$1.724\pm1.581$}
\expandafter\def\csname datum:tables-gears-paired-tex-c0024\endcsname{$1.159\pm0.010$}
\expandafter\def\csname datum:tables-decoder-only-tex-c0001\endcsname{$2.259\pm0.038$}
\expandafter\def\csname datum:tables-decoder-only-tex-c0002\endcsname{$1.698\pm0.069$}
\expandafter\def\csname datum:tables-decoder-only-tex-c0003\endcsname{$79.256\pm3.823$}
\expandafter\def\csname datum:tables-decoder-only-tex-c0004\endcsname{$143.379\pm9.280$}
\expandafter\def\csname datum:tables-decoder-only-tex-c0005\endcsname{$1.082\pm0.007$}
\expandafter\def\csname datum:tables-decoder-only-tex-c0006\endcsname{$0.854\pm0.012$}
\expandafter\def\csname datum:tables-decoder-only-tex-c0007\endcsname{$16.212\pm1.226$}
\expandafter\def\csname datum:tables-decoder-only-tex-c0008\endcsname{$15.145\pm1.766$}
\expandafter\def\csname datum:tables-decoder-only-tex-c0009\endcsname{$2.241\pm0.044$}
\expandafter\def\csname datum:tables-decoder-only-tex-c0010\endcsname{$1.690\pm0.076$}
\expandafter\def\csname datum:tables-decoder-only-tex-c0011\endcsname{$72.223\pm4.315$}
\expandafter\def\csname datum:tables-decoder-only-tex-c0012\endcsname{$150.861\pm8.913$}
\expandafter\def\csname datum:tables-decoder-only-tex-c0013\endcsname{$1.037\pm0.021$}
\expandafter\def\csname datum:tables-decoder-only-tex-c0014\endcsname{$0.836\pm0.029$}
\expandafter\def\csname datum:tables-decoder-only-tex-c0015\endcsname{$15.810\pm1.201$}
\expandafter\def\csname datum:tables-decoder-only-tex-c0016\endcsname{$16.719\pm2.287$}
\expandafter\def\csname datum:tables-decoder-only-tex-c0017\endcsname{$2.246\pm0.044$}
\expandafter\def\csname datum:tables-decoder-only-tex-c0018\endcsname{$1.695\pm0.076$}
\expandafter\def\csname datum:tables-decoder-only-tex-c0019\endcsname{$66.577\pm3.543$}
\expandafter\def\csname datum:tables-decoder-only-tex-c0020\endcsname{$155.635\pm8.886$}
\expandafter\def\csname datum:tables-decoder-only-tex-c0021\endcsname{$1.058\pm0.019$}
\expandafter\def\csname datum:tables-decoder-only-tex-c0022\endcsname{$0.859\pm0.028$}
\expandafter\def\csname datum:tables-decoder-only-tex-c0023\endcsname{$15.394\pm1.384$}
\expandafter\def\csname datum:tables-decoder-only-tex-c0024\endcsname{$18.058\pm2.758$}
\expandafter\def\csname datum:tables-decoder-only-tex-c0025\endcsname{$1.047\pm0.034$}
\expandafter\def\csname datum:tables-decoder-only-tex-c0026\endcsname{$0.828\pm0.020$}
\expandafter\def\csname datum:tables-decoder-only-tex-c0027\endcsname{$16.001\pm1.145$}
\expandafter\def\csname datum:tables-decoder-only-tex-c0028\endcsname{$16.625\pm2.312$}
\expandafter\def\csname datum:tables-decoder-only-tex-c0029\endcsname{$-1.177\pm0.039$}
\expandafter\def\csname datum:tables-decoder-only-tex-c0030\endcsname{3/3}
\expandafter\def\csname datum:tables-decoder-only-tex-c0031\endcsname{$-0.844\pm0.058$}
\expandafter\def\csname datum:tables-decoder-only-tex-c0032\endcsname{3/3}
\expandafter\def\csname datum:tables-decoder-only-tex-c0033\endcsname{$-1.204\pm0.028$}
\expandafter\def\csname datum:tables-decoder-only-tex-c0034\endcsname{3/3}
\expandafter\def\csname datum:tables-decoder-only-tex-c0035\endcsname{$-0.854\pm0.047$}
\expandafter\def\csname datum:tables-decoder-only-tex-c0036\endcsname{3/3}
\expandafter\def\csname datum:tables-decoder-only-tex-c0037\endcsname{$-1.188\pm0.027$}
\expandafter\def\csname datum:tables-decoder-only-tex-c0038\endcsname{3/3}
\expandafter\def\csname datum:tables-decoder-only-tex-c0039\endcsname{$-0.836\pm0.047$}
\expandafter\def\csname datum:tables-decoder-only-tex-c0040\endcsname{3/3}
\expandafter\def\csname datum:tables-controlled-ablation-tex-c0001\endcsname{3}
\expandafter\def\csname datum:tables-controlled-ablation-tex-c0002\endcsname{$23.337\pm5.417$}
\expandafter\def\csname datum:tables-controlled-ablation-tex-c0003\endcsname{$15.333\pm2.244$}
\expandafter\def\csname datum:tables-controlled-ablation-tex-c0004\endcsname{3}
\expandafter\def\csname datum:tables-controlled-ablation-tex-c0005\endcsname{$114.354\pm20.126$}
\expandafter\def\csname datum:tables-controlled-ablation-tex-c0006\endcsname{$15.421\pm2.135$}
\expandafter\def\csname datum:tables-controlled-ablation-tex-c0007\endcsname{3}
\expandafter\def\csname datum:tables-controlled-ablation-tex-c0008\endcsname{$61.438\pm15.002$}
\expandafter\def\csname datum:tables-controlled-ablation-tex-c0009\endcsname{$22.149\pm3.796$}
\expandafter\def\csname datum:tables-controlled-ablation-tex-c0010\endcsname{2/3}
\expandafter\def\csname datum:tables-controlled-ablation-tex-c0011\endcsname{$2.2845\pm.0321$}
\expandafter\def\csname datum:tables-controlled-ablation-tex-c0012\endcsname{$\mathbf{2.2604\pm.0325}$}
\expandafter\def\csname datum:tables-controlled-ablation-tex-c0013\endcsname{2/3}
\expandafter\def\csname datum:tables-controlled-ablation-tex-c0014\endcsname{$2.3372\pm.0430$}
\expandafter\def\csname datum:tables-controlled-ablation-tex-c0015\endcsname{$\mathbf{2.2656\pm.0317}$}
\expandafter\def\csname datum:tables-controlled-ablation-tex-c0016\endcsname{1/3}
\expandafter\def\csname datum:tables-controlled-ablation-tex-c0017\endcsname{2.2671}
\expandafter\def\csname datum:tables-controlled-ablation-tex-c0018\endcsname{\textbf{2.2156}}
\expandafter\def\csname datum:tables-controlled-ablation-tex-c0019\endcsname{4496}
\expandafter\def\csname datum:tables-controlled-ablation-tex-c0020\endcsname{$11696^{\dagger}$}
\expandafter\def\csname datum:tables-controlled-ablation-tex-c0021\endcsname{4496}
\expandafter\def\csname datum:tables-controlled-ablation-tex-c0022\endcsname{2/3}
\expandafter\def\csname datum:tables-controlled-ablation-tex-c0023\endcsname{9896}
\expandafter\def\csname datum:tables-controlled-ablation-tex-c0024\endcsname{6296}
\expandafter\def\csname datum:tables-controlled-ablation-tex-c0025\endcsname{2696}
\expandafter\def\csname datum:tables-controlled-ablation-tex-c0026\endcsname{3/3}
\expandafter\def\csname datum:tables-controlled-ablation-tex-c0027\endcsname{2696}
\expandafter\def\csname datum:tables-controlled-ablation-tex-c0028\endcsname{2696}
\expandafter\def\csname datum:tables-controlled-ablation-tex-c0029\endcsname{2696}
\expandafter\def\csname datum:tables-controlled-ablation-tex-c0030\endcsname{3/3}
\expandafter\def\csname datum:tables-module-ablation-tex-c0001\endcsname{$13.327\pm 0.648$}
\expandafter\def\csname datum:tables-module-ablation-tex-c0002\endcsname{$0.0000\pm 0.0000$}
\expandafter\def\csname datum:tables-module-ablation-tex-c0003\endcsname{$\mathbf{13.320}\pm 0.647$}
\expandafter\def\csname datum:tables-module-ablation-tex-c0004\endcsname{$-0.0071\pm 0.0016$}
\expandafter\def\csname datum:tables-module-ablation-tex-c0005\endcsname{$15.245\pm 1.377$}
\expandafter\def\csname datum:tables-module-ablation-tex-c0006\endcsname{$1.9174\pm 1.3410$}
\expandafter\def\csname datum:tables-module-ablation-tex-c0007\endcsname{$\mathbf{0.333}\pm 0.577$}
\expandafter\def\csname datum:tables-module-ablation-tex-c0008\endcsname{2/3}
\expandafter\def\csname datum:tables-module-ablation-tex-c0009\endcsname{$4.333\pm 0.577$}
\expandafter\def\csname datum:tables-module-ablation-tex-c0010\endcsname{0/3}
\expandafter\def\csname datum:tables-module-ablation-tex-c0011\endcsname{$2.3077\pm 0.0293$}
\expandafter\def\csname datum:tables-module-ablation-tex-c0012\endcsname{$\mathbf{23.9778}\pm 0.7817$}
\expandafter\def\csname datum:tables-module-ablation-tex-c0013\endcsname{$2.7120\pm 0.0189$}
\expandafter\def\csname datum:tables-module-ablation-tex-c0014\endcsname{$280.1258\pm 0.5836$}
\expandafter\def\csname datum:tables-module-ablation-tex-c0015\endcsname{$\mathbf{2.3067}\pm 0.0333$}
\expandafter\def\csname datum:tables-module-ablation-tex-c0016\endcsname{$24.8983\pm 0.3083$}
\expandafter\def\csname datum:tables-module-ablation-tex-c0017\endcsname{$2.3091\pm 0.0292$}
\expandafter\def\csname datum:tables-module-ablation-tex-c0018\endcsname{$23.9997\pm 0.8275$}
\expandafter\def\csname datum:appendix-ablation-complete-tex-c0001\endcsname{$25.232\pm3.391$}
\expandafter\def\csname datum:appendix-ablation-complete-tex-c0002\endcsname{$2.1892\pm0.1470$}
\expandafter\def\csname datum:appendix-ablation-complete-tex-c0003\endcsname{0/1/0}
\expandafter\def\csname datum:appendix-ablation-complete-tex-c0004\endcsname{$16.278\pm1.553$}
\expandafter\def\csname datum:appendix-ablation-complete-tex-c0005\endcsname{$\mathbf{2.1651}\pm0.1441$}
\expandafter\def\csname datum:appendix-ablation-complete-tex-c0006\endcsname{0/1/0}
\expandafter\def\csname datum:appendix-ablation-complete-tex-c0007\endcsname{$16.278\pm1.553$}
\expandafter\def\csname datum:appendix-ablation-complete-tex-c0008\endcsname{$\mathbf{2.1651}\pm0.1441$}
\expandafter\def\csname datum:appendix-ablation-complete-tex-c0009\endcsname{0/1/0}
\expandafter\def\csname datum:appendix-ablation-complete-tex-c0010\endcsname{$\mathbf{14.146}\pm0.850$}
\expandafter\def\csname datum:appendix-ablation-complete-tex-c0011\endcsname{$2.1726\pm0.1428$}
\expandafter\def\csname datum:appendix-ablation-complete-tex-c0012\endcsname{0/1/0}
\expandafter\def\csname datum:appendix-ablation-complete-tex-c0013\endcsname{5}
\expandafter\def\csname datum:appendix-ablation-complete-tex-c0014\endcsname{21.971}
\expandafter\def\csname datum:appendix-ablation-complete-tex-c0015\endcsname{2.08521}
\expandafter\def\csname datum:appendix-ablation-complete-tex-c0016\endcsname{5}
\expandafter\def\csname datum:appendix-ablation-complete-tex-c0017\endcsname{16.069}
\expandafter\def\csname datum:appendix-ablation-complete-tex-c0018\endcsname{2.06321}
\expandafter\def\csname datum:appendix-ablation-complete-tex-c0019\endcsname{5}
\expandafter\def\csname datum:appendix-ablation-complete-tex-c0020\endcsname{16.069}
\expandafter\def\csname datum:appendix-ablation-complete-tex-c0021\endcsname{2.06321}
\expandafter\def\csname datum:appendix-ablation-complete-tex-c0022\endcsname{5}
\expandafter\def\csname datum:appendix-ablation-complete-tex-c0023\endcsname{13.869}
\expandafter\def\csname datum:appendix-ablation-complete-tex-c0024\endcsname{2.07161}
\expandafter\def\csname datum:appendix-ablation-complete-tex-c0025\endcsname{6}
\expandafter\def\csname datum:appendix-ablation-complete-tex-c0026\endcsname{28.740}
\expandafter\def\csname datum:appendix-ablation-complete-tex-c0027\endcsname{6}
\expandafter\def\csname datum:appendix-ablation-complete-tex-c0028\endcsname{17.925}
\expandafter\def\csname datum:appendix-ablation-complete-tex-c0029\endcsname{6}
\expandafter\def\csname datum:appendix-ablation-complete-tex-c0030\endcsname{17.925}
\expandafter\def\csname datum:appendix-ablation-complete-tex-c0031\endcsname{6}
\expandafter\def\csname datum:appendix-ablation-complete-tex-c0032\endcsname{15.100}
\expandafter\def\csname datum:appendix-ablation-complete-tex-c0033\endcsname{7}
\expandafter\def\csname datum:appendix-ablation-complete-tex-c0034\endcsname{24.984}
\expandafter\def\csname datum:appendix-ablation-complete-tex-c0035\endcsname{2.29315}
\expandafter\def\csname datum:appendix-ablation-complete-tex-c0036\endcsname{7}
\expandafter\def\csname datum:appendix-ablation-complete-tex-c0037\endcsname{14.840}
\expandafter\def\csname datum:appendix-ablation-complete-tex-c0038\endcsname{2.26705}
\expandafter\def\csname datum:appendix-ablation-complete-tex-c0039\endcsname{7}
\expandafter\def\csname datum:appendix-ablation-complete-tex-c0040\endcsname{14.840}
\expandafter\def\csname datum:appendix-ablation-complete-tex-c0041\endcsname{2.26705}
\expandafter\def\csname datum:appendix-ablation-complete-tex-c0042\endcsname{7}
\expandafter\def\csname datum:appendix-ablation-complete-tex-c0043\endcsname{13.468}
\expandafter\def\csname datum:appendix-ablation-complete-tex-c0044\endcsname{2.27358}
\expandafter\def\csname datum:appendix-ablation-complete-tex-c0045\endcsname{2/3}
\expandafter\def\csname datum:appendix-ablation-complete-tex-c0046\endcsname{$2696/11696^*/4496$}
\expandafter\def\csname datum:appendix-ablation-complete-tex-c0047\endcsname{0/1/0}
\expandafter\def\csname datum:appendix-ablation-complete-tex-c0048\endcsname{3/3}
\expandafter\def\csname datum:appendix-ablation-complete-tex-c0049\endcsname{$6296/6296/2696$}
\expandafter\def\csname datum:appendix-ablation-complete-tex-c0050\endcsname{0/0/0}
\expandafter\def\csname datum:appendix-ablation-complete-tex-c0051\endcsname{3/3}
\expandafter\def\csname datum:appendix-ablation-complete-tex-c0052\endcsname{$2696/2696/2696$}
\expandafter\def\csname datum:appendix-ablation-complete-tex-c0053\endcsname{0/0/0}
\expandafter\def\csname datum:appendix-ablation-complete-tex-c0054\endcsname{2/3}
\expandafter\def\csname datum:appendix-ablation-complete-tex-c0055\endcsname{$1800/11696^*/3600$}
\expandafter\def\csname datum:appendix-ablation-complete-tex-c0056\endcsname{0/1/0}
\expandafter\def\csname datum:appendix-unified-evidence-tex-c0001\endcsname{$1.118\pm.134$}
\expandafter\def\csname datum:appendix-unified-evidence-tex-c0002\endcsname{2.180}
\expandafter\def\csname datum:appendix-unified-evidence-tex-c0003\endcsname{$.938\pm.021$}
\expandafter\def\csname datum:appendix-unified-evidence-tex-c0004\endcsname{$2.161\pm.078$}
\expandafter\def\csname datum:appendix-unified-evidence-tex-c0005\endcsname{5.014}
\expandafter\def\csname datum:appendix-unified-evidence-tex-c0006\endcsname{$1.859\pm.093$}
\expandafter\def\csname datum:appendix-unified-evidence-tex-c0007\endcsname{$1.771\pm.433$}
\expandafter\def\csname datum:appendix-unified-evidence-tex-c0008\endcsname{2.085}
\expandafter\def\csname datum:appendix-unified-evidence-tex-c0009\endcsname{$.729\pm.228$}
\expandafter\def\csname datum:appendix-unified-evidence-tex-c0010\endcsname{$57.011\pm2.039$}
\expandafter\def\csname datum:appendix-unified-evidence-tex-c0011\endcsname{102.146}
\expandafter\def\csname datum:appendix-unified-evidence-tex-c0012\endcsname{$93.292\pm.179$}
\expandafter\def\csname datum:appendix-unified-evidence-tex-c0013\endcsname{$3.461\pm1.367$}
\expandafter\def\csname datum:appendix-unified-evidence-tex-c0014\endcsname{4.101}
\expandafter\def\csname datum:appendix-unified-evidence-tex-c0015\endcsname{$1.091\pm.341$}
\expandafter\def\csname datum:appendix-unified-evidence-tex-c0016\endcsname{$393.035\pm10.084$}
\expandafter\def\csname datum:appendix-unified-evidence-tex-c0017\endcsname{700.012}
\expandafter\def\csname datum:appendix-unified-evidence-tex-c0018\endcsname{$663.595\pm5.185$}
\expandafter\def\csname datum:appendix-unified-evidence-tex-c0019\endcsname{$2.082\pm.073$}
\expandafter\def\csname datum:appendix-unified-evidence-tex-c0020\endcsname{$2.116\pm.068$}
\expandafter\def\csname datum:appendix-unified-evidence-tex-c0021\endcsname{$19.382\pm11.244$}
\expandafter\def\csname datum:appendix-unified-evidence-tex-c0022\endcsname{$16.208\pm2.870$}
\expandafter\def\csname datum:appendix-unified-evidence-tex-c0023\endcsname{$32.297\pm40.164$}
\expandafter\def\csname datum:appendix-unified-evidence-tex-c0024\endcsname{$18.480\pm2.970$}
\expandafter\def\csname datum:appendix-unified-evidence-tex-c0025\endcsname{$.370\pm.025$}
\expandafter\def\csname datum:appendix-unified-evidence-tex-c0026\endcsname{$.399\pm.046$}
\expandafter\def\csname datum:appendix-unified-evidence-tex-c0027\endcsname{$.314\pm.065$}
\expandafter\def\csname datum:appendix-unified-evidence-tex-c0028\endcsname{$.331\pm.050$}
\expandafter\def\csname datum:tables-baseline-health-tex-c0001\endcsname{[0.849, 0.984]}
\expandafter\def\csname datum:tables-baseline-health-tex-c0002\endcsname{86/100}
\expandafter\def\csname datum:tables-baseline-health-tex-c0003\endcsname{[0.220, 0.797]}
\expandafter\def\csname datum:tables-baseline-health-tex-c0004\endcsname{120/120}
\expandafter\def\csname datum:tables-baseline-health-tex-c0005\endcsname{[0.919, 0.986]}
\expandafter\def\csname datum:tables-baseline-health-tex-c0006\endcsname{86/100}
\expandafter\def\csname datum:tables-baseline-health-tex-c0007\endcsname{[-0.005, 0.044]}
\expandafter\def\csname datum:tables-baseline-health-tex-c0008\endcsname{120/120}
\expandafter\def\csname datum:tables-baseline-health-tex-c0009\endcsname{[0.920, 0.995]}
\expandafter\def\csname datum:tables-baseline-health-tex-c0010\endcsname{98/100}
\expandafter\def\csname datum:tables-baseline-health-tex-c0011\endcsname{[0.118, 0.804]}
\expandafter\def\csname datum:tables-baseline-health-tex-c0012\endcsname{120/120}
\expandafter\def\csname datum:tables-baseline-health-tex-c0013\endcsname{[0.902, 0.982]}
\expandafter\def\csname datum:tables-baseline-health-tex-c0014\endcsname{93/100}
\expandafter\def\csname datum:tables-baseline-health-tex-c0015\endcsname{[-0.003, 0.826]}
\expandafter\def\csname datum:appendix-run-index-tex-c0001\endcsname{0.930}
\expandafter\def\csname datum:tables-baseline-health-tex-c0016\endcsname{120/120}

\definecolor{ProjectLinkBlue}{RGB}{40,79,112}
\newcommand{\ProjectWebsiteIcon}{%
  \tikz[baseline=-0.55ex,x=1ex,y=1ex,line width=0.45pt,draw=ProjectLinkBlue]{%
    \draw (0,0) circle (0.68);
    \draw (0,0) ellipse (0.29 and 0.68);
    \draw (-0.68,0)--(0.68,0);
    \draw (-0.58,0.34) .. controls (-0.2,0.23) and (0.2,0.23) .. (0.58,0.34);
    \draw (-0.58,-0.34) .. controls (-0.2,-0.23) and (0.2,-0.23) .. (0.58,-0.34);
  }}
\newcommand{\ProjectWebsiteLink}{%
  \begin{center}\small
  \href{https://dzyy123.github.io/intervene-calibrate-discover/}{%
    \color{ProjectLinkBlue}\ProjectWebsiteIcon\enspace
    \textsf{Project website}\enspace
    \texttt{dzyy123.github.io/intervene-calibrate-discover/}}
  \end{center}}

\makeatletter
\newenvironment{inlinetable}{\par\smallskip\noindent\begin{minipage}{\linewidth}\def\@captype{table}}{\end{minipage}\par\smallskip}
\makeatother
\newcolumntype{L}[1]{>{\raggedright\arraybackslash}p{#1}}
\newcolumntype{Y}{>{\centering\arraybackslash}X}

\newcommand{\E}{\mathbb E}
\DeclareMathOperator{\Pa}{Pa}
\DeclareMathOperator{\De}{De}
\DeclareMathOperator{\doop}{do}
\DeclareMathOperator{\Corr}{Corr}

\renewcommand{\refname}{References}
\title{Intervene, Calibrate, Discover:\\Learning Causal Structure from Pixels}
\author{\parbox{\dimexpr\textwidth-2\tabcolsep\relax}{\centering\normalfont
  Ziyi Ding\textsuperscript{1,5,*}\quad
  Xiaodi Huang\textsuperscript{2,*}\quad
  Xianxin Lai\textsuperscript{3,5}\quad
  Haijier Chen\textsuperscript{4,5}\\[0.5ex]
  Zhang Zhang\textsuperscript{2}\quad
  Xiao-Ping Zhang\textsuperscript{1,\ensuremath{\dagger}}\quad
  Jiayu Chen\textsuperscript{3,5,\ensuremath{\dagger}}\\[1.1ex]
  {\small
  \textsuperscript{1}Tsinghua University\\
  \textsuperscript{2}Institute of Automation, Chinese Academy of Sciences\\
  \textsuperscript{3}The University of Hong Kong\qquad
  \textsuperscript{4}Wuhan University\\
  \textsuperscript{5}INFIFORCE Intelligent Technology, China}\\[0.8ex]
  {\footnotesize\textsuperscript{*}Equal contribution.\quad
  \textsuperscript{\ensuremath{\dagger}}Corresponding authors.\\
  \href{mailto:xpzhang@ieee.org}{\texttt{xpzhang@ieee.org}}\quad
  \href{mailto:jiayuc@hku.hk}{\texttt{jiayuc@hku.hk}}}
}}
\hypersetup{
  pdftitle={Intervene, Calibrate, Discover: Learning Causal Structure from Pixels},
  pdfauthor={Ziyi Ding; Xiaodi Huang; Xianxin Lai; Haijier Chen; Zhang Zhang; Xiao-Ping Zhang; Jiayu Chen},
  pdfsubject={Preprint}
}
\newcommand{\PreprintHeader}{%
  \fancyhead{}\fancyhead[L]{Preprint}\renewcommand{\headrulewidth}{0pt}}

\begin{document}
\maketitle
\PreprintHeader
\ProjectWebsiteLink
\begin{abstract}
Learning high-level causal variables from low-level observations is a central problem for artificial intelligence~\citep{scholkopf2021toward}. With known node semantics but an unknown graph, visual readout errors can misalign model settings with environment intervention values. We address this mismatch by pairing image readouts with verified values attained through fixed round-robin interventions. With encoder and decoder frozen, a dimension-wise invertible calibration map puts visual responses and target settings in shared numerical units. After freezing this map, the agent selects interventions to resolve graph directions and fits local mechanisms to predict consequences; inverse calibration returns counterfactual states to the decoder's original coordinates. Across Domino, Pick-and-Place, Balance, Gears, CAUSAL3D, and CausalWorld, we evaluate graph discovery, intervention consequences, and counterfactual images. Controlled domino experiments show that interventions resolve orientation ambiguity left by equal additional observations. In a matched-data domino control, calibration reduces the mean error of predicted changes in other variables within the same factual context, relative to an uncalibrated model.

\end{abstract}
\section{Introduction}\label{sec:intro}
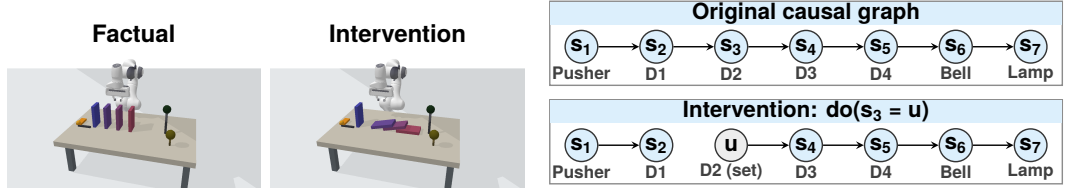
\begin{figure}[htbp]
\centering
\newfontfamily\SceneLabelFont{texgyreheros-bold.otf}
\begin{minipage}{.49\linewidth}
\centering
\begin{minipage}{.49\linewidth}\centering
{\SceneLabelFont\footnotesize Factual}\par
\includegraphics[width=\linewidth]{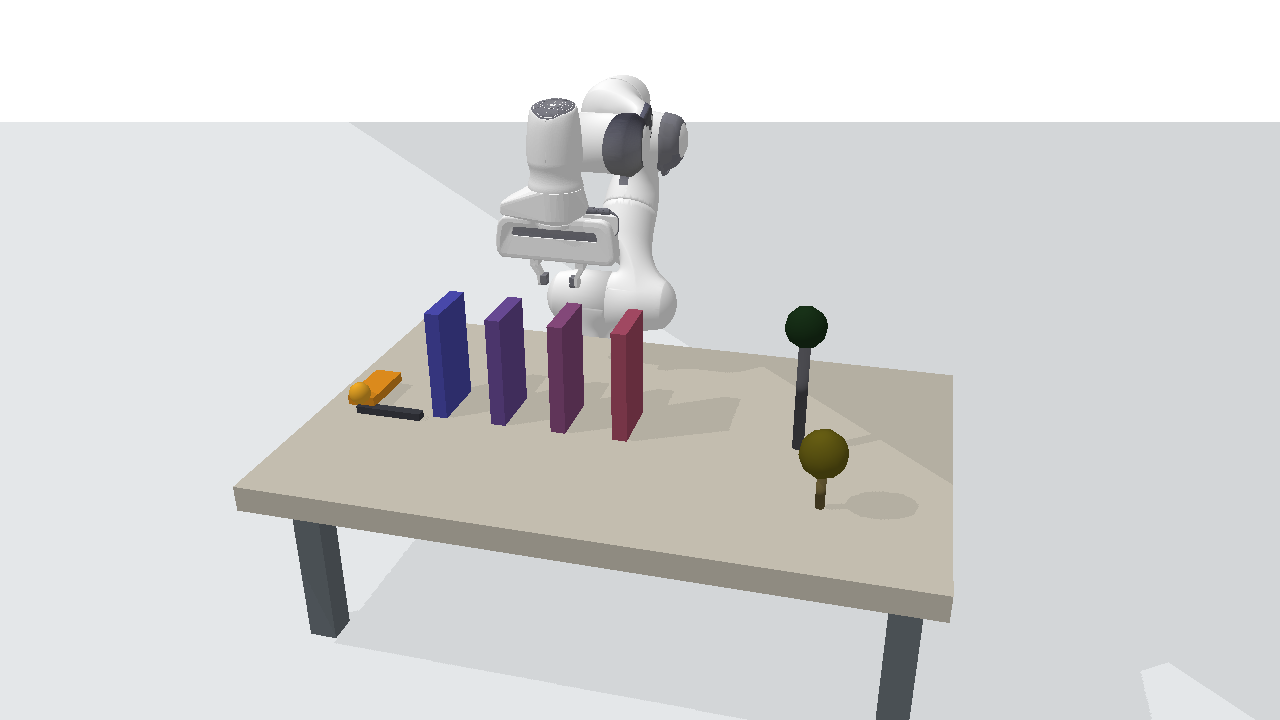}
\end{minipage}\hfill
\begin{minipage}{.49\linewidth}\centering
{\SceneLabelFont\footnotesize Intervention}\par
\includegraphics[width=\linewidth]{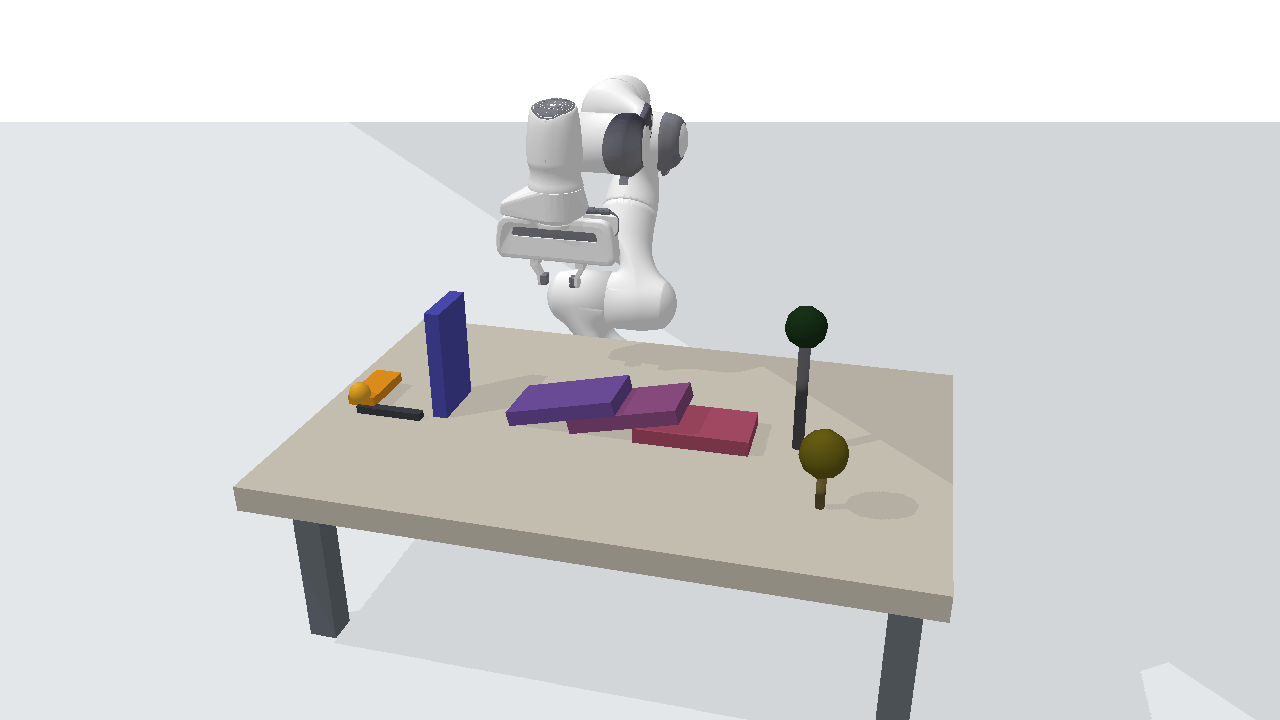}
\end{minipage}
\end{minipage}\hfill
\begin{minipage}{.49\linewidth}
\centering\begingroup
\newfontfamily\DominoFigureFont{texgyreheros-bold.otf}
\definecolor{DominoBlue}{HTML}{DCEFFC}
\resizebox{\linewidth}{!}{%
\begin{tikzpicture}[x=1cm,y=1cm,>=stealth,line width=.7pt,
 every node/.style={font=\DominoFigureFont\fontsize{11}{12}\selectfont},
 state/.style={circle,draw=black!80,fill=DominoBlue,minimum size=.56cm,inner sep=1pt},
 object/.style={font=\DominoFigureFont\fontsize{8.5}{9}\selectfont,text=black!75}]
\filldraw[fill=white,draw=black!45] (-.55,-.12) rectangle (8.35,1.34);
\fill[DominoBlue] (-.54,.94) rectangle (8.34,1.33);
\node at (3.9,1.14) {Original causal graph};
\foreach \i/\x/\name in {1/0/Pusher,2/1.3/D1,3/2.6/D2,4/3.9/D3,5/5.2/D4,6/6.5/Bell,7/7.8/Lamp}{
 \node[state] (a\i) at (\x,.55) {s\textsubscript{\i}};
 \node[object] at (\x,.09) {\name};}
\foreach \i/\j in {1/2,2/3,3/4,4/5,5/6,6/7}\draw[->] (a\i)--(a\j);
\filldraw[fill=white,draw=black!45] (-.55,-1.83) rectangle (8.35,-.37);
\fill[DominoBlue] (-.54,-.77) rectangle (8.34,-.38);
\node at (3.9,-.57) {Intervention: do(s\textsubscript{3} = u)};
\foreach \i/\x/\name in {1/0/Pusher,2/1.3/D1,4/3.9/D3,5/5.2/D4,6/6.5/Bell,7/7.8/Lamp}{
 \node[state] (b\i) at (\x,-1.16) {s\textsubscript{\i}};
 \node[object] at (\x,-1.62) {\name};}
\node[state,fill=black!7] (b3) at (2.6,-1.16) {u};
\node[object] at (2.6,-1.62) {D2 (set)};
\foreach \i/\j in {1/2,3/4,4/5,5/6,6/7}\draw[->] (b\i)--(b\j);
\end{tikzpicture}}
\endgroup
\end{minipage}
\caption{Left: simulator scenes before/after releasing domino 2 from a $20^\circ$ tilt; the robot arm is fixed. Right: a separate seven-node structural causal model (SCM): pusher, four dominoes, bell, and light. $\doop(s_3=u)$ replaces the target mechanism and removes incoming edges. The physical preview is neither a prediction nor validation of this graph.}
\label{fig:intro-environment}
\end{figure}

Can an agent predict intervention consequences from an image? Beyond reconstructing the domino scene in Figure~\ref{fig:intro-environment}, intervention queries require a target, its new value, and resulting responses~\citep{scholkopf2021toward}---a distinction relevant to physical-system counterfactuals~\citep{li2020causalvideos} and robotic transfer under environmental changes~\citep{ahmed2021causalworld}.

Our goal is to query pixel-derived variables in environment intervention units. Differently directed graphs can share conditional independences, forming a Markov equivalence class (MEC). Interventions can distinguish candidates, but unique recovery requires sufficient coverage and identification conditions (Section~\ref{sec:problem})~\citep{hauser2012characterization}.

Label-aligned readouts can still have scale, offset, or nonlinear errors that corrupt queries despite correct graph directions. A \emph{verified intervention value} is the attained target state; queries assume a valid setting without modeling actuator execution.

\paragraph{Contribution.}
To address numerical mismatch, we calibrate node-aligned readouts to intervention values with invertible $T$ supervised by verified direct-target values, preserving the frozen visual model. Settings, graph discovery, and mechanism fitting share coordinates; $T^{-1}$ returns predictions to the decoder. Valid, sufficiently oriented estimated graphs support non-target interventional means and counterfactual images conditioned on factual images.

\section{Related Work}\label{sec:related}
\begin{wrapfigure}{R}{.49\textwidth}
\centering
\begingroup
\definecolor{vblue}{HTML}{EEF3F6}
\definecolor{vgreen}{HTML}{EFF3EC}
\definecolor{vsand}{HTML}{F8F3E9}
\definecolor{vink}{HTML}{364650}
\begin{tikzpicture}[x=.85mm,y=.55mm,every node/.style={align=center,text=vink,font=\fontsize{8}{9}\selectfont,inner sep=1pt}]
\fill[vblue] (23,40) circle (22);
\fill[vgreen,opacity=.65] (47,40) circle (22);
\fill[vsand,opacity=.65] (35,21) circle (22);
\begin{scope}
\clip (23,40) circle (22);
\clip (47,40) circle (22);
\fill[orange!18] (35,21) circle (22);
\end{scope}
\draw[blue!30!gray] (23,40) circle (22);
\draw[green!25!gray] (47,40) circle (22);
\draw[brown!40!gray] (35,21) circle (22);
\node[font=\fontsize{8}{9}\selectfont\bfseries] at (16,67) {Causal representations\\from pixels};
\node[font=\fontsize{8}{9}\selectfont\bfseries] at (54,67) {Feedback-driven\\experiments};
\node[font=\bfseries] at (14,46) {1};
\node[font=\bfseries] at (56,46) {2};
\node[font=\bfseries] at (35,8) {3};
\node[font=\bfseries] at (35,52) {4};
\node[font=\bfseries] at (23,27) {5};
\node[font=\bfseries] at (47,27) {6};
\node[font=\fontsize{8}{10}\selectfont\bfseries,text=blue!40!black] at (35,35) {Our work};
\node[font=\fontsize{8}{9}\selectfont\bfseries] at (35,-8) {Explicit causal structure discovery};
\end{tikzpicture}
\par\vspace{3pt}
\setlength{\tabcolsep}{2pt}
\renewcommand{\arraystretch}{1.08}
\fontsize{7}{8.4}\selectfont
\begin{tabularx}{\linewidth}{@{}>{\raggedright\arraybackslash}X>{\raggedright\arraybackslash}X@{}}
\textbf{1}~~\citeauthor{ahuja2023interventional} \citeyearpar{ahuja2023interventional}
&\textbf{2}~~Causal Curiosity\\
&\hspace*{7pt}\citeauthor{sontakke2021causal} \citeyearpar{sontakke2021causal}\\[2pt]
\textbf{3}~~PC-stable
&\textbf{4}~~ICF\\
\hspace*{7pt}\citeauthor{colombo2014order} \citeyearpar{colombo2014order}
&\hspace*{7pt}\citeauthor{thomas2018disentangling} \citeyearpar{thomas2018disentangling}\\
\hspace*{7pt}GIES: \citeauthor{hauser2012characterization} \citeyearpar{hauser2012characterization}
&\\[2pt]
\textbf{5}~~CausalVAE: \citeauthor{yang2021causalvae} \citeyearpar{yang2021causalvae}
&\textbf{6}~~AIT: \citeauthor{scherrer2021active} \citeyearpar{scherrer2021active}\\
\hspace*{7pt}CITRIS: \citeauthor{lippe2022citris} \citeyearpar{lippe2022citris}
&\hspace*{7pt}CBED: \citeauthor{tigas2022interventions} \citeyearpar{tigas2022interventions}
\end{tabularx}
\endgroup

\caption{\small \textbf{Positioning intervention-value queries.} Sets summarize inputs, acquisition, and structure learning; their intersection denotes a joint setting, not a ranking. We align visual estimates with verified values. Numbered work and classification conditions: Appendix~\ref{sec:positioning-details}.}
\label{fig:positioning-venn}
\end{wrapfigure}
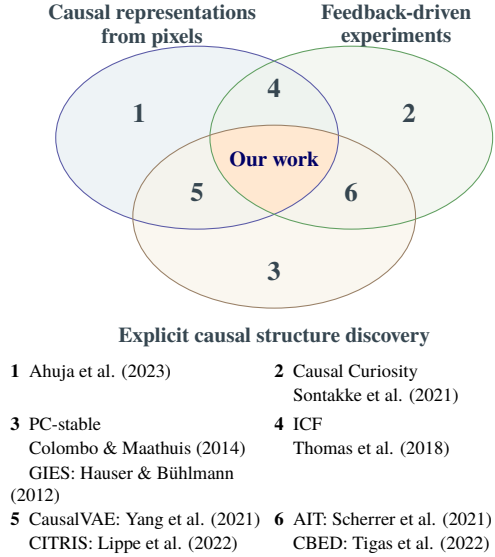
Figure~\ref{fig:positioning-venn} summarizes these distinctions.

\textbf{Visual causal representations.}
CausalVAE learns concept-supervised representations supporting latent-intervention images~\citep{yang2021causalvae}. CITRIS identifies variables from known intervention targets and temporal images, optionally transforming pretrained autoencoder outputs invertibly; iCITRIS handles instantaneous effects, and BISCUIT learns variable--binary-interaction correspondence~\citep{lippe2022citris,lippe2023icitris,lippe2023biscuit}. Implicit latent causal models (ILCM) learn variables and permit structure extraction from intervention pairs~\citep{brehmer2022weakly}; theory also studies intervention-based identification~\citep{ahuja2023interventional}. These capabilities do not ensure numerical alignment with environment settings. We calibrate known-node readouts against verified values to address scale, offset, or nonlinear errors affecting intervention queries.

\paragraph{Active experiments and controllable factors.}
Causal Curiosity distinguishes factors experimentally; independently controllable factors (ICF) learn separately controllable visual factors~\citep{sontakke2021causal,thomas2018disentangling}. Given defined variables, active strategies and AIT choose targets, causal Bayesian experimental design (CBED) also chooses values, and CAASL uses a pretrained policy~\citep{hauser2014two,scherrer2021active,tigas2022interventions,annadani2024amortized}. Visual-readout calibration remains separate: we calibrate round-robin and freeze the map before targeting unoriented edges, giving graph discovery and mechanisms shared intervention-value coordinates.

\paragraph{Response-based coordinate alignment.}
Rep4Ex uses action--latent invariance for intervention extrapolation~\citep{saengkyongam2024identifying}. Ding et al.\ study counterfactual retrieval with a CausalVAE world-model plug-in, aligned during training when state supervision is available~\citep{ding2026causalvae}. We calibrate frozen, known-node visual models against verified values; a matched-data, fully fitted affine control tests nonlinear calibration's incremental benefit.

\section{Problem Formulation}\label{sec:problem}
Given target $j$ and environment value $u$, images inform predictions of settled \emph{post-intervention states}, not frame-by-frame dynamics. Node semantics are known; the direct graph and structural equations are unknown (terminology: Appendix~\ref{app:causal-glossary}).

\paragraph{Environment and true variables.}
Let $V=\{1,\ldots,K\}$ index states $s=(s_1,\ldots,s_K)\in\prod_j\mathcal S_j\subseteq\mathbb R^K$, with domains $\mathcal S_j$ either $\{0,1\}$ for binary events or real intervals for continuous states. Unknown graph $G^\star=(V,\mathcal E^\star)$ has directed edges $\mathcal E^\star$; $\Pa_G(j)$ denotes direct parents. Structural equations $f_j^\star$ and disturbances $\epsilon=(\epsilon_1,\ldots,\epsilon_K)\sim P_\epsilon$ generate states and images:
\begin{equation}
 s_j=f_j^\star(s_{\Pa_{G^\star}(j)},\epsilon_j),\qquad
 x=R(s,\nu)\in[0,1]^{H\times W\times3}.
 \label{eq:short-scm}
\end{equation}
Renderer $R$ produces size-$H,W$ images with imaging factors $\nu$. Neither $R$ nor $f_j^\star$ generates learner predictions; simulator states, graphs, and paired images serve only permitted supervision/evaluation. Superscript $\star$ marks truth; hats mark estimates.

\paragraph{Known inputs and supervision.}
Inputs are observational data $\mathcal D_{\rm obs}=\{(x_i,\ell_i)\}_{i=1}^{N_{\rm obs}}$, eligible targets $\mathcal J\subseteq V$, and feasible values $\mathcal U_j$. Labels $\ell_i\in\mathbb R^K$ supply numerical states and dimension identities, not $\mathcal E^\star$, supervising the $z_j=E_\theta(x)_j$--$s_j$ correspondence. Single-frame ambiguity requires protocol-supplied extra frames or action information, equally available to compared methods.

Decoder coordinates $z_j$ and graph/mechanism readouts $h_j$ both contain visual error. The latter standardize $T_j(z_j)$ with observational statistics; $T_j$ neither permutes nor mixes dimensions. Equation~\eqref{eq:short-coordinates} calibrates images but maps $u$ directly to target setting $q_j(u)$ without $T$.

\paragraph{Environment interventions and budget.}
Operation $a=(j,u)$ specifies target $j\in\mathcal J$ and verified value $u\in\mathcal U_j$. Hard intervention $\doop(s_j=u)$ replaces only the target mechanism:
\begin{equation}
 s_j^{\,a}=u,\qquad
 s_k^{\,a}=f_k^\star(s_{\Pa_{G^\star}(k)}^{\,a},\epsilon_k)
 \quad(k\ne j).
 \label{eq:short-do}
\end{equation}
The environment returns $(j,u,x^{\rm int})$, not full states. $u$ must match the depicted state; release actions and soft interventions require task-specific mechanism-replacement definitions. Equation~\eqref{eq:short-do} is evaluated topologically on acyclic graphs. Population prediction retains the exogenous distribution; individual counterfactuals fix factual $(\epsilon,\nu)$. Observational selection on $s_j=u$ replaces no mechanism and is not intervention.

Budget $B=B_{\rm cal}+B_{\rm graph}$ counts new calibration and graph records, not replay; early stopping reports actual usage. Observational volume is separate. Evidence-dependent targeting precludes treating cross-target records as independent, identically distributed observational samples.

\paragraph{Identification assumptions and scope.}
Assume $G^\star$ is a directed acyclic graph (DAG), no common causes are omitted, $P_\epsilon=\prod_jP_{\epsilon_j}$, and there is no selection bias. Observational states satisfy causal Markov and faithfulness conditions; single-target interventions preserve other mechanisms and the exogenous distribution~\citep{colombo2014order,hauser2012characterization}. Relevant states have finite second moments, and standardization variables have nonzero training variance. Encoder training does not ensure these assumptions. Visual error, approximate tests, and inadequate coverage may prevent recovery. Dimension-wise invertible $T$ adds no population conditional-independence information; unique recovery requires sufficient identification conditions.

Calibration requires a dimension-wise monotone approximation of the discrepancy that remains stable across observational/interventional images. Unintervened dimensions receive only observational-preservation constraints.

\paragraph{Learning objectives.}
We learn encoder $E_\theta$, calibration $T_{\widehat\phi}$, graph $\widehat G$, mechanisms $\widehat f_j$, and decoder $D_\psi$ for non-target responses to node settings. Parameters and normalization statistics use permitted training data only, then freeze (notation and fitted residuals: Appendix~\ref{app:model-objects}).

\textbf{(1) Structure identification.} Compare adjacencies and directions in $\widehat G$ with $G^\star$, retaining partial graphs when evidence is insufficient.
\textbf{(2) Intervention consequence prediction.} For $a=(j,u)$, set node $j$ to verified $u$ and predict each non-target node's mean settled state across contexts:
\[
 \mu_k(j,u)=\mathbb E[s_k\mid\doop(s_j=u)]
 =\mathbb E_{\epsilon\sim P_\epsilon}[s_k^{\,a}(\epsilon)],\qquad k\ne j.
\]
This expectation uses the interventional distribution, not observations selected by $s_j=u$; $s_k^{\,a}(\epsilon)$ follows Equation~\eqref{eq:short-do}. Estimate $\widehat\mu_k(j,u)$ averages across contexts without factual-image conditioning. It is an outcome mean, not an average treatment effect between settings. Excluding $k=j$ tests propagation rather than command copying. Tasks define in-distribution (ID) and out-of-distribution (OOD) splits.
\mbox{\textbf{(3) Counterfactual image generation.}} Compare $\widehat x^{\rm cf}(x,a)$ for factual image $x$ and alternative $a$ with same-context simulator replay. One image need not identify $(\epsilon,\nu)$, so paired evaluation does not establish individual counterfactual identifiability. DAG-based fitting and queries require valid, sufficiently oriented estimates.
\FloatBarrier

\section{Method}\label{sec:method}
\begin{figure}[!htbp]
\centering
\includegraphics[width=\linewidth]{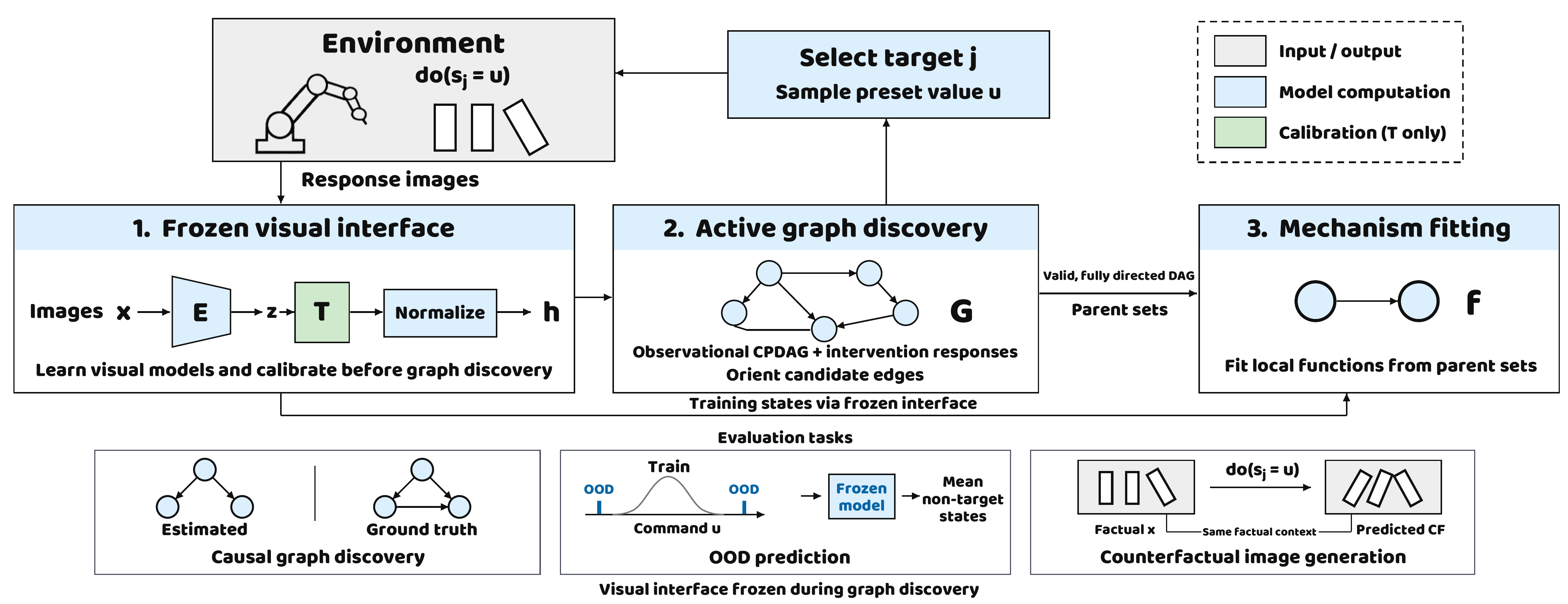}
\caption{Learning and evaluation. The visual model and calibration freeze before active targeting, orientation, and mechanism fitting. $E,T,G,f$: encoder, calibration, graph, mechanisms. Gray/blue/green: inputs/outputs, computation, calibration. OOD excludes A/B-only tests lacking defined shift axes. Curves denote continuous commands (query coordinates: Appendix Figure~\ref{fig:query-interface-r37}).}
\label{fig:method-branches}
\end{figure}

Three stages connect images to intervention outcomes (Figure~\ref{fig:method-branches}). \textbf{First, learn and calibrate readouts.} Concept supervision aligns latents with nodes. Freeze the encoder/decoder, then update only dimension-wise $T$ using verified values from fixed round-robin acquisition. \textbf{Second, recover propagation structure.} Freeze $T$, estimate an observational candidate skeleton, and orient edges using responses to actively targeted environment interventions. \textbf{Third, predict states and images.} Fit local mechanisms on a valid, sufficiently oriented graph for population interventional means and factual-image counterfactuals; inverse calibration returns predictions to decoder coordinates. All modules freeze at test time (query coordinates: Appendix Figure~\ref{fig:query-interface-r37}).

\subsection{Node-Aligned Visual Learning and Readout Calibration}
A concept-supervised variational autoencoder learns reconstruction and latents anchored by standardized state labels (Appendix~\ref{app:standard-modules}). We use the frozen encoder's posterior mean $z=E_\theta(x)$. Jointly trained decoder $D$ may be refitted solely on ID reconstruction as $D'$. Only $T$ is calibrated; each control shares $D$ or $D'$.

\paragraph{Calibrating with verified target states.}
Verified $u$ and encoding $E_\theta(x^{\rm int})_j$ describe the same target state. Training-label mean and standard deviation $\mu_j^s,\sigma_j^s$ define anchor units $\alpha_j(u)=(u-\mu_j^s)/\sigma_j^s$. Superscript $s$ marks a fixed training mean, unlike command-dependent $\mu_k(j,u)$. Records $(j,u,x^{\rm int})$ fit a strictly monotone dimension-wise map
\begin{equation}
 T_{\phi,j}(z_j)=g_jz_j+b_j+
 \sum_{l=1}^{L}w_{jl}\tanh(z_j-\kappa_l),\qquad
 g_j>\sum_l|w_{jl}|.
 \label{eq:short-calibration}
\end{equation}
Parameters $\phi$ comprise scales $g_j$, offsets $b_j$, and weights $w_{jl}$; $L$ hyperbolic-tangent bases have fixed knots $\kappa_l$. The constraint ensures invertibility. For calibration records $\mathcal D_{\rm cal}$, $\lambda\ge0$ balances target supervision and observational preservation; expectations average finite records:
\begin{equation}
 \mathcal L_{\rm cal}=
 \E_{(j,u,x^{\rm int})\sim\mathcal D_{\rm cal}}
 \left[(T_{\phi,j}(E_\theta(x^{\rm int})_j)-\alpha_j(u))^2\right]
 +\lambda\E_{x\sim\mathcal D_{\rm obs}}
 \frac{\|T_\phi(E_\theta(x))-E_\theta(x)\|_2^2}{K}.
 \label{eq:short-cal-loss}
\end{equation}
Only targets receive $u$ supervision; the second term limits observational drift. After predetermined round-robin rounds replaying accumulated records, freeze $T=T_{\widehat\phi}$ at fitted parameters $\widehat\phi$. Training observations supply calibrated means $\mu^c$ and standard deviations $\sigma^c$:
\begin{equation}
 h(x)=\frac{T(E_\theta(x))-\mu^c}{\sigma^c},\qquad
 q_j(u)=\frac{\alpha_j(u)-\mu_j^c}{\sigma_j^c}.
 \label{eq:short-coordinates}
\end{equation}
Readout $h_j(x)$ and setting $q_j(u)$ share coordinates; division is element-wise. Cross-source stability requires independent testing; neither target supervision nor monotonicity ensures it.

\subsection{Observational Structure and Active Orientation}
PC-type conditional-independence tests on $h(\mathcal D_{\rm obs})$ estimate the observational equivalence class as a completed partially directed acyclic graph (CPDAG)~\citep{colombo2014order}; finite-sample $\widehat G_0$ may be wrong. RGB Domino uses linear partial correlations (MEC backends: Appendix~\ref{app:multibench-protocol}).

Each round selects the target with most incident unoriented edges and remaining allowance. Predetermined tie-breaking and ID value distributions use neither the true graph nor test results (Appendix~\ref{app:standard-modules}). New images use frozen $h$. On RGB Domino candidate adjacencies, single-target records $\mathcal I_j$ give
\begin{equation}
 e_{j\to k}=\left|\widehat{\Corr}(h_j,h_k;\mathcal I_j)\right|.
 \label{eq:short-effect}
\end{equation}
Sample correlation $\widehat{\Corr}$ supplies orientation evidence for initial adjacencies, not direct causal effects; targets remain visual estimates, not commands. Bidirectional responses with threshold $\tau=0.15$ and gap $\delta=0.05$ propose orientations. Conflict/acyclicity checks precede Meek propagation~\citep{meek1995}. Appendix~\ref{app:algorithm} specifies sample thresholds, one-sided rules, and rejections. Budget exhaustion retains unresolved edges.

Orientation requires reliable adjacencies and detectable responses. Weak correlation does not exclude nonlinear effects, and graph-consistency checks cannot ensure statistical correctness. Appendix~\ref{app:frozen-diagnostics} tests edge-order dependence and invalid-evidence safeguards.

\subsection{Local Mechanisms and New Commands}\label{sec:downstream-branches}
On valid, sufficiently oriented estimated graphs, ridge regression with an intercept fits $\widehat f_j$ from parent readouts in $h$ coordinates (Appendix~\ref{app:standard-modules}), excluding interventions that replace $j$'s mechanism. Root observations and residuals form sample pools. Invertibility does not imply linearity: nonlinear $T$ can change relationships and residual--parent dependence. Readouts, consequences, and counterfactuals require separate tests (Appendix~\ref{app:mechanism-compatibility}).

\paragraph{Mean-outcome prediction for new commands.}
For $a=(j,u)$, set $q_j(u)$, resample training roots/residuals, and propagate. Convert samples to label units, then average each non-target node $k\ne j$ for $\widehat\mu_k(j,u)$. Inverse-calibrate and decode each sample before averaging images; do not decode a mean latent. Conversions and sampling rule~\eqref{eq:short-rollout} appear in Appendix~\ref{app:query-protocol}.

\subsection{Decoding Counterfactual Images}
Encode, calibrate, and standardize factual $x$ to $h(x)$. For each non-root node, estimate $\eta_j^{\rm fact}=h_j(x)-\widehat f_j(h_{\Pa_{\widehat G}(j)}(x))$. Hold these residuals fixed, set $q_j(u)$, and update descendants to obtain $h^{\rm cf}$. Non-target roots and non-descendants retain factual readouts. Undo standardization and calibration to reach decoder coordinates:
\begin{equation}
 \widehat x^{\rm cf}=D_\psi\!\left[T^{-1}
 \left(\sigma^c\odot h^{\rm cf}+\mu^c\right)\right].
 \label{eq:short-decode}
\end{equation}
Here $\widehat x^{\rm cf}$ is the predicted image, $T^{-1}$ frozen inverse calibration, and $\odot$ element-wise multiplication. Estimated residuals need not equal true exogenous causes. Appendix~\ref{app:query-protocol} also specifies the separately evaluated identity-calibration, conditional-mean configuration without factual residuals.

\section{Experiments}\label{sec:experiments}
Our evidence follows three links: interventions resolve ambiguity left by matched additional observations; same-data calibration lowers mean intervention-effect error in the controlled domino cohort; state and image queries test downstream use. System comparisons assess task performance, not individual components. Decoder-only controls separate rendering gains from calibration (protocol and statistical index: Appendix~\ref{app:evidence-map}).

\subsection{Experimental Setup}

\paragraph{Benchmarks.}
Tasks comprise Domino, Pick-and-Place (PnP), Balance, Gears, CAUSAL3D Spring and Seesaw~\citep{liu2025causal3d}, and CausalWorld~\citep{ahmed2021causalworld}. Domino separately includes a seven-node continuous SCM and four-node contact dynamics. The latter uses binary events $F_i$ for whether each domino falls and evaluates fixed/release commands through same-initial-state replay. PnP, Balance, and Gears contain six, five, and six SCM variables. Figure~\ref{fig:benchmark-scenes-r32} illustrates scenes; Domino configurations and task-table scope appear in Appendices~\ref{app:reported-protocol} and~\ref{app:run-index}.

\paragraph{Baselines.}
System comparisons include CausalVAE~\citep{yang2021causalvae} and ILCM, which learns from intervention-paired images~\citep{brehmer2022weakly}. Each uses its own representation, graph extraction, latent intervention, and decoder. ICM-VAE is a given-graph generation reference~\citep{komanduri2024icmvae}; no-graph controls remove dependencies through their cohort-specific operators (Appendix~\ref{app:implementation-rules}). Appendix~\ref{app:baseline-health} reports ID quality and model selection. Weak baseline results are not attributed to an absence of intervention data.

\paragraph{Metrics.}
Graph discovery reports structural Hamming distance (SHD)$\downarrow$, counting a reversed edge once, and area under the precision--recall curve (AUPRC)$\uparrow$. Consequence prediction reports training-scale-normalized mean squared error (NMSE)$\downarrow$ for non-target mean states. Counterfactual images use RGB mean squared error (MSE)$\downarrow$. For counterfactual images, models receive only a factual image and command; ground truth (GT) replays the same exogenous context or physical initial state. Population prediction instead takes only a command. Errors in the latter two task tables are multiplied by $10^3$. Completion, graph recovery, and executable queries have separate denominators (Appendix~\ref{app:metrics-r32}).

\paragraph{Comparison scope.}
Our method uses learned $T$ in the three task tables. Appendix~\ref{app:comparison-scope} distinguishes unified comparisons, given-graph references, and physical ILCM with different permissions. These descriptive summaries measure conditional error on executed queries, not end-to-end success or superiority on a common query population. Their protocol coverage is specified in Appendix~\ref{app:benchmark-protocol-summary}; the matched-data and matched-acquisition controls, rather than these aggregate gaps, support component attribution. Continuous-task columns retain reported ID/amplitude-shift labels; other columns compare test sets A/B (Appendix~\ref{app:ood-definitions-r30}). Light blue marks our method, bold marks the best point estimate, and ``--'' means not applicable. Acquisition efficiency uses the total budget $B=B_{\rm cal}+B_{\rm graph}$.

\begin{figure}[!htbp]
\centering
\includegraphics[width=\linewidth]{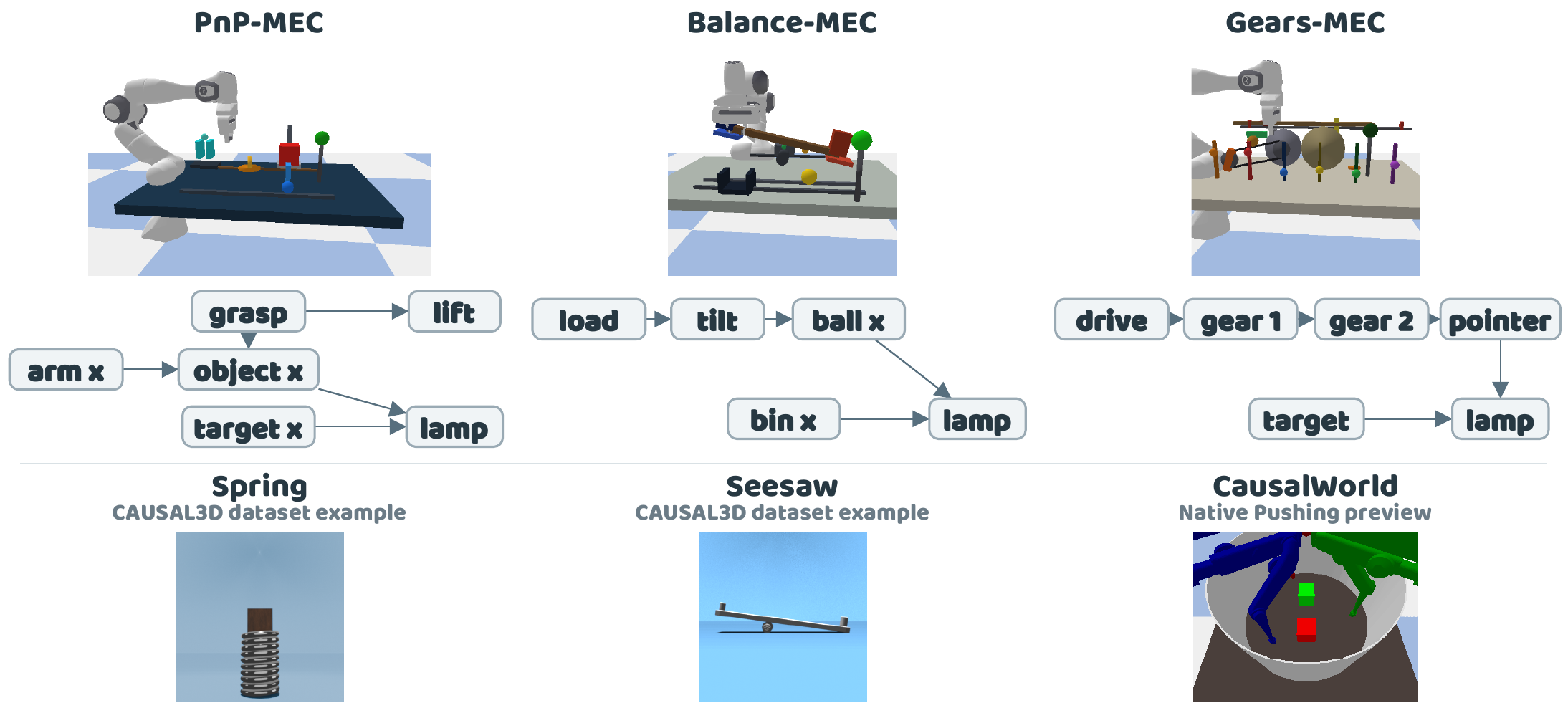}
\caption{Example environments. Top: controlled MEC scenes and generating graphs for PnP, Balance, and Gears. Bottom: original CAUSAL3D Spring and Seesaw images and a native CausalWorld Pushing illustration. Generating graphs are not estimates; scene illustrations do not represent task-table configurations or model predictions. Sources appear in Appendix~\ref{app:benchmark-gallery-r32}.}
\label{fig:benchmark-scenes-r32}
\end{figure}

\subsection{Discovering Graphs through Environment Interventions}\label{sec:exp-graph}
A three-seed control shares the visual and calibration interface and matches the number of added records for graph discovery: both intervention acquisition strategies recover the true graph with 1800 or 3600 records, while equal additional observations leave 6 unoriented edges. In this controlled task, resolving the remaining orientation ambiguity requires intervention responses; equal additional observations do not suffice. Coverage targeting does not save records over round-robin acquisition. Appendices~\ref{app:query-evidence-r38} and~\ref{app:multibench-protocol} give the protocols and diagnostics for different orientation criteria.

\begin{inlinetable}
\centering\scriptsize\setlength{\tabcolsep}{2pt}\renewcommand{\arraystretch}{1.25}
\caption{Graph discovery: each cell reports SHD$\downarrow$/AUPRC$\uparrow$ on its upper/lower line. Bold marks the best point estimate; light blue marks our method. Bottom: completed/planned runs, not graph-recovery rates. See Appendix~\ref{app:unified-evidence} for comparison scope.}
\label{tab:pilot}
\begin{tabularx}{\linewidth}{L{.17\linewidth}*{8}{Y}}
\toprule
 & \multicolumn{2}{c}{Domino} & \multicolumn{3}{c}{Other MEC tasks} & \multicolumn{2}{c}{CAUSAL3D} & \shortstack{Causal\\World}\\
\cmidrule(lr){2-3}\cmidrule(lr){4-6}\cmidrule(lr){7-8}
Method & SCM & Physical & PnP & Balance & Gears & Spring & Seesaw & \\
\midrule
\rowcolor{OurRowBlue}
\textbf{Ours} & \datum{tables-unified-graph-tex-c0001} & \datum{tables-unified-graph-tex-c0002} & \datum{tables-unified-graph-tex-c0003} & \datum{tables-unified-graph-tex-c0004} & \datum{tables-unified-graph-tex-c0005} & \datum{tables-unified-graph-tex-c0006} & \datum{tables-unified-graph-tex-c0007} & \datum{tables-unified-graph-tex-c0008}\\
CausalVAE & \datum{tables-unified-graph-tex-c0009} & \datum{tables-unified-graph-tex-c0010} & \datum{tables-unified-graph-tex-c0011} & \datum{tables-unified-graph-tex-c0012} & \datum{tables-unified-graph-tex-c0013} & \datum{tables-unified-graph-tex-c0014} & \datum{tables-unified-graph-tex-c0015} & \datum{tables-unified-graph-tex-c0016}\\
ILCM & \datum{tables-unified-graph-tex-c0017} & \datum{tables-unified-graph-tex-c0018} & \datum{tables-unified-graph-tex-c0019} & \datum{tables-unified-graph-tex-c0020} & \datum{tables-unified-graph-tex-c0021} & \datum{tables-unified-graph-tex-c0022} & \datum{tables-unified-graph-tex-c0023} & \datum{tables-unified-graph-tex-c0024}\\
\midrule
\multicolumn{9}{l}{\textit{Completed/planned runs$\uparrow$ (not graph-recovery rates)}}\\
\rowcolor{OurRowBlue}
\textbf{Ours} & \datum{tables-unified-graph-tex-c0025} & \datum{tables-unified-graph-tex-c0026} & \datum{tables-unified-graph-tex-c0027} & \datum{tables-unified-graph-tex-c0028} & \datum{tables-unified-graph-tex-c0029} & \datum{tables-unified-graph-tex-c0030} & \datum{tables-unified-graph-tex-c0031} & \datum{tables-unified-graph-tex-c0032}\\
\bottomrule
\end{tabularx}
\end{inlinetable}

In Table~\ref{tab:pilot}, our SHD is best in six of eight columns and AUPRC in seven; SHD is 0 for physical Domino and PnP. The complete system yields more accurate structure estimates on most tasks, while the shared-interface control above shows that graph-stage intervention responses supply orientation evidence still absent from additional observations. The advantage does not cover every metric: SHD is higher than ILCM on Balance and Gears, and AUPRC is 0.242 on Spring and 0.276 on CausalWorld, leaving substantial full-structure recovery error.

\subsection{Intervention-Value Calibration and Effects}\label{sec:ablation-main}
Table~\ref{tab:calibration-controlled} compares readouts, non-target counterfactual (CF) effects, and RGB errors in the same three-seed cohort for the seven-node domino scene rendered as $64\times64$ RGB images. Within each seed, configurations share test contexts and commands. The first three rows share an ID-refitted decoder $D'$. Ordinary joint updating additionally trains a readout head and decoder, so it is included only as a different-permission reference. CF effects compare differences from the factual state and are not pooled with population mean-outcome NMSE.

\begin{inlinetable}
\centering\small\setlength{\tabcolsep}{3pt}
\caption{Readouts, intervention effects, and images in the same paired three-seed cohort. Errors are multiplied by $10^3$; entries are means $\pm$ sample SD across seeds. The first three rows share each seed's frozen encoder, data, learned graph, and ID-refitted decoder $D^{\prime}$, fitting mechanisms in their respective coordinates. Ordinary joint updating additionally trains a readout head and decoder and is reference-only. All seeds 5/6/7 are scored; bold marks the lowest mean in the first three rows. Paired differences appear in Appendix~\ref{app:paired-evidence}.}
\label{tab:calibration-controlled}
\begin{tabularx}{\linewidth}{Xrrrrr}\toprule
 & ID readout & \multicolumn{2}{c}{\shortstack{Non-target CF effect\\NMSE$\downarrow$}} & \multicolumn{2}{c}{RGB MSE$\downarrow$}\\
\cmidrule(lr){3-4}\cmidrule(lr){5-6}
Configuration & MSE$\downarrow$ & ID & Amp. shift & ID & Amp. shift\\\midrule
Uncalibrated & \datum{tables-calibration-controlled-tex-c0001} & \datum{tables-calibration-controlled-tex-c0002} & \datum{tables-calibration-controlled-tex-c0003} & \datum{tables-calibration-controlled-tex-c0004} & \datum{tables-calibration-controlled-tex-c0005}\\
Fully fitted affine & \datum{tables-calibration-controlled-tex-c0006} & \datum{tables-calibration-controlled-tex-c0007} & \datum{tables-calibration-controlled-tex-c0008} & \datum{tables-calibration-controlled-tex-c0009} & \datum{tables-calibration-controlled-tex-c0010}\\
\rowcolor{OurRowBlue}
Nonlinear (ours) & \datum{tables-calibration-controlled-tex-c0011} & \datum{tables-calibration-controlled-tex-c0012} & \datum{tables-calibration-controlled-tex-c0013} & \datum{tables-calibration-controlled-tex-c0014} & \datum{tables-calibration-controlled-tex-c0015}\\
\midrule
Ordinary joint updating (ref.) & \datum{tables-calibration-controlled-tex-c0016} & \datum{tables-calibration-controlled-tex-c0017} & \datum{tables-calibration-controlled-tex-c0018} & \datum{tables-calibration-controlled-tex-c0019} & \datum{tables-calibration-controlled-tex-c0020}\\
\bottomrule\end{tabularx}
\end{inlinetable}

Calibration places visual readouts and target settings in shared intervention units; its effectiveness must be assessed through intervention consequences, not state-readout MSE alone. Ordinary joint updating has lower mean readout error than nonlinear calibration (11.361 versus 12.864), but higher ID/amplitude-shift effect errors (9.985/20.449 versus 4.968/6.515). Ordinary updating changes both the readout head and decoder, so this comparison does not by itself establish the contribution of invertibility.

Nonlinear calibration has lower mean error than no calibration for both effects, improving 2/3 seeds in each case. Relative to affine calibration, its amplitude-shift effect error is lower, but ID effect and RGB errors are higher. These results support a reduction in mean intervention-effect error through calibration in this cohort, not nonlinear superiority across all queries. A two-seed fixed-graph Gears control also slightly improves readouts while increasing query error (Appendix~\ref{app:paired-evidence}). Appendix~\ref{app:retest-r42} defines the effect and within-cohort statistical units.

\subsection{Consequence Prediction under Distribution Shifts}\label{sec:exp-ood}
With models frozen, we predict the other variables' mean post-intervention states under $(j,u)$. Table~\ref{tab:ood-benchmark-matrix} separates reported ID/amplitude-shift columns from A/B tests and gives both absolute NMSEs and their ratio $r$. The second-group error is never smaller than the first-group error. A ratio closer to 1 indicates less relative growth, but increasing the first-group error also raises $r$; absolute errors must therefore accompany its interpretation.

\begin{inlinetable}
\centering\scriptsize\setlength{\tabcolsep}{2pt}\renewcommand{\arraystretch}{1.2}
\caption{Conditional consequence-prediction errors, excluding blocked queries. Unmarked columns retain the reported ID/amplitude-shift labels; entries are NMSEs ($\times10^3$) and ratio $r$. Columns marked $\dagger$ report test sets A, B, and $E_A/E_B$, without inferring an OOD type. Bold marks the smallest listed error summary; light blue marks our method. Lower errors are better; $r$ alone does not define a ranking. Scope and distribution qualifications: Appendices~\ref{app:comparison-scope} and~\ref{app:ood-definitions-r30}.}
\label{tab:ood-benchmark-matrix}
\begin{tabularx}{\linewidth}{L{.17\linewidth}*{8}{Y}}
\toprule
 & \multicolumn{2}{c}{Domino} & \multicolumn{3}{c}{Other MEC tasks} & \multicolumn{2}{c}{CAUSAL3D} & \shortstack{Causal\\World}\\
\cmidrule(lr){2-3}\cmidrule(lr){4-6}\cmidrule(lr){7-8}
Method & SCM & Physical$^\dagger$ & PnP & Balance & Gears & Spring$^\dagger$ & Seesaw$^\dagger$ & $\dagger$\\
\midrule
\rowcolor{OurRowBlue}
\textbf{Ours} & \datum{tables-unified-ood-tex-c0001} & \datum{tables-unified-ood-tex-c0002} & \datum{tables-unified-ood-tex-c0003} & \datum{tables-unified-ood-tex-c0004} & \datum{tables-unified-ood-tex-c0005} & \datum{tables-unified-ood-tex-c0006} & \datum{tables-unified-ood-tex-c0007} & \datum{tables-unified-ood-tex-c0008}\\
CausalVAE & \datum{tables-unified-ood-tex-c0009} & \datum{tables-unified-ood-tex-c0010} & \datum{tables-unified-ood-tex-c0011} & \datum{tables-unified-ood-tex-c0012} & \datum{tables-unified-ood-tex-c0013} & \datum{tables-unified-ood-tex-c0014} & \datum{tables-unified-ood-tex-c0015} & \datum{tables-unified-ood-tex-c0016}\\
ILCM & \datum{tables-unified-ood-tex-c0017} & \datum{tables-unified-ood-tex-c0018} & \datum{tables-unified-ood-tex-c0019} & \datum{tables-unified-ood-tex-c0020} & \datum{tables-unified-ood-tex-c0021} & \datum{tables-unified-ood-tex-c0022} & \datum{tables-unified-ood-tex-c0023} & \datum{tables-unified-ood-tex-c0024}\\
\midrule
No graph & \datum{tables-unified-ood-tex-c0025} & \datum{tables-unified-ood-tex-c0026} & \datum{tables-unified-ood-tex-c0027} & \datum{tables-unified-ood-tex-c0028} & \datum{tables-unified-ood-tex-c0029} & \datum{tables-unified-ood-tex-c0030} & \datum{tables-unified-ood-tex-c0031} & \datum{tables-unified-ood-tex-c0032}\\
\bottomrule
\end{tabularx}
\end{inlinetable}

For executed queries, our point estimates are lowest for both test groups in all eight columns, with $r$ ranging from 0.508--0.896, versus 0.101--0.507 for baselines and the no-graph ablation. Arithmetic differences between the reported second-group conditional errors are approximately 97.8\%--99.8\% relative to the lower CausalVAE/ILCM entry in each column. These gaps describe the listed implementations and test sets, not end-to-end success or equally large OOD advantages. Interpretation requires the distribution definition and ID fitting quality.

The no-graph rows have higher second-group error throughout. This is a cohort-specific dependency ablation, not necessarily an unchanged-mechanism propagation switch (Appendix~\ref{app:implementation-rules}). The factual-conditioned effect NMSE in Section~\ref{sec:ablation-main} evaluates a different query from the population mean error here.

\subsection{Counterfactual RGB Image Generation}\label{sec:exp-cf}
We fix the factual context and generate images through latent interventions and each model's decoder. Among the reported summaries in Table~\ref{tab:cf-pixel-matrix}, our pixel MSE is numerically lowest in seven of eight columns: 68.6\% below CausalVAE on PnP and 26.8\% below ILCM on Spring, but 57.9\% above ILCM on Seesaw. Whole-image MSE combines reconstruction and consequence errors. Causal interpretation requires non-target state effects and independent image readouts of target execution, descendant response, and non-descendant preservation; sharper images alone are insufficient.

\begin{inlinetable}
\centering\scriptsize\setlength{\tabcolsep}{2pt}\renewcommand{\arraystretch}{1.35}
\caption{Counterfactual RGB generation: pixel MSE$\downarrow$ ($\times10^{-3}$) for executed queries only, excluding blocked queries. Bold marks the smallest listed summary; light blue marks our method; ``--'' means not applicable. Physical ILCM is a single-seed reference with different permissions and budget, excluded from unified ranking. Appendix~\ref{app:comparison-scope} specifies the comparison scope.}
\label{tab:cf-pixel-matrix}
\begin{tabularx}{\linewidth}{L{.17\linewidth}*{8}{Y}}
\toprule
 & \multicolumn{2}{c}{Domino} & \multicolumn{3}{c}{Other MEC tasks} & \multicolumn{2}{c}{CAUSAL3D} & \shortstack{Causal\\World}\\
\cmidrule(lr){2-3}\cmidrule(lr){4-6}\cmidrule(lr){7-8}
Method & SCM & Physical & PnP & Balance & Gears & Spring & Seesaw & \\
\midrule
\rowcolor{OurRowBlue}
\textbf{Ours} & \datum{tables-unified-cf-tex-c0001} & \datum{tables-unified-cf-tex-c0002} & \datum{tables-unified-cf-tex-c0003} & \datum{tables-unified-cf-tex-c0004} & \datum{tables-unified-cf-tex-c0005} & \datum{tables-unified-cf-tex-c0006} & \datum{tables-unified-cf-tex-c0007} & \datum{tables-unified-cf-tex-c0008}\\
CausalVAE & \datum{tables-unified-cf-tex-c0009} & \datum{tables-unified-cf-tex-c0010} & \datum{tables-unified-cf-tex-c0011} & \datum{tables-unified-cf-tex-c0012} & \datum{tables-unified-cf-tex-c0013} & \datum{tables-unified-cf-tex-c0014} & \datum{tables-unified-cf-tex-c0015} & \datum{tables-unified-cf-tex-c0016}\\
ILCM & \datum{tables-unified-cf-tex-c0017} & \datum{tables-unified-cf-tex-c0018} & \datum{tables-unified-cf-tex-c0019} & \datum{tables-unified-cf-tex-c0020} & \datum{tables-unified-cf-tex-c0021} & \datum{tables-unified-cf-tex-c0022} & \datum{tables-unified-cf-tex-c0023} & \datum{tables-unified-cf-tex-c0024}\\
ICM-VAE (given graph) & -- & -- & -- & -- & -- & \datum{tables-unified-cf-tex-c0025} & \datum{tables-unified-cf-tex-c0026} & --\\
\midrule
No graph & \datum{tables-unified-cf-tex-c0027} & \datum{tables-unified-cf-tex-c0028} & \datum{tables-unified-cf-tex-c0029} & \datum{tables-unified-cf-tex-c0030} & \datum{tables-unified-cf-tex-c0031} & \datum{tables-unified-cf-tex-c0032} & \datum{tables-unified-cf-tex-c0033} & \datum{tables-unified-cf-tex-c0034}\\
\bottomrule
\end{tabularx}
\end{inlinetable}

\begin{figure}[!htbp]
\centering
\includegraphics[width=\linewidth]{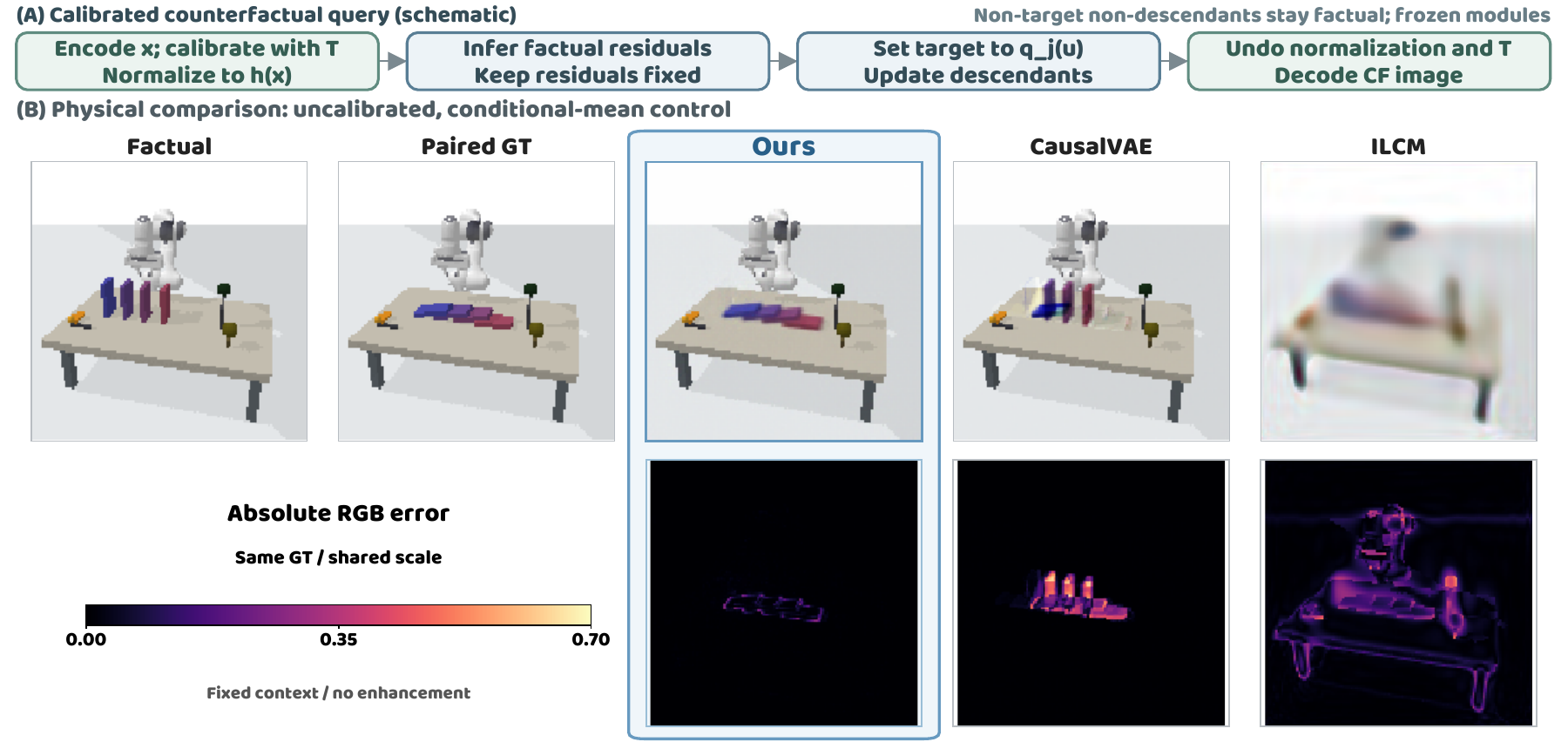}
\caption{(A) Schematic of learned calibration, factual-residual propagation, and inverse-calibrated decoding. (B) Physical comparison for making the first domino fall ($F_1=1$). The blue \emph{Ours} column denotes our conditional-mean comparison branch, not the full calibration-and-residual path in A. Factual images, paired GT, and model outputs are above RGB mean absolute errors with a shared scale. Images are clipped to $[0,1]$ for display only. ILCM is a single-seed reference with different training permissions (Appendix~\ref{app:ilcm-physical-demo}).}
\label{fig:rigidbody-causalvae}
\end{figure}

Figure~\ref{fig:rigidbody-causalvae}A summarizes the proposed counterfactual query path. Panel B compares our physical conditional-mean branch, labeled \emph{Ours}, with CausalVAE and ILCM in one fixed context. The schematic in A and the comparison branch in B are distinct configurations; this image comparison does not measure calibration gains. Appendix Figure~\ref{fig:full-calibrated-cf} gives paired images with learned calibration and factual residuals, and Appendix~\ref{app:interface-r39} reports independent image-readout evaluation.

\paragraph{Paired decoder control.}
Replacing only the original $D$ with the ID-refitted decoder $D'$ lowers amplitude-shift CF pixel MSE by 49.32\% (3/3), with unchanged state predictions. Uncalibrated and affine models also benefit (Appendix~\ref{app:decoder-only-r40}); Table~\ref{tab:calibration-controlled} instead shares $D'$ to compare calibration. Appendix~\ref{app:ablation-complete-r33} gives other component and budget controls.

\FloatBarrier
\section{Conclusion}\label{sec:conclusion}
With known node semantics, verified intervention values calibrate visual readouts so that node settings, causal mechanisms, and decoding use consistent intervention units. Matched controls support improvements for some intervention queries, but the relative benefits of affine and nonlinear calibration depend on the setting; better readouts do not automatically improve downstream predictions. Environment interventions provide direction evidence absent from additional observations. This finding is distinct from targeting efficiency and gains in image reconstruction.

\label{main-text-end}
\clearpage
\section*{AI Use Statement}
Generative AI tools assisted result analysis, writing and translation, methodological and experimental discussions, code implementation, and figure preparation. AI-assisted analyses were subsequently checked by the authors. The authors led the research motivation, methodology, and scientific decisions, and take responsibility for the experiments, reported results, and final manuscript.

\section*{Reproducibility Statement}
The appendix specifies the controlled-cohort protocols and disaggregated results. Appendix~\ref{app:comparison-scope} defines comparison permissions and statistical units, and Appendix~\ref{app:benchmark-protocol-summary} distinguishes specified protocols from aggregate-only system summaries. Training code, raw data, and model checkpoints are private; the project website provides viewing material, not a public implementation release.

\paragraph{Additional website demonstrations.}
The website also includes recorded robot-contact simulations. Its multi-round physical fork uses an exploratory revisable-graph scoring extension, distinct from the observational-skeleton orientation procedure in Section~\ref{sec:method}. These demonstrations are separate from the paper's benchmark tables and do not establish calibration gains or general graph-recovery guarantees.


\clearpage
\appendix
\raggedbottom
\renewcommand{\thetable}{A\arabic{table}}\setcounter{table}{0}
\renewcommand{\thefigure}{A\arabic{figure}}\setcounter{figure}{0}
\renewcommand{\theHtable}{appendix.\arabic{table}}
\renewcommand{\theHfigure}{appendix.\arabic{figure}}

\FloatBarrier
\section{Experimental Protocols, Statistical Units, and Environments}
\subsection{Research questions and protocol index}\label{app:evidence-map}
The appendix separates system performance from component attribution. Tables~\ref{tab:pilot}, \ref{tab:ood-benchmark-matrix}, and~\ref{tab:cf-pixel-matrix} compare complete systems; Appendix~\ref{app:comparison-scope} defines their permissions and denominators. The controlled evidence answers four further questions:
\begin{itemize}[leftmargin=1.4em,itemsep=0pt]
\item \textbf{Orientation:} equal added observations versus interventions in three seeds (Appendix~\ref{app:query-evidence-r38}), distinguished from the eight-seed MEC study (Appendix~\ref{app:multibench-protocol}). Calibration interventions are shared, so the former isolates graph-stage information rather than a wholly passive pipeline.
\item \textbf{Calibration:} Table~\ref{tab:calibration-controlled} fixes each seed's encoder, graph, queries and $D'$, refitting mechanisms in each coordinate system; paired differences are in Appendix~\ref{app:paired-evidence}. Ordinary updating has additional readout/decoder permissions.
\item \textbf{Decoding:} replacing only $D$ with $D'$ leaves the three-seed causal-state predictions fixed (Appendix~\ref{app:decoder-only-r40}).
\item \textbf{Acquisition:} total-budget comparisons include calibration cost and report recovery and exhaustion for all three seeds under batch orientation checks (Appendix~\ref{app:ablation-complete-r33}).
\end{itemize}
All test-time modules, scales and query rules are frozen. Appendix~\ref{app:calibration-chain-r30} uses configuration-specific graphs from shared evidence; Appendix~\ref{app:interface-r39} relearns graphs from a shared trajectory. These differ from the main fixed-graph control and from fixed-graph Gears, which does not test rediscovery. Metrics and shifts are defined once in Appendices~\ref{app:metrics-r32} and~\ref{app:ood-definitions-r30}; affine-favoring conditions and negative Gears results retain their original cohorts.

\subsection{Comparison scope and statistical units}\label{app:comparison-scope}
The system tables use the reported unified data/supervision setting and learned $T$ for our method. Additional permissions define separate references: ICM-VAE receives a graph, while physical ILCM uses different paired data and budget ($n=1$; 32 contexts, four targets, two commands; 256/256 queries). Other MEC baseline diagnostics refer only to the seed300 checkpoints ($n=1$). Cohort configurations are indexed in Appendix~\ref{app:run-index}.

\paragraph{Inputs and scoring.}
Counterfactual generation receives a factual image and target/value; population prediction receives only the command and scores non-target states. Training normalization and all query-time parameters are frozen. Graph, state and image metrics are not pooled. Tables~\ref{tab:ood-benchmark-matrix} and~\ref{tab:cf-pixel-matrix} score executed queries: blocked cases receive neither zero error nor an arbitrary penalty. Different failures can leave different scoring sets, so these conditional errors do not measure end-to-end success or superiority on a common query population. Completion fractions in Table~\ref{tab:pilot} count completed/planned runs, not exact graphs or prediction queries; Table~\ref{tab:benchmark-protocol-summary} keeps these fields separate.

\paragraph{Matched controls and uncertainty.}
Table~\ref{tab:calibration-controlled} uses seeds 5/6/7, all retained, with 64 shared factual contexts, seven targets and two commands per target/split: 896 CF queries per seed/split. Targets and commands are equally weighted within seed; the mean and sample SD are computed over three seed-level errors, not 2688 independent images. Paired differences in Appendix~\ref{app:paired-evidence} use the same seeds; their SD is not the difference of marginal SDs. Uncertainty and denominators are not transferred to aggregate cells lacking them, nor are separate physical aggregates used for paired calibration gains.

The three calibration rows share encoder, data, graph and $D'$; ordinary updating also trains readout/decoder parameters. Architectures need not match under equal data access. No-graph operators are cohort-specific (Appendix~\ref{app:implementation-rules}); equal-total-budget policy comparisons are confined to Appendix~\ref{app:ablation-complete-r33}. Decoder and orientation gains alone do not identify calibration's contribution.

\subsection{Evaluation configurations and statistical scope}\label{app:run-index}
The task tables report learned $T$ under a unified data and supervision protocol. Table~\ref{tab:run-index} distinguishes system comparisons, physical-scene adaptations, and mechanism controls; counts and configurations are interpreted separately for each cohort.

\FloatBarrier
\begin{inlinetable}
\centering\small\setlength{\tabcolsep}{3pt}
\caption{Physical and MEC cohort configurations. System-table comparison rules are in Appendix~\ref{app:comparison-scope}; denominators below belong only to their named cohorts.}
\label{tab:run-index}
\begin{tabularx}{\linewidth}{L{.23\linewidth}L{.37\linewidth}X}
\toprule
Result location & Configuration and record & Queries, permissions, and scope\\\midrule
Physical ILCM in Table~\ref{tab:cf-pixel-matrix}
& \texttt{run\_s42\_v1}, seed 42; native implicit mechanisms and its own decoder; epoch 120 selected by ID validation.
& 1280 training pairs, 256 validation pairs; 32 contexts$\times$4 targets$\times$2 commands, 256/256 queries. Paired GT matches archived seed-42 queries, not necessarily the retained sets of other task-table summaries. Permissions and budget differ (Appendix~\ref{app:ilcm-physical-demo}).\\
MEC mechanism controls in Appendix~\ref{app:new-mec-queries}
& Seeds 300/301, learned $T$; frozen representation, calibration, and graph mechanisms; a shared decoder trained on 2400 independent ID images.
& PnP/Gears retain 2/2 seeds and Balance 1/2; respectively 384/384/320 CF queries per valid seed. Command ranges and 2048-sample outcome evaluation apply only to this cohort.\\
Uncalibrated physical control (aggregate)
& No numerical calibration; aggregate full-image pixel MSE is \datum{appendix-run-index-tex-c0001} ($\times10^{-3}$).
& Reported separately from the paired control below; its seed count, SD, query count, and training budget are not transferred. This aggregate is not used with the main-table learned-$T$ result to estimate paired calibration gains.\\
Paired conditional-mean control in Appendix~\ref{app:reported-protocol}
& No numerical calibration or addition of factual residuals to descendants; common completed seeds 42 and 314.
& 256 common queries per seed; our method completes 2/3 seeds and CausalVAE 3/3, with more concept labels for the latter. Disaggregated errors are reported in Appendix~\ref{app:reported-protocol}.\\
\bottomrule
\end{tabularx}
\end{inlinetable}

\paragraph{Benchmark protocol summary.}\label{app:benchmark-protocol-summary}
Table~\ref{tab:benchmark-protocol-summary} records the protocol coverage of the system-table summaries. The specified controls below separately support claims about calibration and intervention information. Common input, metric, freeze, and exclusion rules (Appendix~\ref{app:comparison-scope}) do not supply missing numerical sampling ranges or executed-query sets.

\begin{inlinetable}
\centering\footnotesize\setlength{\tabcolsep}{3pt}
\caption{Protocol coverage of the system-table summaries. Completed/planned run counts in Table~\ref{tab:pilot} are not prediction seed counts $n$, executed queries $Q$, or calibration-inclusive totals $B$. Those fields are not reported for the aggregate cells and are not inferred from the controls below.}
\label{tab:benchmark-protocol-summary}
\begin{tabularx}{\linewidth}{L{.22\linewidth}L{.35\linewidth}X}
\toprule
Benchmark & Task variables & Operation and test grouping\\
\midrule
Domino SCM & 7 continuous states: pusher, four dominoes, bell, lamp & Single-state setting; reported ID/amplitude groups\\
\addlinespace
Physical Domino & 4 binary final toppling events & Fix upright / tilt and release; test sets A/B\textsuperscript{*}\\
\addlinespace
PnP & 6 SCM variables & Single-state setting; reported ID/amplitude groups\\
\addlinespace
Balance & 5 SCM variables & Single-state setting; reported ID/amplitude groups\\
\addlinespace
Gears & 6 SCM variables & Single-state setting; reported ID/amplitude groups\\
\addlinespace
CAUSAL3D Spring & Scored variables not reported & Test sets A/B; targets and ranges not reported\\
\addlinespace
CAUSAL3D Seesaw & Scored variables not reported & Test sets A/B; targets and ranges not reported\\
\addlinespace
CausalWorld & Task variant and scored variables not reported & Test sets A/B; targets and ranges not reported\\
\bottomrule
\end{tabularx}
\par\smallskip\raggedright
Unreported fields are not zero or failed runs. The aggregate scores alone do not identify training/test generators, numerical ranges, uncertainty, or a common retained-query population. In particular, the Spring, Seesaw, and CausalWorld A/B summaries do not establish a specific OOD shift or acquisition efficiency; official benchmark defaults and scene illustrations cannot determine these runs' settings. \textsuperscript{*}Physical ILCM is a separately documented reference: seed 42, 1280 training pairs, 256 validation pairs, and 256/256 CF queries. Its 3072 pair-generation image calls use different permissions/accounting, not a matched $B_{\rm cal}+B_{\rm graph}$ total. These values do not fill other methods' cells.
\end{inlinetable}
\FloatBarrier

\paragraph{Independently specified controls.}
These controls retain their own sampling protocols rather than supplying missing task-table fields.
\begin{itemize}[leftmargin=1.4em,itemsep=0pt]
\item \textbf{Main matched-data calibration control (Table~\ref{tab:calibration-controlled}; Appendix~\ref{app:paired-evidence}).}
Seeds 5/6/7 each share 64 factual contexts, seven targets, and two commands per split: 896 queries per seed/split, or 1792 across ID and amplitude shift. Commands are $\mu_j^s\pm0.75\sigma_j^s$ and $\mu_j^s\pm2\sigma_j^s$, respectively. The three calibration variants share each seed's frozen encoder, learned graph, records, and ID-refitted decoder $D'$; mechanisms refit in their respective coordinates. Means and sample SD use three seed-level errors, not independent image counts.
\item \textbf{Matched-acquisition Domino (Appendix~\ref{app:interface-r39}).}
All seven targets, seeds 5/6/7, 6000 observations, 896 calibration records, and 1800/3600/3600 new graph records give $B=896+1800$ for seed 5 and $B=896+3600$ for 6/7. Commands use training-label statistics: $\mu_j^s\pm0.75\sigma_j^s$ on independent ID contexts and $\mu_j^s\pm2\sigma_j^s$ for amplitude tests. Each seed executes 1792 CF queries over both splits; graph/decoder conditions differ from Table~\ref{tab:calibration-controlled}.
\item \textbf{Paired physical control (Appendix~\ref{app:reported-protocol}).}
Four binary events use fix/release commands; fitting uses 512 labeled images, 128 ID validation images and 512 independent observations. Seeds 42/314 execute and 2718 is blocked ($n=2/3$); 32 held-out initial states give 256 CF queries per completed seed. The known acquisition cost is 256 graph records, not a calibration-inclusive total or a denominator for the separate aggregate in Table~\ref{tab:run-index}.
\item \textbf{MEC mechanism controls (Appendix~\ref{app:new-mec-queries}).}
Node semantics, including PnP's command markers and fixed HOME arm, are in Appendix~\ref{app:benchmark-gallery-r32}. All targets use the training mapping's physical origin $\pm0.75$/$\pm2$ scale units. The shared decoder fits 2400 independent ID images. Thirty-two contexts and two commands per target give 384/320/384 CF queries per executable seed for PnP/Balance/Gears. Seeds 300/301 execute for PnP/Gears; only 300 for Balance. Population tests use 2048 model and simulator samples per command. Full encoder-training distributions and calibration-inclusive totals are unspecified.
\end{itemize}
Scene illustrations do not fill unspecified task-table configurations.

\paragraph{Configuration distinctions and interpretation.}
Appendix~\ref{app:unified-evidence} disaggregates learned-$T$ MEC controls; the uncalibrated physical control does not estimate paired calibration gains. Aggregate point estimates do not recover per-run variability or common query sets. Missing Spring/Seesaw/CausalWorld configurations cannot be supplied by Figure~\ref{fig:benchmark-scenes-r32} or other controls. Table~\ref{tab:ood-benchmark-matrix} retains reported continuous ID/amplitude labels and otherwise A/B groups; named control distributions are in Appendix~\ref{app:ood-definitions-r30}.

\subsection{Causal terminology}\label{app:causal-glossary}
In $G=(V,E)$, edge $i\to j$ denotes direct dependence in $j$'s structural equation; $E$ is distinct from encoder $E_\theta$. A DAG is acyclic. Parents are $\Pa_G(j)=\{i:i\to j\in E\}$; descendants $\De_G(i)$ are nodes reachable from $i$, excluding $i$ itself. An effect along $1\to2\to3$ does not require edge $1\to3$.

For disjoint sets $A,B,C$, $s_A\perp\!\!\!\perp s_B\mid s_C$ denotes conditional independence. Markovness links graphical $d$-separation to independence; faithfulness excludes additional independences from exact cancellation~\citep{colombo2014order}. Zero correlation is insufficient. A skeleton omits directions, a PDAG permits unoriented edges, and a CPDAG represents an observational equivalence class~\citep{hauser2012characterization}; arbitrary finite-sample completion does not identify that class.

$\doop(s_j=u)$ replaces a mechanism, unlike observational conditioning. Here $u$ is a verified attained value, not unchecked $u_{\rm cmd}$; calibration changes the visual mapping rather than the environment. Causal sufficiency, independent disturbances and absence of selection bias are assumptions, not consequences of encoding. Labels anchor node identities. Fitted residuals approximate, rather than reveal, true noise (Appendix~\ref{app:model-objects}); preserving them gives factual-conditioned queries, while resampling gives population queries (Appendix~\ref{app:query-protocol}). Hidden single-frame information can still preclude individual identification. Frozen coordinate and metric definitions appear in Appendix~\ref{app:metrics-r32}; acquisition budgets count returned records, not optimization steps.

\subsection{Full model notation and fitted residuals}\label{app:model-objects}
The true SCM comprises the graph, structural equations, and exogenous distribution defined in the main text:
\[
 \mathcal S^\star=(G^\star,\{f_j^\star\}_{j=1}^K,P_\epsilon),
 \qquad G^\star=(V,\mathcal E^\star).
\]
The learned objects can be written as
\[
 \widehat{\mathcal M}
 =\big(E_\theta,T_{\widehat\phi},\widehat G,
       \{\widehat f_j\}_{j=1}^{K},D_\psi\big),\qquad
 h_j=\widehat f_j(h_{\Pa_{\widehat G}(j)})+\eta_j.
\]
$E_\theta,D_\psi$ are the encoder and decoder, $T_{\widehat\phi}$ the fitted calibration, $\widehat G$ the estimated graph, and $\widehat f_j$ the local predictive mechanisms. The fitted residual $\eta_j$ is not recovered true exogenous noise. Observational, interventional, and factual images enter this model through the same frozen interface. The main method and Appendix~\ref{app:query-protocol} detail its numerical approximations and query procedure.

\subsection{Evaluation notation and scene-image provenance}\label{app:metrics-r32}
Graph scores are $\mathrm{SHD}=N_{\rm add}+N_{\rm del}+N_{\rm rev}$, where additions, deletions, and reversals transform the estimate into the true graph and a reversal counts once, and $\mathrm{AUPRC}=\int_0^1\operatorname{Prec}(\rho)\,\mathrm d\rho$, where $\rho$ is recall and $\operatorname{Prec}$ precision. AUPRC ranks edge scores; it is distinct from the completion fractions in Table~\ref{tab:pilot}.

For MEC partial-graph controls, each unordered pair $i<j$ contributes at most one error: absent/extraneous adjacency, reverse direction, an unoriented true edge, or bidirectional conflict. For an adjacency shared by both graphs, only the correct single arrow scores zero; the total is bounded by $K(K-1)/2$. Wrong directions and unresolved edges are also reported separately; SHD is not a graph-validity or execution gate. System-table scoring scope is given in Appendix~\ref{app:unified-evidence}. Response correlation is neither a total-effect estimate nor an edge to every affected descendant; its decision rules appear in Appendix~\ref{app:algorithm}.

\paragraph{State means, images, and normalization.}
Table~\ref{tab:ood-benchmark-matrix} evaluates the model mean $\widehat\mu_k(j,u)$ against the simulator's interventional mean $\mu_k(j,u)$ for each non-target node $k\ne j$ under $a=(j,u)$, with per-node scales fixed from training data before squaring errors. The target is excluded; low error evaluates prediction, not unique recovery of the true mechanism. This is distinct from factual-conditioned $\widehat x^{\rm cf}(x,a)$ and next-frame prediction. Scene distributions are qualified in Appendix~\ref{app:ood-definitions-r30}.

The ratio $r=E_{\rm A}/E_{\rm B}$ uses rounded error summaries, not a mean of per-run ratios or a significance test. For $E_{\rm B}\ge E_{\rm A}>0$, higher $r$ means less relative growth but may also reflect worse first-group error; absolute errors are required. A zero denominator gives no finite ratio. A/B mean ID/amplitude shift only where that shift is defined.

\paragraph{Notation and fixed quantities.}
$q_\theta(z\mid x)$ is the encoding posterior, while $q_j(u)$ is a model node setting; $U_t$ is an unoriented-edge set, $\mathcal U_j$ a feasible value range, and $\mathcal A_t\subseteq\mathcal J$ the available targets. The coefficient vector $a_j$ differs from operation $a$. Training labels determine $\mu_j^s,\sigma_j^s$, and frozen-$T$ training observations determine $\mu_j^c,\sigma_j^c$. The superscripted $\mu_j^s,\mu_j^c$ are fixed normalization statistics, distinct from the command-dependent interventional mean $\mu_k(j,u)$. Standard deviations use stabilization $10^{-8}$ and are not re-estimated on test data. Fitted calibration parameters $\widehat\phi$ and residuals $\eta_j$ are not true environmental parameters/noises. Equation~\eqref{eq:short-cal-loss} averages permitted training records, with $\lambda$ balancing target error and observational preservation.

\subsection{Environment scenes and generating-graph provenance}\label{app:benchmark-gallery-r32}
Figure~\ref{fig:benchmark-scenes-r32} uses the first observational state at seed 20260924 for the three MEC scenes, rendered without augmentation or prediction. Display sizes are $512\times768$ for PnP and $512\times512$ for Balance/Gears, without changing experiment inputs. Graphs come from environment adjacency matrices (only node positions are rearranged), not discovery from robot contacts; the arm retains its fixed HOME pose.

\begin{description}[leftmargin=1.7cm,style=nextline]
\item[PnP-MEC] $\mathrm{arm\_x}\to\mathrm{obj\_x}\leftarrow\mathrm{grasp}$, $\mathrm{grasp}\to\mathrm{obj\_lift}$, $\mathrm{obj\_x}\to\mathrm{led}\leftarrow\mathrm{target\_x}$. In the figure, arm x and grasp denote persistent markers of commanded position and grasp opening, not the live end-effector position of the arm fixed at HOME. Object x and lift denote the object's horizontal and vertical states, target x denotes target position, and lamp denotes indicator state.
\item[Balance-MEC] $\mathrm{load\_delta}\to\mathrm{tilt}\to\mathrm{ball\_x}\to\mathrm{led}\leftarrow\mathrm{bin\_x}$, corresponding to load difference, beam tilt, ball position, indicator state, and collection-bin position.
\item[Gears-MEC] $\mathrm{drive}\to\mathrm{g1}\to\mathrm{g2}\to\mathrm{ptr}\to\mathrm{led}\leftarrow\mathrm{target}$, corresponding to drive input, the angular states of two gear stages, pointer position, indicator state, and target-dial position.
\end{description}
These are the controlled MEC implementations in Appendices~\ref{app:multibench-protocol} and~\ref{app:new-mec-queries}, not substitute graphs for other task-table environments.

The lower row uses original $256\times256$ PNGs Spring:1004 and Seesaw:1001, the first entries in fixed official test lists, without error- or appearance-based selection. The CausalWorld illustration uses native Pushing in \texttt{causal-world==1.2}: seed 20260925, one default reset, original cameras/geometry, and CPU PyBullet DIRECT/TinyRenderer at $320\times240$, without action search. These are appearance illustrations, not paired GT or predictions; the Pushing illustration does not specify the main-table task.

\subsection{Scene-specific test distributions}\label{app:ood-definitions-r30}
ID/OOD denotes a relation between training and test generators, not a score. Independent contexts from the training generator are ID; changes in commands, initial states, appearance, or mechanisms define different shifts. Models, normalization, and evaluation seeds are frozen. The following ranges belong to their specified diagnostic cohorts, not every aggregate column.

\begin{description}[leftmargin=2cm,style=nextline]
\item[Domino-SCM] The independent pixel-backbone controls use training-anchor mean $\pm0.75$ standard deviations for ID and $\pm2$ for amplitude tests. A separate protocol uses $\pm1.15/\pm1.75$. Gaussian training commands have full support, so these are amplitude-distribution changes rather than strict out-of-support extrapolation.
\item[Other MEC] PnP/Balance/Gears independent baseline comparisons use training-concept mean $\pm0.75$ standard deviations (ID) and $\pm2$ (amplitude shift). These ranges do not define every main-table run.
\item[Physical Domino] Four final-state events are binary: command 0 fixes the target upright and 1 tilts/releases it. Held-out contexts from the same initial-state generator are paired ID tests; binary commands alone do not define amplitude OOD.
\item[Spring, Seesaw, CausalWorld] Table~\ref{tab:ood-benchmark-matrix} reports two test datasets, A and B, not method or supervision groups. Their scores do not identify an amplitude, initial-state, appearance, or mechanism shift. Other CausalWorld task protocols cannot supply this column's configuration or establish cross-domain robustness.
\end{description}

The $\dagger$ columns are reported conditional A/B summaries without an identified OOD type or established common retained-query population; the reported labels on continuous columns retain the scope in Appendix~\ref{app:run-index}. Identifying a shift additionally requires the actual training/test command or context distributions and changed dimensions. No missing ranges are supplied from a different cohort.

\subsection{Experimental settings and complete disaggregated results}\label{app:reported-protocol}\label{app:cf-details}\label{sec:rigidbody-native-cf}
This appendix provides training data, model selection, sample denominators, and disaggregated errors for the identity-calibration controls and rigid-body queries. Each configuration uses the training and query protocol described below.

\paragraph{Shared coordinate settings.}
The uncalibrated controls in this appendix use encoder readouts directly and are reported separately from the learned-calibration configurations in the main tables. SCM controls share the frozen visual model and data pool for each seed. Rigid-body controls represent four toppling events and propagate intervention outcomes through conditional means. These results analyze graph propagation, not the gain from nonlinear calibration.

\paragraph{SCM dominoes.}
The seven nodes are a pusher, four dominoes, a bell, and a lamp; images are generated by kinematic rendering driven by structural equations. RGB64 and RGB128 each use 10 training seeds and 6000 observational training records, with 350 epochs of visual-model training. The subsequent interface collects 14 batches of 64 prespecified ID intervention records; the same record pool fits mechanisms in the corresponding coordinates. Each seed has 64 same-noise paired counterfactual queries per target. RGB64 has 8 downstream-executable seeds, including 1 with a complete but incorrect graph; RGB128 has 9. The remaining seeds retain unoriented edges. Means and standard deviations are computed over valid seeds; both resolutions have a planned denominator of 10. Detailed results appear in Appendix~\ref{app:unified-evidence}.

\paragraph{Paired rigid-body conditional-mean control.}
The settings and disaggregated statistics below describe runs with seeds 42, 314, and 2718. They are reported separately from both the uncalibrated physical aggregate in Table~\ref{tab:run-index} and the main-table learned-calibration summary, and do not supply seed denominators, standard deviations, or budgets for either.
Four nodes indicate whether each domino ultimately falls; a deviation from vertical of at least $10^\circ$ is labeled 1. $F_j=0$ is implemented by fixing the domino upright, and $F_j=1$ by initially tilting and then releasing it. The generator checks the target's actual settled value. The robot arm, bell, and lamp are fixed scene components. Factual and counterfactual scenes share the physical initial state and other parameters; inputs are RGB128 images and intervention commands. The 32 test initial states, four targets, and two settings form 256 paired queries.
This evaluation directly concerns same-initial-state replays under fix/release commands. Checking target attainment alone does not establish modularity of a final-state SCM: physical actions may still affect other variables through paths absent from the binary-event representation. It therefore does not establish that the ideal hard-intervention assumption in Eq.~\eqref{eq:short-do} holds in the physical system.
Our method uses 512 concept-labeled training images, 128 ID validation images, and 512 independent observational images for skeleton learning. The two complete runs each execute two rounds of 128 intervention records. The visual model is trained for 100 epochs; the encoder is then frozen while the decoder is post-fitted for 100 epochs on the same training images, reconstructing them from the frozen encoder's posterior means. Candidates are selected only by ID validation. The three prespecified seeds are 42, 314, and 2718; the last stops before downstream queries because of partial-graph ambiguity.

\paragraph{CausalVAE.}
We retain the authors' four-concept model with four dimensions per concept, training losses, DAG module, and latent-intervention path, adapting RGB resolution and label scales; the graph is initialized at zero. The native intervention operator uses different branches for the first two and last two concept slots, and this table reports the task adaptation of that fixed implementation. Inference uses deterministic latent means and takes only the factual image and permitted command as inputs. The 1280-image group contains 512 anchoring, 512 observational, and 256 intervention images, with all concept labels supplied to the model. It therefore uses more labels than our method; both are scored on the same 32-initial-state query set. We report both the original decoder and an additionally ID-refitted decoder; counterfactual comparisons use the stronger original decoder.

\begin{table}[htbp]
\centering\small\setlength{\tabcolsep}{3pt}
\caption{Disaggregated errors for the paired rigid-body dominoes control, in units of $10^{-3}$. Full-image results are means $\pm$ sample standard deviations over common completed seeds 42 and 314; they do not describe the uncalibrated physical aggregate in Table~\ref{tab:run-index} or the main-table learned-calibration summary.}
\label{tab:rigidbody-causalvae}
\begin{tabularx}{\linewidth}{Xrrr}
\toprule
Configuration & Full-image MSE & Keep upright & Make fall\\
\midrule
Our method (conditional-mean propagation) & \datum{appendix-reported-protocol-tex-c0001} & \datum{appendix-reported-protocol-tex-c0002} & \datum{appendix-reported-protocol-tex-c0003}\\
Factual-residual propagation control & \datum{appendix-reported-protocol-tex-c0004} & \datum{appendix-reported-protocol-tex-c0005} & \datum{appendix-reported-protocol-tex-c0006}\\
Target-only latent editing & \datum{appendix-reported-protocol-tex-c0007} & \datum{appendix-reported-protocol-tex-c0008} & \datum{appendix-reported-protocol-tex-c0009}\\
CausalVAE, 512 images, original decoder & \datum{appendix-reported-protocol-tex-c0010} & \datum{appendix-reported-protocol-tex-c0011} & \datum{appendix-reported-protocol-tex-c0012}\\
CausalVAE, 1280 images, original decoder & \datum{appendix-reported-protocol-tex-c0013} & \datum{appendix-reported-protocol-tex-c0014} & \datum{appendix-reported-protocol-tex-c0015}\\
CausalVAE, 512 images, ID refit & \datum{appendix-reported-protocol-tex-c0016} & \datum{appendix-reported-protocol-tex-c0017} & \datum{appendix-reported-protocol-tex-c0018}\\
CausalVAE, 1280 images, ID refit & \datum{appendix-reported-protocol-tex-c0019} & \datum{appendix-reported-protocol-tex-c0020} & \datum{appendix-reported-protocol-tex-c0021}\\
\bottomrule
\end{tabularx}
\end{table}

\paragraph{Interpreting the disaggregated differences.}
Conditional-mean propagation directly uses local regression means for descendants without adding factual residuals back. Its advantage mainly comes from cascading toppling. For keep-upright queries, CausalVAE's original decoder outperforms this branch; retaining factual residuals mitigates the difference. The four binary events may omit hidden initial triggers that determine a particular fall, so pixel error and preservation of individual outcomes are reported separately. Decoder refitting improves ID reconstruction but raises CausalVAE's counterfactual error, showing that the two metrics measure different abilities.

\begin{figure}[htbp]
\centering\setlength{\tabcolsep}{1pt}
\begin{tabular}{cccc}
\small Factual & \small GT & \small Our method & \small CausalVAE\\
\multicolumn{4}{c}{\small Keep the first domino upright: $F_1=0$, first two fixed test initial states}\\
\includegraphics[width=.24\linewidth]{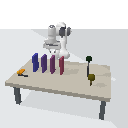}&\includegraphics[width=.24\linewidth]{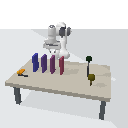}&\includegraphics[width=.24\linewidth]{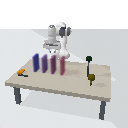}&\includegraphics[width=.24\linewidth]{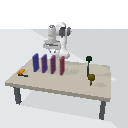}\\
\includegraphics[width=.24\linewidth]{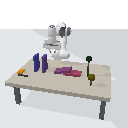}&\includegraphics[width=.24\linewidth]{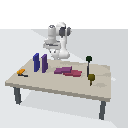}&\includegraphics[width=.24\linewidth]{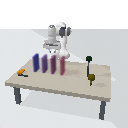}&\includegraphics[width=.24\linewidth]{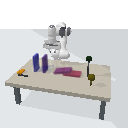}\\
\end{tabular}
\caption{The same seed 42 and factual initial states as in the main text, with the command changed to $F_1=0$. The second row shows conditional-mean propagation changing a descendant that fell independently in the factual scene. Predicted pixels retain the original output.}
\label{fig:cf-full-demo}
\end{figure}

\paragraph{Intervention-distribution evaluation.}
ID commands are $\pm1.15$, and amplitude-shift commands are $\pm1.75$. Model and GT samples are generated separately, and their mean images and means in frozen original-encoder coordinates are compared. Original-coordinate error is visual-readout error, not independent physical-state error. Seeds that fail downstream are not filled with zero error. The two resolutions in Appendix~\ref{app:unified-evidence} use different valid-seed sets and are not inserted into the main-text NMSE summary table.

\paragraph{Inputs to graph-backend comparisons.}
PC receives the shared encoder's observational matrix $Z_{\rm obs}\in\mathbb R^{N\times K}$; intervention-aware graph learners additionally receive the corresponding records and target mask $M\in\{0,1\}^{N\times K}$. $N$ is the number of records and $K$ the number of nodes. An entry of 1 in $M$ indicates that the corresponding node was directly intervened on in that record. The same encoder, normalization statistics, and data permissions localize differences to the graph-learning backend. This comparison is interpreted separately from the end-to-end comparison with CausalVAE, which trains its own representation from images.

\paragraph{ILCM adaptation to the physical scene in Figure~\ref{fig:rigidbody-causalvae}B.}\label{app:ilcm-physical-demo}
ILCM uses the same physical environment and RGB128 rendering settings as the paired GT in Figure~\ref{fig:rigidbody-causalvae}B, training a separate four-dimensional continuous-latent model. Its official image network does not directly support resolution 128, so one native residual scaling layer is added to both the encoder and decoder of the 64-resolution branch. The implicit mechanisms, Gaussian interface, and training objective retain the original implementation. Training uses seed 42, 1280 training image pairs, and 256 independent ID validation pairs. Each pair shares an initial context, and pairs with no post-operation change are retained. Training consists of 20 pretraining epochs and 100 mechanism-training epochs, with Adam learning rate $10^{-4}$ and batch size 32; the checkpoint is selected by the ID validation objective over the full mechanism stage. Training accesses neither intervention-target labels nor the true graph. After training, only training-set event labels are used to match the latent permutation, and latent means for the two event classes define command endpoints.

For queries, the factual image is first encoded into noise coordinates. The native conditional solution function converts the target value to target noise while retaining all other noise coordinates; ILCM's own decoder then generates the image. Generation calls neither our mechanisms nor our decoder and does not take GT as input. Figure~\ref{fig:rigidbody-causalvae}B fixes seed 42, context 59000000 (test-initial-state index 0), and the command to topple the first domino, without selecting a case by image quality. The paired data require 3072 simulation calls, a different training budget from the other models; this run's mean over 256 queries appears in the corresponding main-table cell. The four binary toppling events and physical fix/release commands do not fully satisfy the identification assumptions of continuous ILCM, so algebraic intervention checks do not establish identification of causal semantics.

This adaptation produces blurry outputs and weak command responses. Across 32 fixed test contexts, changing the first domino's command from 0 to 1 gives an RGB mean absolute change of $2.28\times10^{-5}$ in ILCM predictions, compared with $9.91\times10^{-3}$ in paired physical GT. This configuration is not an adequately trained counterfactual baseline. Figure~\ref{fig:rigidbody-causalvae}B shows its actual outputs; the conclusion is limited to this single-seed adaptation and does not characterize ILCM's general capabilities. The effects of representation compression, optimization convergence, and physical-task adaptation cannot yet be separated from these results.

\FloatBarrier
\section{Algorithm, Query Implementation, and Mechanism Approximation}
\subsection{Complete procedure and execution conditions}
\label{app:algorithm}
Algorithm 1 fits the visual model and calibration, freezes that interface for graph acquisition, then fits mechanisms. The main procedure learns $T$ by Eq.~\eqref{eq:short-cal-loss}; omitting numerical calibration defines a separate control. All inference parameters are frozen, and budgets count returned environment records.

\begin{center}
\fbox{\begin{minipage}{.94\linewidth}
\textbf{Algorithm 1: Visual-representation calibration and active graph discovery}\par\smallskip
\textbf{Input:} Training images and concept labels $\mathcal D_{\rm obs}$; intervenable targets $\mathcal J$ and a frozen ID command distribution; environment-call budget $B=B_{\rm cal}+B_{\rm graph}$; training, calibration, independence-test, and orientation configurations.\par
\textbf{Output:} The upstream procedure returns frozen representation $h$, estimated graph $\widehat G$ with unoriented-edge markers, and orientation evidence. When execution conditions are met, downstream processing fits shared local mechanisms and provides counterfactual decoding and outcome-computation interfaces based on $\hat f$.
\begin{enumerate}[leftmargin=1.8em,itemsep=.2em]
\item Learn $E_\theta,D_\psi$ from training images and concept labels, then freeze them. Do not access test states, the true graph, or test results.
\item Call the environment in a prespecified round-robin order to collect $B_{\rm cal}$ records $(j,u,x^{\rm int})$. A batch of $n$ records counts as $n$ calls, including repeated targets. Optimization steps that replay accumulated records do not add environment calls.
\item Fit $T_\phi$ using Eq.~\eqref{eq:short-cal-loss}. Freeze $T$ after the prespecified training rounds; compute $\mu^c,\sigma^c$ only from training observations and construct the shared interface $h$ in Eq.~\eqref{eq:short-coordinates}. The uncalibrated control retains encoder readouts unchanged and is reported separately.
\item Run PC-style adjacency search on $h(\mathcal D_{\rm obs})$, recording separating sets, colliders, and Meek propagation to obtain the initial partial graph $\widehat G_0$. If the structure is inconsistent, return a failure record rather than replacing it with the true graph.
\item Initialize the graph-discovery call count $b=0$ and an empty intervention-record set. Retain the observational orientations and initial unoriented adjacencies of $\widehat G_0$.
\item While unoriented edges remain and $b<B_{\rm graph}$:
  \begin{enumerate}[label=(\alph*),leftmargin=1.8em,itemsep=.1em]
  \item Select a target covering current unoriented edges using Eq.~\eqref{eq:short-policy}. The RGB sequential-update cohort breaks ties by smallest node index and caps each target at six blocks (Appendix~\ref{app:standard-modules}). Do not access the true graph.
  \item Sample commands from the frozen ID distribution, execute the environmental operations, and obtain settled-state images. The last batch must not exceed the remaining budget. After obtaining $n$ records, set $b\leftarrow b+n$.
  \item Pass the new images through the same frozen $h$. Append the encoded responses and single-target intervention masks, which identify the directly operated node in each record, to the accumulated evidence. Do not update the encoder, calibration, or normalization statistics online.
  \item Reconstruct the current partial graph from $\widehat G_0$ and accumulated evidence. Test only the initially unoriented candidate edges in the CPDAG, preserving observationally compelled directions. Propose orientations using the response decision table in this appendix.
  \item Tentatively add each candidate orientation to a graph copy. Reject it if it creates an opposite-direction conflict, a cycle, or an inconsistent DAG extension. If the checks pass, apply Meek orientation propagation until no further direction follows, then commit that update. Record reasons and sources for acceptance, rejection, and propagation.
  \end{enumerate}
\item Stop when no unoriented edges remain, target rules permit no further query, or the budget is exhausted. Return the final partial graph and all unresolved directions; do not complete edges by node index.
\item Fit the local mechanisms described in the main text only if the graph is valid, has zero unoriented edges, and is acyclic. Otherwise return a downstream-blocked status. Do not treat an executable DAG completion as a graph identified by the main method.
\end{enumerate}
\end{minipage}}
\end{center}

\paragraph{Full-DAG gate and blocking unit.}
Here ``sufficiently oriented'' strictly means zero unoriented edges in the whole graph, not just in a query-relevant subgraph. Appendix~\ref{app:interface-r39} checks validity and zero unoriented edges for each seed--configuration, then mechanism construction rejects cycles through topological sorting. A failed configuration has no predictor: all its planned CF and population queries are blocked, with no target-specific salvage from the partial graph. This is a seed--configuration gate, not a query-by-query answerability test; a seed execution fraction cannot substitute for query-level answerability. The prespecified consistent-extension control in Appendix~\ref{app:static-factor} is separate.

\paragraph{Using the trained model.}
Factual-image queries preserve estimated residuals and decode through Eq.~\eqref{eq:short-decode}; command-only outcomes resample training roots/residuals through Eq.~\eqref{eq:short-rollout}. Both use the same fitted $\hat f$, without true test states (Appendix~\ref{app:query-protocol}). Static graph evaluation is not temporal evolution; temporal prediction would require a trajectory interface, cross-time graph, and transition mechanisms.

\paragraph{Implementation scope and failure reporting.}
Marginal response correlations on randomized single-target records are unavailable for $n\leq30$. The rule assumes valid interventions and reliable candidate adjacencies; weak one-sided evidence can reverse an edge and cannot generally distinguish direct from ancestral effects. Fixed-order sequential closure is reproducible, not order-invariant. Thresholds, skeleton errors, visual mixing, and finite samples may misorient edges; Meek propagation cannot correct statistically wrong premises.

\paragraph{Direct-response criterion.}
Let $c_i=|\widehat{\Corr}(h_i,h_j;\mathcal I_i)|$, with $c_j$ defined symmetrically. Each side requires at least 31 single-target records with binary intervention masks to be available. With $\tau=0.15,\delta=0.05$, the decisions are as follows; a proposed direction is accepted only after graph-consistency checks.
\begin{center}\small
\begin{tabularx}{\linewidth}{L{.19\linewidth}Xl}
\toprule
Evidence & Condition & Proposal\\\midrule
Both sides available & $c_i>\tau,\ c_j<\tau,\ c_i-c_j\ge\delta$ & $i\to j$\\
Both sides available & Swap $i,j$ in the preceding row & $j\to i$\\
Both sides available & Otherwise (including both strong, both weak, or an insufficient gap) & Unoriented\\
Only side $i$ & $|c_i-\tau|<\delta$ & Unoriented\\
Only side $i$ & Outside the preceding interval and $c_i>\tau$ & $i\to j$\\
Only side $i$ & Outside the preceding interval and $c_i\le\tau$ & $j\to i$\\
Only side $j$ & Symmetric treatment & Symmetric\\
Neither side & No available response score & Unoriented\\\bottomrule
\end{tabularx}
\end{center}
The implementation records non-numeric correlations caused by constant variables as 0 rather than automatically abstaining. When only that side is available, this can trigger the weak-evidence reverse-direction rule. Graph-consistency checks therefore do not guarantee statistically correct orientations.

\paragraph{Calibration parameterization and inversion.}
The seven-node nonlinear calibration fixes $L=5$ and $\kappa=(-2,-1,0,1,2)$, with training parameterization $g_j=1+0.25\tanh(\xi_j)$ and $w_{jl}=0.1\tanh(\omega_{jl})$; biases $b_j$ are unrestricted. Weights may be positive or negative. These ranges give a derivative lower bound of at least 0.25 and a total of 49 learnable scalars. The query loader checks $g_j>\sum_l|w_{jl}|$. If all weights are zero, inversion uses the analytic expression $(y-b_j)/g_j$. Otherwise, with $W_j=\sum_l|w_{jl}|$, elementwise bisection runs for exactly 60 iterations within $[(y-b_j-W_j)/g_j-10^{-6},(y-b_j+W_j)/g_j+10^{-6}]$. The round-trip error threshold on training observations is $10^{-8}$, not a global numerical-accuracy guarantee for every extrapolated input. The fully fitted affine control instead uses weighted least squares with a positive slope lower bound of $10^{-6}$ and no upper bound, a different parameter range from the bounded affine control.

\paragraph{Budgets and logs.}
The budget unit is the number of independent records returned by environment calls, not optimization steps or acquisition rounds. The number and training cost of observational samples used to train the encoder are reported separately; calibration and graph discovery together must not exceed $B$. Implementation logs save at least sample seeds, commands and realized values, stage, cumulative call count, graph states before and after updates, effect scores, rejection reasons, and hashes of frozen parameters, code, and data.

\subsection{Confidence orientation backend and empty-graph query operator}\label{app:implementation-rules}
The frozen criterion, held-out run records, and query implementation specify two particular controls: eight-seed Domino-MEC confidence orientation and the three-scene empty-graph control in Appendix~\ref{app:new-mec-queries}. These rules do not automatically apply to other aggregate cells with the same method names.

\paragraph{Response statistic and resampling.}\label{app:confidence-backend}
For an unoriented adjacency $i-j$ in the initial observational CPDAG, retain only records whose mask has exactly one target. Within interventions on $i$ and $j$, respectively, compute the dimensionless absolute Pearson correlations $\widehat d_i=|\widehat\Corr(z_i,z_j\mid\doop(i))|$ and $\widehat d_j=|\widehat\Corr(z_i,z_j\mid\doop(j))|$. A side is available only with at least 200 records. Resample whole rows within that side with replacement 2000 times, re-centering and recomputing the correlation each time to obtain $D_i^{(b)}$ and $D_j^{(b)}$. Intervals $[L_i,U_i]$ and $[L_j,U_j]$ use the 0.005 and 0.995 quantiles with linear interpolation. Independent deterministic streams serve the two sides. When both exist, $\Delta^{(b)}=D_i^{(b)}-D_j^{(b)}$ gives $[L_\Delta,U_\Delta]$ by the same quantiles; this is not a difference between paired individuals across interventions.

Nonfinite encodings are rejected; fewer than 200 records make a side unavailable. Bootstrap correlations with denominators at most $10^{-12}$, point correlations with denominators below $10^{-12}$, and nonfinite correlations are set to zero. Constant inputs cannot alone trigger strong-signal orientation, although the weak side of a two-sided decision may be constant. These gates do not establish a true causal direction.

\paragraph{Calibrated noise ceilings and frozen selection.}
For each calibration edge with both sides available, select the side with the smaller point correlation (side $i$ on a tie) and add its upper confidence endpoint to the multiset $\mathcal V_k$ for that intervention target. Let $\mathcal V=\bigcup_k\mathcal V_k$. The target-specific noise ceiling is
\begin{equation}
 b_k=\min\!\left\{1,\ 0.03+
 \begin{cases}
 \max\{\max\mathcal V_k,Q_{0.95}(\mathcal V)\},&|\mathcal V_k|\ge4,\\
 \max\mathcal V,&|\mathcal V_k|<4.
 \end{cases}\right\}.
 \label{eq:bootstrap-null-bound}
\end{equation}
Here $Q_p$ denotes the empirical $p$-quantile with linear interpolation.
Thus 0.03 is a margin added to a data-derived ceiling, not a zero-effect threshold applied directly to correlation. Under the paper's one-based node indexing, the frozen values are
\begin{equation}
 \begin{aligned}
 b_1=b_6&=0.32561556156709537,\\
 b_5&=0.32477945538327324,\\
 b_2=b_3=b_4=b_7&=0.3211601465259132.
 \end{aligned}
 \label{eq:bootstrap-frozen-bounds}
\end{equation}
Ceiling fitting itself does not read graph truth; an empty $\mathcal V$ raises a calibration error. Criterion selection uses leave-one-seed-out predictions on seeds 10--17, seals predictions, then uses the \emph{calibration-set true graphs} to reject candidates introducing wrong arrows or invalid graphs. Remaining candidates are ranked by maximizing complete graphs, maximizing correct edges, minimizing unresolved edges, then fixed priority; ceilings are fitted to all calibration records. The four prespecified (confidence, added margin, signal excess, directional contrast) tuples are $(0.99,0.03,0.10,0.15)$, $(0.95,0.03,0.10,0.15)$, $(0.95,0.02,0.05,0.10)$, and $(0.90,0.02,0.05,0.10)$; the first was frozen. If no candidate passes, a safe mode abstains everywhere. Inference on held-out seeds 20--27 does not read true graphs. This statistical-criterion calibration is distinct from anchor calibration of the numerical map $T$.

\paragraph{Direction decision and commitment.}
With only side $i$ available, propose $i\to j$ only if $L_i>b_i+0.10$; side $j$ is symmetric. Weak one-sided evidence never proposes the reverse direction. With both sides available, use
\begin{equation}
 \begin{aligned}
 i\to j:\quad&\widehat d_i>\widehat d_j,\quad L_i>b_i+0.10,\quad U_j\le b_j,\quad L_\Delta>0.15;\\
 j\to i:\quad&\widehat d_j>\widehat d_i,\quad L_j>b_j+0.10,\quad U_i\le b_i,\quad U_\Delta<-0.15.
 \end{aligned}
 \label{eq:bootstrap-direction-rule}
\end{equation}
With neither side available, equal point estimates, or any failed condition, leave the edge unoriented; an invalid initial graph yields no proposal. Each round recomputes from the frozen initial CPDAG and cumulative records, with \emph{sequential commitment} along its unoriented-edge list. A proposal must preserve acyclicity and then pass Meek closure with a consistent-extension check before the edge and propagated arrows are committed. Failed updates are discarded, and edges already oriented by closure are skipped. Observationally compelled arrows cannot be overwritten. The 0.99 level is nominal for each interval, not a probability that the whole graph is correct.

The fixed-threshold replacement shares held-out seeds, frozen encoders, initial CPDAGs, the target-by-block-index data cache, and acquisition-budget settings, but uses the original $\tau=0.15,\delta=0.05$ rule. Backend-dependent graphs can change subsequent targets, so a shared cache does not imply identical realized action sequences. The 8/8 versus 4/8 result compares final complete recovery, not matched action paths.

This held-out cohort uses 1800 records per block, at most six blocks, and a six-block per-target cap. Acquisition ceases once the graph is valid with no unoriented edges; incomplete runs retain their final states. This runner evaluates graphs only and constructs no downstream query predictor, so 8/8 complete graph recovery is not a query-execution denominator.

\paragraph{Empty-graph population operator and factual branch.}\label{app:nograph-operator}
This paragraph applies to the MEC mechanism controls with learned $T$ in Appendix~\ref{app:new-mec-queries}. Let $w$ denote the mechanism coordinates produced by their frozen visual interface and $v_j(u)$ its training-unit conversion of the command. The empty-graph model is \emph{rebuilt from the same encoded records and intervention masks} with zero adjacency, making every node a root. Each node stores its observational empirical marginal $\widehat P_k^{\rm obs}$, falling back to records not intervening on that node only if the observational pool is empty. Original parent coefficients and residual pools are not retained, and no intercept regression is fitted for roots.

The command-only population query samples non-target nodes independently with replacement and clamps the target:
\begin{equation}
 \widehat P_{\varnothing}(w\mid j,u)
 =\delta_{v_j(u)}(w_j)\prod_{k\ne j}\widehat P_k^{\rm obs}(w_k).
 \label{eq:nograph-population-operator}
\end{equation}
Former non-roots therefore use observational marginals, not their original fitted residuals; whole observational rows are not resampled to preserve joint dependence. Both methods receive the same commands and GT means, each generate 2048 model samples per command, and average after conversion to protocol units. For each command, the same seed initializes a fresh random stream for each method, but graph-dependent topological traversal differs, so noises are not paired coordinatewise.

The factual-image branch does not sample these pools: it copies $w^{\rm fact}$, sets only $w_j=v_j(u)$, retains all other coordinates, and uses the shared decoder. This empty-graph object has no descendants, so its factual operator differs from the population product of marginals. The module ablation that retains fitted mechanisms but disables descendant updates is interpreted separately: identical target-edit outputs would not establish identical population predictors or cohorts. These implementation records bind to the valid diagnostic seeds, 2/2 each for PnP/Gears and 1/2 for Balance. The other Balance seed is excluded for both methods because of unresolved structure; this does not supply operators or query denominators for other main-table aggregates.

\subsection{Visual-learning, coverage-targeting, and local-mechanism formulas}\label{app:standard-modules}
This section specifies the visual-model objective, coverage-targeting rule, and linear mechanisms that support the calibration and query procedure in the main text.

\paragraph{Node-aligned variational autoencoder.}
Encoder $E_\theta$ estimates state-aligned latents $z$, and decoder $D_\psi$ reconstructs images from them. Concept labels $\widetilde\ell_i$, standardized by observational training-label means and standard deviations, anchor node identities:
\begin{equation}
 \mathcal L_{\rm enc}(x_i,\ell_i)=\E_{z\sim q_\theta(\cdot\mid x_i)}
 \|D_\psi(z)-x_i\|_2^2
 +\beta D_{\rm KL}\!\left(q_\theta(z\mid x_i)\,\|\,\mathcal N(\widetilde\ell_i,I)\right).
 \label{eq:short-encoder}
\end{equation}
Here $q_\theta$ is the encoding posterior, $D_{\rm KL}$ the Kullback--Leibler divergence, $\beta$ the regularization weight, and $I$ identity covariance; the two terms constrain reconstruction and label matching. After training, the encoder is frozen and its posterior mean $z=E_\theta(x)$ is used. The jointly trained base decoder is denoted by $D$; with the encoder fixed, it may be post-fitted using ID reconstruction alone to obtain $D'$. Calibration updates only $T$, each control shares the same $D$ or $D'$, and all are frozen during queries.

\paragraph{Target selection by unoriented-edge coverage.}
Let $U_t$ be the set of unoriented node pairs at round $t$, and $\mathcal A_t\subseteq\mathcal J$ the intervenable targets with remaining acquisition allowance. The coverage heuristic selects a target adjacent to the most unoriented edges:
\begin{equation}
 j_t\in\arg\max_{j\in\mathcal A_t}
 \sum_{\{p,q\}\in U_t}\mathbf 1[j\in\{p,q\}].
 \label{eq:short-policy}
\end{equation}
Here $t$ is the acquisition round, $\mathbf 1[\cdot]$ is the indicator, and $\arg\max$ returns maximal-coverage targets. The RGB sequential-update cohort (Appendix~\ref{app:interface-r39}) breaks ties by smallest node index. Its $\mathcal A_t$ contains all targets with fewer than six acquired blocks; each block has 1800 records, at most six blocks are acquired, and an empty set stops acquisition. Repeated targets are allowed with no additional smaller cap; a six-block per-target cap cannot bind before the six-block total horizon. This cohort uses no incident-node fallback and does not relax the acquisition cap. Values come from a predetermined ID command distribution without true-graph or test access, and new images pass through frozen $h$. Coverage is a heuristic, not a proven budget-optimal policy.

\paragraph{Linear local mechanisms and ridge regression.}
Local mechanisms are fitted in the same $h$ coordinates as $q_j(u)$ and executed only for valid, acyclic estimated graphs with zero unoriented edges (Appendix~\ref{app:algorithm}). The linear parameterization with an intercept is
\begin{equation}
 h_j=\widehat f_j(h_{\Pa_{\widehat G}(j)})+\eta_j,\qquad
 \widehat f_j(v)=c_j+a_j^\top v.
 \label{eq:short-mechanism}
\end{equation}
Here $v$ contains parent readouts, $c_j$ is an intercept, $a_j$ a coefficient vector, $\top$ denotes transpose, and $\eta_j$ is an additive residual. Ridge regression estimates $c_j,a_j$, excluding records that directly intervene on $j$ because its mechanism has been replaced. Root observations and fitted residuals form sample pools. Linear mechanisms are predictive approximations in calibrated coordinates, not consequences of invertibility. Nonlinear $T$ can change local relationships and residual--parent dependence, so readouts, propagated outcomes, and counterfactuals are evaluated separately (Appendix~\ref{app:mechanism-compatibility}).

\subsection{Implementation of distribution prediction and individual counterfactuals}\label{app:query-protocol}
Both entry points share frozen $E_\theta,T,\widehat G,\widehat f_j,D_\psi$ and coordinate statistics. They differ in whether a factual context is held fixed. Shared coordinates do not guarantee equally accurate observational/interventional readouts or preserve a linear additive mechanism family; cross-source and image-readout tests are given in Appendices~\ref{app:query-evidence-r38} and~\ref{app:interface-r39}.

\paragraph{Execution conditions and scoring sets.}
The whole graph must be valid, acyclic and fully oriented; otherwise all queries for that seed/configuration are blocked. Paired ablations retain common valid seeds. Executable counts in Appendices~\ref{app:calibration-chain-r30} and~\ref{app:interface-r39} remain cohort-specific under Appendix~\ref{app:comparison-scope}. The consistent-extension control in Appendix~\ref{app:static-factor} does not alter this gate.

\subsubsection{Given a factual image: individual counterfactual decoding}
Given factual RGB image $x$, target $j$ and verified value $u$, compute $z^{\rm fact}=E_\theta(x)$, $h^{\rm fact}=h(x)$ and the frozen target mapping $q_j(u)$ through $\alpha_j$. For each non-root node, estimate its factual residual:
\begin{equation}
 \eta_k^{\rm fact}=h_k^{\rm fact}
 -\widehat f_k(h_{\Pa_{\widehat G}(k)}^{\rm fact}).
\end{equation}
This approximates context through fitted residuals, not known true noise. Non-target roots retain factual readouts without resampling; the true test graph is never substituted.

\paragraph{Executing the intervention and preserving unrelated variables.}
Let $\mathcal D_j=\De_{\widehat G}(j)$ denote the target's descendants. Copy the factual vector, set the target value, and update descendants in topological order:
\begin{equation}
 h_k^{\rm cf}=
 \begin{cases}
 q_j(u), & k=j,\\
 \widehat f_k(h_{\Pa_{\widehat G}(k)}^{\rm cf})+\eta_k^{\rm fact},
       & k\in\mathcal D_j,\\
 h_k^{\rm fact}, & k\notin\mathcal D_j\cup\{j\}.
 \end{cases}
 \label{eq:query-cf-detail}
\end{equation}
Only the target is set; descendants propagate through mechanisms and non-descendants retain factual values. This preservation rule depends on the estimated graph and can fail when its edges are wrong.

\paragraph{Returning to decoder coordinates.}
First undo graph-interface standardization, then invert calibration:
\begin{equation}
 z^{\rm cf}=T^{-1}(\sigma^c\odot h^{\rm cf}+\mu^c),
 \qquad \widehat x^{\rm cf}=D_\psi(z^{\rm cf}).
\end{equation}
The decoder receives original $z$ coordinates. The model uses neither GT counterfactual images/outcome states nor a replacement paired-outcome prediction head. Invertibility provides compatibility, not extrapolation accuracy.

\paragraph{Rigid-body conditional-mean branch.}
The rigid-body conditional-mean propagation control omits numerical calibration and recomputes descendants without the factual residuals in Eq.~\eqref{eq:query-cf-detail}. Both propagation variants share encoding, graph, and decoding interfaces; disaggregated results appear in Appendix Table~\ref{tab:rigidbody-causalvae}. Binary events may omit hidden initial triggering information, so conditional-mean propagation can incorrectly change descendants that fell independently in the factual scene.

\subsubsection{Given only a command: intervention-distribution prediction}
\paragraph{Training mechanisms and sample pools.}
Fit ridge regressions with intercepts on the parent sets specified by the graph, excluding records directly intervening on node $k$ when fitting that node. The ridge coefficient for the linear mechanisms is $10^{-3}$, with no penalty on the intercept. Root nodes retain observational encoded samples; non-root nodes retain fitted residuals. Simulation samples with replacement from each node's corresponding empirical pool. This is a model sampling approximation, not evidence that visual residuals equal independent true physical noise.

\paragraph{Generating samples.}
The input is only an operation $a=(j,u)$; the estimand is the mean non-target state of its intervention distribution. With no factual $x$, $\eta^{\rm fact}$ is not reused. Let $M$ denote the prediction sample count, with $h^{(m)}$ generated for each $m=1,\ldots,M$. Let $\widetilde r_k$ be a readout from root node $k$'s training observational pool and $\widetilde\eta_k$ a residual sampled with replacement from its training residual pool. Evaluate nodes in topological order:
\begin{equation}
 \widetilde h_k=
 \begin{cases}
 q_j(u),&k=j,\\
 \widetilde r_k,&k\ne j,\ \Pa_{\widehat G}(k)=\varnothing,\\
 \widehat f_k(\widetilde h_{\Pa_{\widehat G}(k)})+\widetilde\eta_k,
 &k\ne j,\ \Pa_{\widehat G}(k)\ne\varnothing.
 \end{cases}
 \label{eq:short-rollout}
\end{equation}
Repeated samples of $\widetilde h_k$ approximate the settled-state distribution; topological order is not physical time. Counts, seeds, command ranges and configurations are fixed before test evaluation. Metric and shift definitions are in Appendices~\ref{app:metrics-r32} and~\ref{app:ood-definitions-r30}; the empty-graph operator is in Appendix~\ref{app:nograph-operator}.

\paragraph{Order of averaging and decoding.}
State predictions are restored to label units using $\widehat s^{(m)}=\mu^s+\sigma^s\odot(\sigma^c\odot h^{(m)}+\mu^c)$, equivalently $\widehat s_j=\mu_j^s+\sigma_j^s T_j(z_j)$ componentwise. The sample mean for each non-target node is $\widehat\mu_k(j,u)=M^{-1}\sum_{m=1}^M\widehat s_k^{(m)}$, $k\ne j$; scoring units are fixed by the task protocol. Image generation requires inverse calibration back to the original $z$ coordinates, followed by per-sample decoding and averaging:
\begin{equation}
 \overline x_{\rm pred}=\frac1M\sum_{m=1}^M
 D_\psi\!\left(T^{-1}(\sigma^c\odot h^{(m)}+\mu^c)\right).
 \label{eq:query-distribution-image}
\end{equation}
Because the decoder and calibration are generally nonlinear, Eq.~\eqref{eq:query-distribution-image} cannot be replaced by averaging latent vectors before decoding. Means in original encoder coordinates are likewise computed after inverse transformation. GT distribution means come from independent sampling of the permitted environment under the same intervention command; model and GT samples need not be paired one-to-one.

\paragraph{Interpretation of the evaluation.}
Mean-image error evaluates population-average appearance, not individual outcomes. Encoder-coordinate error compares image-set means through the same frozen encoder, not an independent physical-state measurement; both remain separate from state NMSE. This SCM protocol uses $\pm1.15$ ID commands and $\pm1.75$ amplitude tests. Gaussian training support makes this a distribution shift rather than strict support extrapolation; reconstruction and CF examples do not substitute for its OOD evaluation.

\subsection{Query interface for operation values and visual readouts}
\label{app:query-interface-r37}
\begin{figure}[htbp]
\centering
\includegraphics[width=\linewidth]{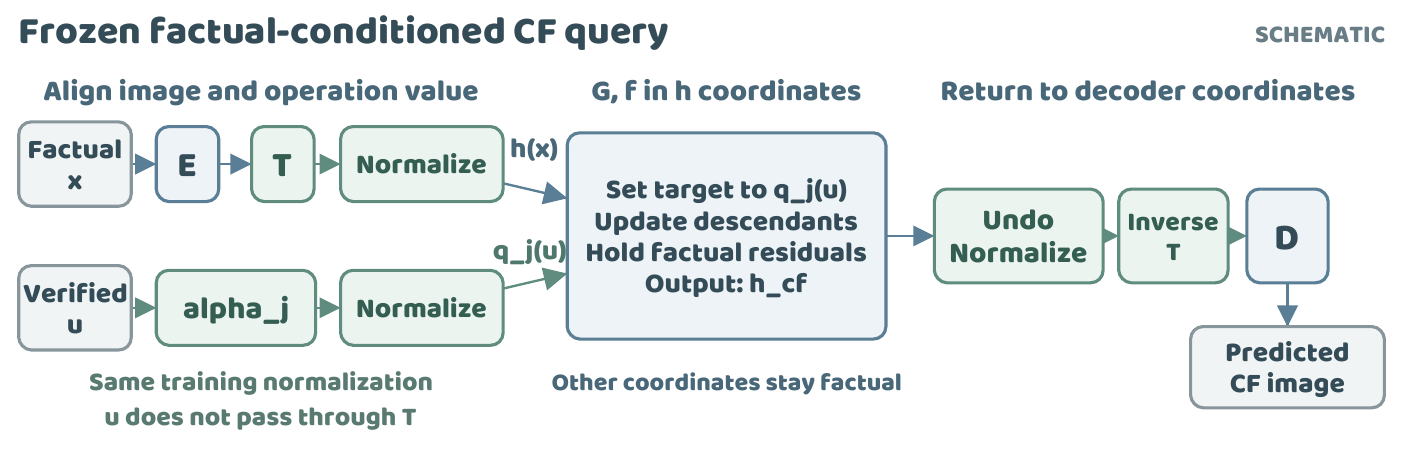}
\caption{Coordinate transformations for a factual-conditioned counterfactual query (schematic). An image passes through $E,T$ to produce calibrated readouts, while the verified value passes through $\alpha_j$. Both paths use the same observational normalization to yield $h(x)$ and $q_j(u)$, respectively; the latter does not pass through $T$ again. After setting the target, retaining factual residuals, and updating descendants, normalization and $T$ are inverted before decoding with $D$. $E,D,G,f$ abbreviate the encoder, decoder, estimated graph, and local mechanisms, respectively. All mappings and statistics are frozen. Population prediction without a factual input is described in Section~\ref{sec:downstream-branches}.}
\label{fig:query-interface-r37}
\end{figure}

Figure~\ref{fig:method-branches} in the main text shows the complete learning procedure. Figure~\ref{fig:query-interface-r37} adds the factual-conditioned counterfactual query in the frozen model, showing how visual readouts and verified intervention values enter the same computational coordinates and how predictions return to the decoder's training coordinates.

\subsection{Calibrated coordinates and mechanism approximation}\label{app:mechanism-compatibility}
Invertible calibration allows queries to return to the decoder's original coordinates; it does not ensure that a specified regression family is preserved by the transformation. To distinguish these properties, temporarily omit affine standardization and suppose the original readout relation is $z_j=f_j(z_{\Pa(j)})+e_j$. The exact transformed relation is
\[
 \widetilde z_j=T_j\!\left[f_j\!\left(T_{\Pa(j)}^{-1}(\widetilde z_{\Pa(j)})\right)+e_j\right].
\]
Here $\widetilde z=T(z)$, and $T_{\Pa(j)}^{-1}$ acts coordinatewise on the parents. If $T_j$ is nonlinear, the increment $T_j(m+e_j)-T_j(m)$ produced by the same $e_j$ generally depends on $m=f_j(z_{\Pa(j)})$. A linear additive model refitted after transformation is therefore an approximation. Invertibility does not imply that residuals remain independent of parent states or that individual noise has been identified.

For positive-affine $T_j(z_j)=g_jz_j+b_j$, exact moment normalization cancels the gain and bias in image readouts $h_j$ (ignoring numerical stabilization). The command setting $q_j(u)$ still depends on them through $\mu_j^c,\sigma_j^c$ in Eq.~\eqref{eq:short-coordinates}; unchanged image coordinates do not imply unchanged queries. Nonlinear calibration can bring distorted visual relations closer to the chosen mechanism family, but this is not guaranteed. Readout, mechanism, and end-to-end query accuracy must be evaluated separately.

\subsection{Orientation and mechanism diagnostics on frozen records}\label{app:frozen-diagnostics}
This analysis uses existing encoders, calibrations, and active-acquisition records for seeds 5, 6, and 7, without training new models. Identity, fully fitted affine, and nonlinear calibration share graph-acquisition records. Prefix-by-prefix replay fixes thresholds and margins and compares ascending, descending, and 30 prespecified random edge orders. Permutations are algorithm-diagnostic conditions, not independent training seeds. We also compare a bilateral-evidence safeguard requiring more than 30 records in both directions and nonzero readout variance. This safeguard is a separate control and does not change the main-table backend.

Mechanism diagnostics use 64 held-out ID factual readouts per seed. For each non-root node, we report frozen-local-mechanism residuals and their relation to parent readouts; bin boundaries for parent readouts are determined only from training observations. Residuals are not errors against ground-truth predictions. Their dependence may arise from visual error, mechanism misspecification, or finite samples and cannot be uniquely attributed to calibration. A pure coordinate diagnostic fixes residual quantiles in original coordinates and computes the standard deviation of the transformed increment as parent states vary; identity and affine transformations provide numerical-zero checks.

\begin{table}[htbp]
\centering\small
\caption{Frozen diagnostics. Graph columns use the final acquisition prefix; all three coordinate systems give the same results, and each cell is unchanged across 32 traversal orders. SHD counts each unoriented true edge as one unresolved relation and therefore differs from the number of wrong arrows. The last column is the SD of transformed increments for fixed original-coordinate residuals under nonlinear calibration, averaged equally over non-root nodes and three training-residual quantiles, in training-label-standardized coordinates. It is a coordinate diagnostic, not prediction error.}
\label{tab:frozen-diagnostics}
\begin{tabular}{ccccc}
\toprule
Seed & Original-rule SHD & Bilateral SHD & Bilateral unoriented edges & Nonlinear increment SD\\
\midrule
\datum{appendix-content-repairs-tex-c0001} & \datum{appendix-content-repairs-tex-c0002} & \datum{appendix-content-repairs-tex-c0003} & \datum{appendix-content-repairs-tex-c0004} & \datum{appendix-content-repairs-tex-c0005}\\
\datum{appendix-content-repairs-tex-c0006} & \datum{appendix-content-repairs-tex-c0007} & \datum{appendix-content-repairs-tex-c0008} & \datum{appendix-content-repairs-tex-c0009} & \datum{appendix-content-repairs-tex-c0010}\\
\datum{appendix-content-repairs-tex-c0011} & \datum{appendix-content-repairs-tex-c0012} & \datum{appendix-content-repairs-tex-c0013} & \datum{appendix-content-repairs-tex-c0014} & \datum{appendix-content-repairs-tex-c0015}\\
\bottomrule
\end{tabular}
\end{table}

Default replay exactly matches the archived graphs. The final prefix contains 1800 records for seed 5 and 3600 each for seeds 6 and 7, so the table is not an equal-budget comparison across seeds. Changing traversal order does not change final graphs on these records; this does not establish order independence for arbitrary graphs or data. Requiring valid bilateral evidence leaves 6 unoriented edges for seed 5, while the other two seeds remain query-executable. The safeguard rejects constant and weak one-sided evidence but reduces orientation coverage under a finite budget, so it is evaluated separately from the main-table algorithm. Degenerate-input tests also show that the original rule can interpret 31 constant one-sided records as a reverse-direction proposal, demonstrating that numerical validity and graph consistency are different checks.

Fixed-residual increment SDs for identity and affine transformations are both zero within tolerance $10^{-10}$; nonlinear results appear in Table~\ref{tab:frozen-diagnostics}. Each seed averages 18 SD values from 6 non-root nodes and 3 training-residual quantiles. Each SD is computed across 64 factual contexts, not across seeds. This gives a concrete example of a changed additive interpretation after transformation, but does not establish that this change is the main cause of downstream error. Local residuals in different coordinates cannot replace outcome evaluation against shared environmental ground truth.

\FloatBarrier
\section{Environmental Interventions and Graph Orientation}
\subsection{Multi-environment structure protocols and ablations}\label{app:multibench-protocol}
The environments share a procedure but differ in state generation, data volumes, and orientation criteria.

\paragraph{Environments and statistical objects.}
Domino-MEC is a seven-node chain, PnP-MEC a six-variable pick-and-place system, Balance-MEC a five-variable balancing system, and Gears-MEC a six-variable transmission system. Explicit SCMs generate responses; renderers provide settled-state frames after actuators return to the fixed HOME pose. Statistical tests exclude action-transition frames. The arm illustrates operation semantics, not contact-dynamic generation of every SCM outcome. Physical contact instead determines the four-event rigid-body domino environment, scored separately under Appendix~\ref{app:reported-protocol}.

Training labels establish node-aligned encodings; graph discovery uses frozen encoded observations, intervention images, and known targets, not the true skeleton. Held-out inference uses the true graph only for offline scoring; criterion selection on calibration seeds uses graph references as specified below. Table~\ref{tab:pilot} reports learned-CPDAG mode, which learns the initial partial graph, excluding true-CPDAG/DAG diagnostics. Structure and counterfactual experiments have separately fixed seed lists; Domino structural holdout and mechanism queries also have separate seeds and protocols.

\paragraph{Domino-MEC: independent holdout and orientation criterion.}
Graph recovery uses held-out seeds 20--27, separate from development seeds 0--9 and criterion-selection/calibration seeds 10--17; these are not numerical-$T$ calibration seeds. Noise ceilings are fitted without graph truth, but criterion selection scores held-out-fold calibration predictions against calibration true graphs. The frozen criterion uses confidence 0.99, 2000 bootstrap resamples, at least 200 records, additive null margin 0.03, signal excess 0.10, and directional contrast 0.15. The margin augments target-specific noise bounds rather than defining a zero-effect interval; one-/two-sided tests and selection rules are given in Appendix~\ref{app:implementation-rules}. Inputs are 6000 observational images and 1800 per intervention block, with 350 encoder epochs and at most six target-selection rounds. All 8 held-out coverage runs complete within two rounds. Rounds are sampling blocks; calls count returned records.

With the same held-out seeds, frozen encoder, initial observational CPDAG, and target/block-index cache, final complete recovery is 8/8 for unoriented-edge coverage (coverage), random targeting (random), and fixed cycling (round-robin), versus 0/8 for observation-only (passive), which retains 6 unoriented edges in every run. Shared cache entries contain identical pixels, but changed graph states can change later targets and realized trajectories. Each round recomputes from the initial CPDAG using cumulative evidence, accepting directions sequentially with Meek closure. This final-state ablation does not establish equal-budget policy superiority. Replacing coverage's confidence criterion with fixed thresholds $\tau=0.15,\delta=0.05$ gives 4/8 complete recovery, retaining errors and unresolved states.

\paragraph{PnP-MEC, Balance-MEC, and Gears-MEC.}
PnP uses independent held-out seeds 200--207; both initial CPDAG and final DAG recovery are 8/8, with zero final errors and unoriented edges. Balance uses 8 development-configuration seeds. After diagnosing extra adjacencies, the skeleton-effect threshold was adjusted from 0.04 to 0.06 while reusing the frozen encoder. Final recovery is correct in 7/8 runs; the remaining run retains an extra adjacency. Balance results therefore analyze development samples rather than provide evidence of independent held-out generalization. Gears uses seeds 0--7, 4200 training images, 5000 observational images, 900 intervention images per block, and 180 encoder-training epochs. Its observational-skeleton threshold is 0.06, with orientation threshold 0.20 and margin 0.08. All 8 learned-graph runs resolve 3 direction ambiguities in one round with zero wrong arrows; the passive controls each retain 3 unoriented edges at termination.

\paragraph{Relation to the shared interface.}
The shared sequence is visual encoding, observational partial graph, target selection, intervention evidence, and graph update; criteria are frozen per environment. Results concern this structure-learning procedure, not an additional non-identity calibrator. Appendix~\ref{app:static-factor} permits protocol-fixed consistent extensions for downstream computation, without counting them as identification success; Appendix~\ref{app:query-protocol}'s strict query entry point does not. Factor-outcome NMAE, pixel-counterfactual MSE, and distribution-prediction error remain separate metrics.

\subsection{On-demand acquisition and frozen-calibration queries}
\label{app:query-evidence-r38}
This control uses a seven-node domino scene rendered as $64\times64$ RGB images and encoder seeds 5, 6, and 7 to evaluate evidence acquisition, readouts, and query outcomes separately. Encoder, decoder, calibration, training statistics, and scoring are frozen before test-data generation, without test-error model selection. Conclusions are limited to this simulation.

\subsubsection{What do additional observations and interventions provide at equal record counts?}
All groups share 6000 observations, 896 calibration records, and a frozen nonlinear mapping. Additional independent observations re-estimate PC at prefixes of 1800 and 3600. Intervention groups start from the same graph, use coverage or fixed round-robin targeting in batches of 1800 up to 3600, and stop at a valid, sufficiently oriented graph. Each reached intervention prefix is paired with equally many new observations. Both sides pay the same calibration cost, but calibration records are excluded from graph-stage evidence: the comparison isolates \emph{the source of graph-stage evidence}, not a wholly intervention-free procedure.

An inconsistent candidate-direction batch is rejected in full under rules and thresholds fixed before acquisition. This control and Appendices~\ref{app:ablation-results} and~\ref{app:ablation-complete-r33} use the same batch-commitment implementation, unlike the sequential commitment in Appendix~\ref{app:algorithm}. The true graph is used only to score fixed trajectories, never for selection, stopping, or graph estimation.

\paragraph{Correspondence among the three acquisition controls.}
Within each training seed, all three controls reuse the same encoder checkpoint, 6000 observations, and nonlinear $T$, with identical file hashes in their frozen manifests. Their initial graphs are also identical field by field: a seven-node chain with 6 unoriented edges. A shared training seed does not imply shared graph-stage samples. Batch $b$ for target $j$ uses sampling seed $N+10007s+101j+b$, where $s\in\{5,6,7\}$ and $j,b$ are zero-based. The equal-record evidence control here uses $N=142000000$; the three-policy control in Appendix~\ref{app:ablation-results} uses $N=2600000$; the equal-total-budget control in Appendix~\ref{app:ablation-complete-r33} uses $N=12600000$. All start with an empty graph-intervention pool and retain the same ID command distribution and orientation thresholds. The graph-record cap is 3600 here and 10800 for the nonlinear configurations in the other two controls.

\begin{table}[htbp]
\centering\small
\caption{Graph-evidence sources under the same frozen visual and calibration interface. Each row compares equal numbers of new records; both sides include the same 896 calibration records and 6000 initial observations. Intervention and additional-observation graphs are scored at matching prefixes; records not acquired after early stopping are not treated as available evidence.}
\label{tab:r38-acquisition}
\begin{tabular}{llrrr}\toprule
Acquisition policy & Seed & New records & Intervention SHD & Added-observation SHD\\\midrule
Coverage targeting & \datum{tables-r38-acquisition-tex-c0001} & \datum{tables-r38-acquisition-tex-c0002} & \datum{tables-r38-acquisition-tex-c0003} & \datum{tables-r38-acquisition-tex-c0004}\\
Fixed round-robin & \datum{tables-r38-acquisition-tex-c0005} & \datum{tables-r38-acquisition-tex-c0006} & \datum{tables-r38-acquisition-tex-c0007} & \datum{tables-r38-acquisition-tex-c0008}\\
Coverage targeting & \datum{tables-r38-acquisition-tex-c0009} & \datum{tables-r38-acquisition-tex-c0010} & \datum{tables-r38-acquisition-tex-c0011} & \datum{tables-r38-acquisition-tex-c0012}\\
Fixed round-robin & \datum{tables-r38-acquisition-tex-c0013} & \datum{tables-r38-acquisition-tex-c0014} & \datum{tables-r38-acquisition-tex-c0015} & \datum{tables-r38-acquisition-tex-c0016}\\
Coverage targeting & \datum{tables-r38-acquisition-tex-c0017} & \datum{tables-r38-acquisition-tex-c0018} & \datum{tables-r38-acquisition-tex-c0019} & \datum{tables-r38-acquisition-tex-c0020}\\
Fixed round-robin & \datum{tables-r38-acquisition-tex-c0021} & \datum{tables-r38-acquisition-tex-c0022} & \datum{tables-r38-acquisition-tex-c0023} & \datum{tables-r38-acquisition-tex-c0024}\\
\bottomrule\end{tabular}
\end{table}

Both intervention policies recover the true graph for all 3 seeds; matched additional observations leave 6 unoriented edges, not 6 wrong directions. Round-robin uses 1800 intervention records per run, whereas coverage uses 3600 for two seeds. This supports additional orientation information from interventions, not a coverage-budget advantage or universal inability of observations to orient edges.

Seed 6's coverage trajectory explains the cross-table difference. Here, targets 1 then 0 yield acceptance of $0\to1$ and complete orientation after 3600 graph records. The other two controls retain $0$--$1$ after the same target prefix, then repeatedly sample target 0 until budget exhaustion. Their fixed target-1 samples yield reverse-response scores of 0.15265 and 0.15226, respectively, both above threshold 0.15, so the two-sided rule continues to abstain. The 3/3 here and 2/3 in those controls therefore refer to different specified graph-sampling streams, not different calibration checkpoints or commitment rules. The present 3/3 does not establish stable coverage-policy success across sampling repetitions.

\subsubsection{Do calibrated readouts transfer to other response sources?}
Per seed, 512 independent exogenous contexts generate natural observations and 7 direct-intervention types with original ID values. Three calibration configurations share frozen encodings; target-state GT comes from the SCM. Each node is evaluated under direct targeting, natural observation, and intervention on other nodes. The last source includes ancestors and non-ancestors; records stratify the ancestor subset rather than treating all responses as indirect causal effects.

Errors cover the full sampled range and common bins, fixed at intervals of 0.5 over $[-2,2]$ in training-label-standardized coordinates. Retained bins contain at least 8 records per source; weighting is equal across bins, then nodes. All seeds have 7 scorable nodes, with ranges, counts, and exclusions saved. GT stratifies evaluation only; it neither updates mappings nor ensures identical within-bin distributions.

\begin{table}[htbp]
\centering\small
\caption{Cross-source readout MSE, multiplied by $10^3$, reported as mean $\pm$ sample SD over 3 seeds with equal weighting of 7 nodes per seed. Common bins are fixed in training-label units and require at least 8 records per source, node, and bin; common bins then receive equal weight. This controls coarse common support, not full distribution matching. Bold denotes the lowest point estimate in each panel, not statistical significance.}
\label{tab:r38-sources}
\begin{tabular}{lrrr}\toprule
Calibration & Direct intervention & Natural observation & Other-node intervention\\\midrule
\multicolumn{4}{l}{\textbf{Full sampled range}}\\
No calibration & \datum{tables-r38-sources-tex-c0001} & \datum{tables-r38-sources-tex-c0002} & \datum{tables-r38-sources-tex-c0003}\\
Fully fitted affine & \datum{tables-r38-sources-tex-c0004} & \datum{tables-r38-sources-tex-c0005} & \datum{tables-r38-sources-tex-c0006}\\
\rowcolor{OurRowBlue}
Nonlinear calibration & \datum{tables-r38-sources-tex-c0007} & \datum{tables-r38-sources-tex-c0008} & \datum{tables-r38-sources-tex-c0009}\\
\midrule
\multicolumn{4}{l}{\textbf{Fixed common bins}}\\
No calibration & \datum{tables-r38-sources-tex-c0010} & \datum{tables-r38-sources-tex-c0011} & \datum{tables-r38-sources-tex-c0012}\\
Fully fitted affine & \datum{tables-r38-sources-tex-c0013} & \datum{tables-r38-sources-tex-c0014} & \datum{tables-r38-sources-tex-c0015}\\
\rowcolor{OurRowBlue}
Nonlinear calibration & \datum{tables-r38-sources-tex-c0016} & \datum{tables-r38-sources-tex-c0017} & \datum{tables-r38-sources-tex-c0018}\\
\bottomrule
\end{tabular}
\end{table}

In Table~\ref{tab:r38-sources}, nonlinear calibration has the lowest mean common-bin error for all three sources, but fully fitted affine calibration is better over the full range for observations and other-node interventions. Readout gains thus depend on state range. Common bins do not fully match distributions, and unequal source errors do not establish elimination of source differences.
\FloatBarrier

\subsubsection{From calibrated coordinates to true-state and image queries}
Queries use the same-record comparison's frozen graphs and mechanisms for no calibration, fully fitted positive affine calibration, and nonlinear calibration, not the acquisition-policy graphs above. Seed 6 has unoriented edges in all three graphs and remains in the planned denominator but is blocked; seeds 5 and 7 have identical executable graphs.

Each valid seed uses 64 new factual contexts, 7 targets, ID commands at the training-label mean $\pm0.75$ standard deviations, and amplitude-shift commands at $\pm2$, giving 896 CF queries per class. GT replays the same exogenous context. Models receive only the factual image, target, and value, retain factual residuals, and inverse-calibrate before decoding. Population queries have no fixed factual image: model and GT means use 4096 samples from empirical pools and the SCM, respectively. Models share test contexts and random seeds; amplitude tests do not adjust models.

Predictions return to environmental units via training-concept scales, are compared with SCM states, and are normalized by those scales. CF-state columns measure individual outcomes; population columns measure each non-target node's interventional mean $\mu_k(j,u)$, $k\ne j$, from Section~\ref{sec:problem}. Both exclude the set target, so command copying is not predictive performance. Descendant, non-descendant, and factual-relative change errors are saved separately. State metrics are not independent decoder-semantic readouts; image metrics are pixel MSE of raw floating-point decoder outputs.

\begin{table}[htbp]
\centering\small\setlength{\tabcolsep}{3pt}
\caption{Query results with frozen graphs, mechanisms, and decoders; errors are multiplied by $10^3$. State columns use true simulator states as GT and exclude the direct target; the image column reports MSE of each model's own decoder output. Values are means $\pm$ sample SD over common valid seeds 5 and 7 out of 3 planned seeds; seed 6 is blocked because its frozen graph is not sufficiently oriented. Bold denotes the lowest point estimate within each command group; amplitude shift is not strict out-of-support extrapolation.}
\label{tab:r38-queries}
\begin{tabular}{llrrr}\toprule
Calibration & Command & CF-state NMSE & Population-mean NMSE & Image MSE\\\midrule
No calibration & ID & \datum{tables-r38-queries-tex-c0001} & \datum{tables-r38-queries-tex-c0002} & \datum{tables-r38-queries-tex-c0003}\\
Fully fitted affine & ID & \datum{tables-r38-queries-tex-c0004} & \datum{tables-r38-queries-tex-c0005} & \datum{tables-r38-queries-tex-c0006}\\
\rowcolor{OurRowBlue}
Nonlinear calibration & ID & \datum{tables-r38-queries-tex-c0007} & \datum{tables-r38-queries-tex-c0008} & \datum{tables-r38-queries-tex-c0009}\\
No calibration & Amplitude shift & \datum{tables-r38-queries-tex-c0010} & \datum{tables-r38-queries-tex-c0011} & \datum{tables-r38-queries-tex-c0012}\\
Fully fitted affine & Amplitude shift & \datum{tables-r38-queries-tex-c0013} & \datum{tables-r38-queries-tex-c0014} & \datum{tables-r38-queries-tex-c0015}\\
\rowcolor{OurRowBlue}
Nonlinear calibration & Amplitude shift & \datum{tables-r38-queries-tex-c0016} & \datum{tables-r38-queries-tex-c0017} & \datum{tables-r38-queries-tex-c0018}\\
\bottomrule\end{tabular}
\end{table}

On the 2 jointly executable seeds, nonlinear calibration reduces mean non-target CF-state NMSE versus no calibration by 6.51\% for ID and 4.41\% for amplitude shift; population-mean NMSE falls by 39.26\% and 12.33\%. These gains are unstable: both query errors rise on seed 5 and fall on seed 7. Against fully fitted affine calibration, only amplitude-shift population-mean error is lower; every other column is higher. Mean rankings do not establish significance.

For factual-relative \emph{change}, nonlinear calibration instead raises non-target effect NMSE versus no calibration by 0.88\% for ID and 11.77\% for amplitude shift. Absolute state, change, population mean, and image errors measure distinct abilities. Together with blocked seed 6, these results verify executable frozen-interface queries and range-limited readout gains, not a stable end-to-end calibration advantage.
\FloatBarrier

\subsection{Static mechanism comparisons on four MEC tasks}\label{app:static-factor}
Fixed static instances of Domino-MEC, PnP, Balance-MEC, and Gears-MEC use native encoder-training protocols and 8 seeds each; PnP uses held-out seeds 200--207. Within each environment, visual representations and mechanism families are fixed, varying only the graph interface. Greedy Interventional Equivalence Search (GIES)~\citep{hauser2012characterization} and passive graph extension receive the same numerical representations as our method. The metric is factor-level counterfactual NMAE: normalized absolute error in variable outcomes, not pixels. These graph-backend/mechanism comparisons are independent of the main-text RGB and rigid-body image benchmarks and do not evaluate image MSE or dynamic prediction.

\begin{table}[htbp]
\centering\small\setlength{\tabcolsep}{3pt}
\caption{Static factor-level counterfactual NMAE, lower is better. Means and 95\% bootstrap intervals come from the corresponding per-seed records. Metrics with different normalization definitions are not ranked together.}
\begin{tabular}{lrrrr}
\toprule
Method & Domino & PnP & Balance & Gears\\
\midrule
Our method & \datum{appendix-static-factor-tex-c0001} & \datum{appendix-static-factor-tex-c0002} & \datum{appendix-static-factor-tex-c0003} & \datum{appendix-static-factor-tex-c0004} \\
GIES & \datum{appendix-static-factor-tex-c0005} & \datum{appendix-static-factor-tex-c0006} & \datum{appendix-static-factor-tex-c0007} & \datum{appendix-static-factor-tex-c0008} \\
Passive graph extension & \datum{appendix-static-factor-tex-c0009} & \datum{appendix-static-factor-tex-c0010} & \datum{appendix-static-factor-tex-c0011} & \datum{appendix-static-factor-tex-c0012} \\
Shuffled graph & \datum{appendix-static-factor-tex-c0013} & \datum{appendix-static-factor-tex-c0014} & \datum{appendix-static-factor-tex-c0015} & \datum{appendix-static-factor-tex-c0016} \\
No graph propagation & \datum{appendix-static-factor-tex-c0017} & \datum{appendix-static-factor-tex-c0018} & \datum{appendix-static-factor-tex-c0019} & \datum{appendix-static-factor-tex-c0020} \\
True graph (structural reference) & \datum{appendix-static-factor-tex-c0021} & \datum{appendix-static-factor-tex-c0022} & \datum{appendix-static-factor-tex-c0023} & \datum{appendix-static-factor-tex-c0024} \\
\bottomrule
\end{tabular}
\end{table}

\emph{Passive graph extension} selects a compatible DAG from the observational graph by a prespecified rule; \emph{shuffled graph} changes node correspondence, and \emph{no graph propagation} removes inter-variable transmission. Protocol-fixed consistent extensions of partial graphs enable computation, not intervention-based identification. The supplied true graph is a structural reference with mechanisms fitted on the same visually encoded data. GIES has slightly lower error point estimates on PnP and Gears; our method is lower on Domino and Balance, subject to the reported intervals and task conditions.

Balance-MEC's skeleton-effect threshold was selected on development samples, unlike PnP's independent held-out seeds 200--207. Results retain these environment-specific selection and evaluation scopes.

\subsection{What-if and pixel-counterfactual protocols for three scenes}\label{app:new-mec-queries}
Native PnP, Balance, and Gears MEC implementations each use seeds 300 and 301 with learned $T$; the encoder, calibration map, and normalization statistics are frozen for graph discovery and queries. Structure learning matches the main-text implementation but uses an independent seed list and denominator. PnP selects among 3 encoder initializations by training-anchor validation error; Balance uses a frozen skeleton threshold of 0.06 and 90 epochs, and Gears uses 180 epochs. Discovery receives pixel encodings, not the true graph. Queries require a fully directed acyclic graph, without offline-SHD seed selection. All six runs remain: PnP and Gears execute 2/2, Balance 1/2, with seed 301 blocked by orientation ambiguity. Reported errors condition on valid runs; blocked runs are not zero errors.

\paragraph{Mechanism prediction.}
Local ridge mechanisms use ID observations and original target-selected intervention records processed by the same frozen visual interface, excluding each node's directly intervened samples. Roots sample saved observational readouts and non-roots fitted residuals, with replacement. Parameters are fixed before prediction, without ID/OOD test-query selection. Each node has two ID and two OOD commands at the training mapping's physical origin $\pm0.75$ and $\pm2$ scale units. Both methods share the original data adapter's training-normalization scale; test outcomes determine neither commands nor denominators.

The no-graph population comparator constructs a new empty-adjacency model from the same encoded records and masks. Every node becomes a root, independently sampled with replacement from its observational marginal pool (falling back to rows not intervening on that node), before clamping the target. It reuses neither parent coefficients nor residual pools. The shared random seed does not pair coordinate-wise noises because traversal order differs; implementation details appear in Appendix~\ref{app:nograph-operator}.

For target $j$ and command $u$, let $\overline s^{\rm GT}$ and $\overline s^{\rm pred}$ be 2048-sample simulator and model means in this protocol's scoring units, with training-fixed scale $d_k$:
\begin{equation}
E(j,u)=\frac{1}{K-1}\sum_{k\ne j}
\left(\frac{\overline s_k^{\rm pred}-\overline s_k^{\rm GT}}{d_k}\right)^2.
\end{equation}
Targets and positive/negative commands are equally weighted within each valid seed, then summarized by across-seed mean and sample SD; Balance reports its sole valid seed's point value. Excluding the set target avoids rewarding command copying, while retaining non-descendants measures preservation alongside propagation. Lower degradation factor $g=E_{\rm OOD}/E_{\rm ID}$ means less relative error growth, opposite to the stability ratio in Table~\ref{tab:ood-benchmark-matrix}: $r=1/g$ only for identical error summaries. Different run sets prohibit reciprocating values across these protocols. Raw records also retain worst-query and distribution-standard-deviation errors. This tests intervention-amplitude shifts, not strict out-of-support generalization or full-distribution distance.

\paragraph{Image queries.}
With the same encoder, learned graph, and mechanisms, a decoder fits reconstruction on 2400 independent ID observations and frozen encodings for 200 epochs, batch size 256, and Adam learning rate 0.001, without test factual--counterfactual pairs. Both methods share this decoder. Our method sets the factual target, retains residuals, and propagates to descendants before decoding; the empty-graph comparator sets only the target and preserves all other factual coordinates, without pool sampling. GT replays the same exogenous noise. The 32 fixed contexts, $K$ targets, and two commands yield $64K$ queries. Generated images are decoder outputs; the simulator supplies only factual images and paired GT.

\begin{figure}[!ht]
\centering\setlength{\tabcolsep}{2pt}
\begin{tabular}{cccc}
Factual image & Paired GT & Our method & No-graph ablation\\
\multicolumn{4}{c}{PnP: seed 300, fixed test context 0, target 0, positive command}\\
\includegraphics[width=.23\linewidth]{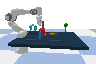}&\includegraphics[width=.23\linewidth]{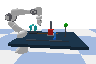}&\includegraphics[width=.23\linewidth]{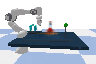}&\includegraphics[width=.23\linewidth]{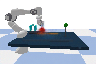}\\
\multicolumn{4}{c}{Balance: the same prespecified case-selection rule}\\
\includegraphics[width=.23\linewidth]{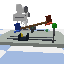}&\includegraphics[width=.23\linewidth]{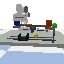}&\includegraphics[width=.23\linewidth]{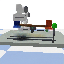}&\includegraphics[width=.23\linewidth]{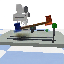}\\
\multicolumn{4}{c}{Gears: the same prespecified case-selection rule}\\
\includegraphics[width=.23\linewidth]{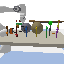}&\includegraphics[width=.23\linewidth]{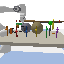}&\includegraphics[width=.23\linewidth]{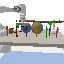}&\includegraphics[width=.23\linewidth]{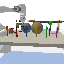}\\
\end{tabular}
\caption{Fixed queries in three controlled MEC scenes. Cases are prespecified by test index, not filtered by prediction quality. PnP uses $64\times96$ images and Balance/Gears use $64\times64$; this group is not pooled with rigid-body domino or RGB128 image protocols.}
\end{figure}

These within-implementation graph comparisons are limited by scene scale, valid seeds, and blocking rates. The no-graph ablation is not an independently published model and does not establish cross-scene superiority over CausalVAE.

\FloatBarrier
\section{Intervention-Value Calibration and Query Capability}
\subsection{Query metrics and interpretation of intervention-value calibration}
\label{app:retest-r42}
Table~\ref{tab:calibration-controlled} uses one three-seed cohort for readout, effect, and RGB comparisons, with the following metric and update permissions.

\paragraph{Counterfactual effect error.}
For one query, non-target factual-to-counterfactual change error is
\[
 \frac1{K-1}\sum_{k\ne j}
 \left[
 \frac{(\widehat s_k^{\rm cf}-\widehat s_k^{\rm fact})
 -(s_k^{\rm cf}-s_k^{\rm fact})}{\sigma_k^s}
 \right]^2.
\]
This is neither absolute-state NMSE nor population-mean NMSE without a factual image. Table~\ref{tab:calibration-controlled} reports readout, effect, and RGB errors; population means, graph discovery, and independent image readouts have separate evaluations.

\paragraph{Trainable modules and comparison scope.}
Identity, fully fitted affine, and nonlinear calibration share an ID-refitted decoder. Ordinary joint updating changes both readout heads and decoder, precluding attribution solely to invertibility, parameter count, or one loss term. Main-table means and SD aggregate equally weighted targets and commands within seeds, then across seeds; improved-seed counts and paired differences appear in Appendix~\ref{app:paired-evidence}.

\paragraph{Trade-offs in other controls.}
Appendix~\ref{app:interface-r39} uses the original decoder and sequential orientation updates, whereas the main cohort uses an ID-refitted decoder. Each cohort fixes its own test contexts and shares a decoder among calibration variants; readout, effect, and RGB statistics are not pooled across cohorts. Appendix~\ref{app:decoder-only-r40} isolates decoder replacement.

\subsection{Restricted affine and nonlinear calibration with a frozen decoder}\label{app:calibration-chain-r30}
Independent ID validation uses the seven-node domino scene with $64\times64$ RGB images. It reuses the frozen encoders/decoders for seeds 5, 6, and 7, 6000 observations, 896 target records, and shared coverage-targeting data, without new training or OOD selection. Each configuration has its own normalization, frozen graph, and mechanisms; graphs use the same evidence without resampling.

Restricted positive-affine calibration fixes curvature weights to 0; nonlinear calibration adds strictly monotone terms under the same learning schedule and data access. The uncalibrated control does not fit $T$ or reallocate its saved calibration budget to graph discovery.

All nine mappings $T$, graphs, and mechanisms are frozen before validation. Each seed contributes 64 independent random ID intervention readouts for each of 7 targets, totaling 1344 records. Graph-valid seeds additionally use 64 independent factual contexts and two commands per target at the training-anchor mean $\pm0.75$ standard deviations. Shared factual images and commands are paired with GT replay under fixed exogenous context. Configurations retain factual residuals in their own frozen coordinates/mechanisms and apply inverse-$T$ before the frozen decoder; predictors never receive GT states.

Seed 6 retains one unoriented edge in all configurations and is blocked, without arbitrary completion or zero errors. Readout/graph statistics use 3 seeds; CF uses the 2 jointly valid seeds, each with 896 paired queries. The 3264 evaluation calls comprise 1344 readout, 1792 CF GT, and 128 factual records. Another 32 consistency records are excluded; there are no training calls.

In Table~\ref{tab:calibration-chain-r30}, nonlinear calibration has the best readouts but unchanged graph recovery, smaller pixel gains, and slightly higher mean CF error than restricted affine. Invertible coordinate-wise maps add no population conditional-independence information; finite-sample and mechanism-fitting changes remain separate. Fully fitted affine and equal-total-budget controls appear in Appendix~\ref{app:ablation-complete-r33}.

\begin{table}[htbp]
\centering\small\setlength{\tabcolsep}{3pt}
\caption{Restricted affine and nonlinear calibration with a frozen decoder. ID readout and CF errors are multiplied by $10^3$, with mean $\pm$ sample SD across seeds. The graph column lists SHD of each configuration's frozen graphs for seeds 5/6/7. Seed 6 is blocked by the graph gate, so no CF error is entered.}
\label{tab:calibration-chain-r30}
\begin{tabularx}{\linewidth}{Xrrr}
\toprule
Configuration & Target MSE (3/3) & SHD & CF pixel MSE (2/3)\\
\midrule
Uncalibrated & \datum{tables-calibration-chain-tex-c0001} & \datum{tables-calibration-chain-tex-c0002} & \datum{tables-calibration-chain-tex-c0003}\\
Restricted positive affine & \datum{tables-calibration-chain-tex-c0004} & \datum{tables-calibration-chain-tex-c0005} & \datum{tables-calibration-chain-tex-c0006}\\
\rowcolor{OurRowBlue}
Nonlinear calibration & \datum{tables-calibration-chain-tex-c0007} & \datum{tables-calibration-chain-tex-c0008} & \datum{tables-calibration-chain-tex-c0009}\\
\bottomrule
\end{tabularx}
\end{table}

\subsection{Matched-record validation of intervention-value calibration and standard supervised updating}
\label{app:interface-r39}
This matched-record control uses the seven-node domino scene with $64\times64$ RGB images and encoder seeds 5, 6, and 7 to separate readout, numerical-query, and image changes. It jointly evaluates calibration, sequential graph updates, local mechanisms, and both query types; every graph is learned from this experiment's acquisition records.

\subsubsection{Shared data and update permissions}
Each seed reuses the frozen backbone, 6000 observations, and 896 verification records. Original seeds and commands replay training images, individually checked against saved encodings, without adding independent data. Uncalibrated, fully fitted positive-affine, and nonlinear configurations share frozen encoder/decoder weights. Nonlinear calibration has 49 coefficients: gain, bias, and five basis weights for each of seven coordinates.

Ordinary joint updating is a same-supervision control, not a published method. With convolutional features frozen, seven linear heads are refitted on 256-dimensional features, using 1799 parameters. New labels come only from verified direct-target values; observation-preservation weight is 0.01. Four-fold cross-validation within the original 896 records selects ridge weight from $\{10^{-4},10^{-2},1\}$, followed by fitting all records. The shared decoder checkpoint refits only reconstruction of 6000 observation and 896 calibration images for 40 epochs, batch size 128, Adam learning rate $10^{-4}$, using the final epoch. Its 1,154,659 parameters receive no test CF pairs. Shared data and labels but different trainable modules/parameter counts prevent separate attribution to heads or decoder.

Table~\ref{tab:current-interface-r39} combines readout and query results under these update permissions. Its original decoder and test queries differ from the ID-refitted decoder and queries in Table~\ref{tab:calibration-controlled}; the cohorts are not pooled.

\subsubsection{From newly acquired graphs to frozen queries}
After all visual interfaces freeze, nonlinear calibration selects targets by the main-text coverage rule: 1800 new pixel observations per batch, at most six batches. Appendix~\ref{app:algorithm}'s sequential tentative additions, consistency checks, and Meek closure replace batch commitment. Other configurations receive exactly the acquired images and commands, learn graphs in their own coordinates, and fit local linear mechanisms. Acquisition records are matched; the four configurations do not optimize separate active policies.

The seeds use 1800, 3600, and 3600 graph records. Every configuration pays for 896 calibration records; observations are reported separately. Stopping is truth-blind and structure is scored after model freezing. All four configurations have SHD 0 and no unoriented edges in all three seeds: every query executes, without evidence of improved recovery from calibration. The batch-commitment 2/3 denominator belongs to a different update procedure and is not directly comparable.

Each seed uses 64 new factual contexts, 7 targets, and four commands per target: training-label mean $\pm0.75$ standard deviations for ID and $\pm2$ for amplitude shift. Each seed/split has $64\times7\times2=896$ paired CF queries: 1792 per configuration/seed across splits and 5376 across three seeds, independent of the 896 training verification records. GT replays fixed exogenous context; models receive only factual images and commands. Population evaluation compares non-target means from 4096 model and simulator outcomes each. Graphs, mechanisms, visual modules, and statistics stay frozen.

\begin{table}[htbp]
\centering\footnotesize\setlength{\tabcolsep}{3pt}
\caption{Matched-record controls with sequential direction updates in the seven-node domino scene ($64\times64$ RGB images). Errors are multiplied by $10^3$ and reported as mean $\pm$ sample SD across three encoder seeds; all configurations have 3/3 valid seeds and 896 CF queries per seed and split. The top panel reports target-readout MSE in standardized label units. Updated modules are, respectively, none, affine $T$, nonlinear $T$, and readout heads plus $D$; joint updating shares supervision but differs in parameter count. The bottom panel excludes the direct target from state, effect, and population-mean NMSE. Bold marks the lowest comparable point estimate; blue marks nonlinear calibration.}
\label{tab:current-interface-r39}
\label{tab:current-queries-r39}
{\small
\begin{tabularx}{\linewidth}{Xcccc}\toprule
Configuration & Uncalibrated & \shortstack{Fully fitted\\affine} & \shortstack{Nonlinear\\calibration} & \shortstack{Ordinary joint\\updating}\\\midrule
Target-readout MSE & \datum{tables-current-interface-tex-c0001} & \datum{tables-current-interface-tex-c0004} & \cellcolor{OurRowBlue}\datum{tables-current-interface-tex-c0007} & \datum{tables-current-interface-tex-c0010}\\
\bottomrule\end{tabularx}
}
\par\medskip
\begin{tabularx}{\linewidth}{Xlrrrr}\toprule
Configuration & Command & \shortstack{State\\NMSE} & \shortstack{Effect\\NMSE} & \shortstack{Mean\\NMSE} & \shortstack{Image\\MSE}\\\midrule
Uncalibrated & ID & \datum{tables-current-queries-tex-c0001} & \datum{tables-current-queries-tex-c0002} & \datum{tables-current-queries-tex-c0003} & \datum{tables-current-queries-tex-c0004}\\
Fully fitted affine & ID & \datum{tables-current-queries-tex-c0005} & \datum{tables-current-queries-tex-c0006} & \datum{tables-current-queries-tex-c0007} & \datum{tables-current-queries-tex-c0008}\\
\rowcolor{OurRowBlue}
Nonlinear calibration & ID & \datum{tables-current-queries-tex-c0009} & \datum{tables-current-queries-tex-c0010} & \datum{tables-current-queries-tex-c0011} & \datum{tables-current-queries-tex-c0012}\\
Ordinary joint updating & ID & \datum{tables-current-queries-tex-c0013} & \datum{tables-current-queries-tex-c0014} & \datum{tables-current-queries-tex-c0015} & \datum{tables-current-queries-tex-c0016}\\
\midrule
Uncalibrated & Amplitude shift & \datum{tables-current-queries-tex-c0017} & \datum{tables-current-queries-tex-c0018} & \datum{tables-current-queries-tex-c0019} & \datum{tables-current-queries-tex-c0020}\\
Fully fitted affine & Amplitude shift & \datum{tables-current-queries-tex-c0021} & \datum{tables-current-queries-tex-c0022} & \datum{tables-current-queries-tex-c0023} & \datum{tables-current-queries-tex-c0024}\\
\rowcolor{OurRowBlue}
Nonlinear calibration & Amplitude shift & \datum{tables-current-queries-tex-c0025} & \datum{tables-current-queries-tex-c0026} & \datum{tables-current-queries-tex-c0027} & \datum{tables-current-queries-tex-c0028}\\
Ordinary joint updating & Amplitude shift & \datum{tables-current-queries-tex-c0029} & \datum{tables-current-queries-tex-c0030} & \datum{tables-current-queries-tex-c0031} & \datum{tables-current-queries-tex-c0032}\\
\bottomrule\end{tabularx}
\end{table}

Nonlinear target-readout error falls by 46.26\% versus no calibration and 15.21\% versus fully fitted affine, improving all three seeds. Affine nevertheless has lower mean numerical-query error. Nonlinear population-mean error versus no calibration falls by 24.10\% for ID with only one seed improving, and 16.64\% for amplitude shift with two improving: aggregate reductions are not consistent gains. Against joint updating, its amplitude-shift population-mean error falls by 71.03\% and improves all three seeds, but joint updating has lower image error. These evaluations are not interchangeable.

\subsubsection{Independent evaluation of visible states in generated images}
An independent image readout $Q$, with no shared weights, evaluates renderer-visible activations $\gamma(s_j)=1/(1+\exp(-3.8s_j))$, not arbitrary-precision recovery of $s_j$ from saturated pixels. Independently sampled node activations break causal-correlation shortcuts: an equal mixture of $\mathcal U(0.001,0.999)$ and Beta$(0.5,0.5)$ is clipped to $[10^{-5},1-10^{-5}]$, inverse-mapped, and rendered. Its 8192 training and 1024 independent validation images train none of the four configurations.

$Q$ has three stride-2 convolutions (32/64/96 channels), a 256-dimensional fully connected layer, and seven sigmoid outputs. Adam uses learning rate $10^{-3}$, batch size 128, and 80 epochs, with random $3\times3$ average blur. Checkpoints minimize the mean clean/blurred independent-validation MSE. All seven nodes pass preset gates of MAE at most 0.08 and $R^2$ at least 0.5 on both validation types: clean MAE is 0.0066--0.0190, blurred 0.0066--0.0249. This learned measurement is not error-free.

For direct-target, true-descendant, or true-non-descendant sets $A$, absolute error compares $Q_k(\widehat x^{\rm cf})$ to true $\gamma(s_k^{\rm cf})$; $Q_k(x^{\rm GT})$ supplies only a measurement-error reference. Change evaluation uses factual reconstruction $x^{\rm rec}=D_\psi(E_\theta(x))$:
\[
 \big[Q_k(\widehat x^{\rm cf})-Q_k(x^{\rm rec})\big]
 \quad\text{with}\quad
 \big[\gamma(s_k^{\rm cf})-\gamma(s_k^{\rm fact})\big].
\]
Absolute differences average within nonempty $A$, equally across targets/commands, then across seeds for mean and SD. Empty sets are not zero. True graphs stratify offline evaluation only.

\begin{table}[htbp]
\centering\footnotesize\setlength{\tabcolsep}{2.6pt}
\caption{Visible-state MAE measured by an independent image readout model, multiplied by $10^3$; mean $\pm$ SD across three encoder seeds. Target, descendant, and non-descendant readouts are scored against true simulator activations; changes use each model's factual reconstruction as the reference. GT-image readout errors are reported separately as a measurement-error reference, not as a substitute for ground truth. Bold marks the lowest point estimate within each split and column.}
\label{tab:current-semantics-r39}
\begin{tabularx}{\linewidth}{Xlrrrrr}\toprule
Configuration & Command & Target & Desc. & Non-desc. & \shortstack{Desc.\\change} & \shortstack{Non-desc.\\change}\\\midrule
Uncalibrated & ID & \datum{tables-current-semantics-tex-c0001} & \datum{tables-current-semantics-tex-c0002} & \datum{tables-current-semantics-tex-c0003} & \datum{tables-current-semantics-tex-c0004} & \datum{tables-current-semantics-tex-c0005}\\
Fully fitted affine & ID & \datum{tables-current-semantics-tex-c0006} & \datum{tables-current-semantics-tex-c0007} & \datum{tables-current-semantics-tex-c0008} & \datum{tables-current-semantics-tex-c0009} & \datum{tables-current-semantics-tex-c0010}\\
\rowcolor{OurRowBlue}
Nonlinear calibration & ID & \datum{tables-current-semantics-tex-c0011} & \datum{tables-current-semantics-tex-c0012} & \datum{tables-current-semantics-tex-c0013} & \datum{tables-current-semantics-tex-c0014} & \datum{tables-current-semantics-tex-c0015}\\
Ordinary joint updating & ID & \datum{tables-current-semantics-tex-c0016} & \datum{tables-current-semantics-tex-c0017} & \datum{tables-current-semantics-tex-c0018} & \datum{tables-current-semantics-tex-c0019} & \datum{tables-current-semantics-tex-c0020}\\
\midrule
Uncalibrated & Amplitude & \datum{tables-current-semantics-tex-c0021} & \datum{tables-current-semantics-tex-c0022} & \datum{tables-current-semantics-tex-c0023} & \datum{tables-current-semantics-tex-c0024} & \datum{tables-current-semantics-tex-c0025}\\
Fully fitted affine & Amplitude & \datum{tables-current-semantics-tex-c0026} & \datum{tables-current-semantics-tex-c0027} & \datum{tables-current-semantics-tex-c0028} & \datum{tables-current-semantics-tex-c0029} & \datum{tables-current-semantics-tex-c0030}\\
\rowcolor{OurRowBlue}
Nonlinear calibration & Amplitude & \datum{tables-current-semantics-tex-c0031} & \datum{tables-current-semantics-tex-c0032} & \datum{tables-current-semantics-tex-c0033} & \datum{tables-current-semantics-tex-c0034} & \datum{tables-current-semantics-tex-c0035}\\
Ordinary joint updating & Amplitude & \datum{tables-current-semantics-tex-c0036} & \datum{tables-current-semantics-tex-c0037} & \datum{tables-current-semantics-tex-c0038} & \datum{tables-current-semantics-tex-c0039} & \datum{tables-current-semantics-tex-c0040}\\
\bottomrule\end{tabularx}
\end{table}

\FloatBarrier

Under amplitude shift, nonlinear target image-readout error falls by 15.52\% versus no calibration and 7.31\% versus affine, improving all three seeds; affine has better ID target readouts. Raw MAEs (not multiplied by $10^3$) show non-descendant error 0.18491, above uncalibrated 0.17422 and affine 0.18121, versus GT-image reference 0.01335. Model non-descendant states stay invariant across commands, but decoded images show substantial readout changes, localizing this preservation failure to the frozen decoding path rather than graph recovery or masking. Joint updating alleviates it without separating head updates, decoder refitting, or their interaction.

\subsubsection{Per-seed results}
\begin{table}[htbp]
\centering\small\setlength{\tabcolsep}{3pt}
\caption{Per-seed results for matched-record controls with sequential direction updates; errors are multiplied by $10^3$. The four configurations within each seed share acquired pixels and test contexts. All graphs have SHD=0, and all three prespecified seeds are scored.}
\label{tab:current-seeds-r39}
\begin{tabularx}{\linewidth}{cXrrrr}\toprule
Seed & Configuration & \shortstack{Readout\\MSE} & \shortstack{ID mean\\NMSE} & \shortstack{Amplitude mean\\NMSE} & \shortstack{ID image\\MSE}\\\midrule
\datum{tables-current-per-seed-tex-c0001} & Uncalibrated & \datum{tables-current-per-seed-tex-c0002} & \datum{tables-current-per-seed-tex-c0003} & \datum{tables-current-per-seed-tex-c0004} & \datum{tables-current-per-seed-tex-c0005}\\
\datum{tables-current-per-seed-tex-c0006} & Fully fitted affine & \datum{tables-current-per-seed-tex-c0007} & \datum{tables-current-per-seed-tex-c0008} & \datum{tables-current-per-seed-tex-c0009} & \datum{tables-current-per-seed-tex-c0010}\\
\rowcolor{OurRowBlue}
\datum{tables-current-per-seed-tex-c0011} & Nonlinear calibration & \datum{tables-current-per-seed-tex-c0012} & \datum{tables-current-per-seed-tex-c0013} & \datum{tables-current-per-seed-tex-c0014} & \datum{tables-current-per-seed-tex-c0015}\\
\datum{tables-current-per-seed-tex-c0016} & Ordinary joint updating & \datum{tables-current-per-seed-tex-c0017} & \datum{tables-current-per-seed-tex-c0018} & \datum{tables-current-per-seed-tex-c0019} & \datum{tables-current-per-seed-tex-c0020}\\
\midrule
\datum{tables-current-per-seed-tex-c0021} & Uncalibrated & \datum{tables-current-per-seed-tex-c0022} & \datum{tables-current-per-seed-tex-c0023} & \datum{tables-current-per-seed-tex-c0024} & \datum{tables-current-per-seed-tex-c0025}\\
\datum{tables-current-per-seed-tex-c0026} & Fully fitted affine & \datum{tables-current-per-seed-tex-c0027} & \datum{tables-current-per-seed-tex-c0028} & \datum{tables-current-per-seed-tex-c0029} & \datum{tables-current-per-seed-tex-c0030}\\
\rowcolor{OurRowBlue}
\datum{tables-current-per-seed-tex-c0031} & Nonlinear calibration & \datum{tables-current-per-seed-tex-c0032} & \datum{tables-current-per-seed-tex-c0033} & \datum{tables-current-per-seed-tex-c0034} & \datum{tables-current-per-seed-tex-c0035}\\
\datum{tables-current-per-seed-tex-c0036} & Ordinary joint updating & \datum{tables-current-per-seed-tex-c0037} & \datum{tables-current-per-seed-tex-c0038} & \datum{tables-current-per-seed-tex-c0039} & \datum{tables-current-per-seed-tex-c0040}\\
\midrule
\datum{tables-current-per-seed-tex-c0041} & Uncalibrated & \datum{tables-current-per-seed-tex-c0042} & \datum{tables-current-per-seed-tex-c0043} & \datum{tables-current-per-seed-tex-c0044} & \datum{tables-current-per-seed-tex-c0045}\\
\datum{tables-current-per-seed-tex-c0046} & Fully fitted affine & \datum{tables-current-per-seed-tex-c0047} & \datum{tables-current-per-seed-tex-c0048} & \datum{tables-current-per-seed-tex-c0049} & \datum{tables-current-per-seed-tex-c0050}\\
\rowcolor{OurRowBlue}
\datum{tables-current-per-seed-tex-c0051} & Nonlinear calibration & \datum{tables-current-per-seed-tex-c0052} & \datum{tables-current-per-seed-tex-c0053} & \datum{tables-current-per-seed-tex-c0054} & \datum{tables-current-per-seed-tex-c0055}\\
\datum{tables-current-per-seed-tex-c0056} & Ordinary joint updating & \datum{tables-current-per-seed-tex-c0057} & \datum{tables-current-per-seed-tex-c0058} & \datum{tables-current-per-seed-tex-c0059} & \datum{tables-current-per-seed-tex-c0060}\\
\bottomrule\end{tabularx}
\end{table}

\FloatBarrier
Table~\ref{tab:current-seeds-r39} retains all three prespecified seeds. Percentage improvements are ratios of means.

\subsection{Seed-paired comparisons for the main-text controlled experiment}\label{app:paired-evidence}
For seeds 5, 6, and 7, no calibration, fully fitted affine, and nonlinear calibration share fixed encoders, learned graphs, fitting records, queries, and a seed-specific ID-refitted decoder; the original decoder is a separate sensitivity condition. Each seed uses 64 factual contexts, 7 targets, and four commands, totaling 1792 CF queries. ID/amplitude commands are training-label mean $\pm0.75$/$\pm2$ standard deviations. All 3 training seeds remain the statistical denominator, not image count.

For configuration $a$ and seed $s$, $E_{a,s}$ equally weights targets and commands within that seed. Paired difference $d_s=E_{a,s}-E_{b,s}$ favors $a$ when negative. Table~\ref{tab:paired-native-calibration} reports $\overline d$, paired sample SD, and counts with $d_s<0$. These three seeds descriptively assess repeatability of this frozen cohort, not other tables' counts or variances.

\begin{table}[htbp]
\centering\small\setlength{\tabcolsep}{4pt}
\caption{Paired differences for the same three-seed cohort as main-text Table~\ref{tab:calibration-controlled}. Errors and SDs are multiplied by $10^3$; parentheses give the number of improved seeds/3. SD is computed from per-seed differences, not by subtracting the two groups' SDs.}
\label{tab:paired-native-calibration}
\begin{tabularx}{\linewidth}{Xlcc}
\toprule
\shortstack[l]{Comparison\\(first minus second)} & Query & \shortstack{Non-target effect\\NMSE difference} & \shortstack{RGB MSE\\difference}\\
\midrule
Affine $-$ uncalibrated & ID & \datum{appendix-paired-evidence-tex-c0001} & \datum{appendix-paired-evidence-tex-c0002}\\
Affine $-$ uncalibrated & Amplitude shift & \datum{appendix-paired-evidence-tex-c0003} & \datum{appendix-paired-evidence-tex-c0004}\\
Nonlinear $-$ uncalibrated & ID & \datum{appendix-paired-evidence-tex-c0005} & \datum{appendix-paired-evidence-tex-c0006}\\
Nonlinear $-$ uncalibrated & Amplitude shift & \datum{appendix-paired-evidence-tex-c0007} & \datum{appendix-paired-evidence-tex-c0008}\\
Nonlinear $-$ affine & ID & \datum{appendix-paired-evidence-tex-c0009} & \datum{appendix-paired-evidence-tex-c0010}\\
Nonlinear $-$ affine & Amplitude shift & \datum{appendix-paired-evidence-tex-c0011} & \datum{appendix-paired-evidence-tex-c0012}\\
\bottomrule
\end{tabularx}
\end{table}

Nonlinear calibration lowers mean non-target effect error versus no calibration in both query types, but worsens one seed each; amplitude-shift RGB improves in 2/3 seeds yet has a slightly higher mean. Against fully fitted affine, nonlinear mean amplitude effect error is lower, but ID effect and both RGB errors are higher. Means and improved-seed counts support only a conditional calibration benefit, not universal nonlinear superiority.

Standard supervised updating uses the same direct-target labels but updates heads and decoder. Its state/effect errors are comparable on the same queries and physical units; its image results cannot support attribution to $T$ alone. Main-table point estimates without per-seed records do not borrow this cohort's SDs or denominator.

\subsubsection{Gears-MEC transfer with fixed learned graphs}\label{app:gears-paired}
Gears-MEC's six-node $drive\to g_1\to g_2\to ptr\to led\leftarrow target$ includes negative coefficients and a collider, differing from domino while remaining a linear additive SCM. Seeds 300 and 301 supply independently trained aligned grayscale CNNs and RGB decoders with $64\times64$ inputs. Encoders, learned complete graphs, and decoders stay frozen; only calibration and coordinate-specific local linear mechanisms refit. This does not vary front-end training or test new graph-recovery gains.

Each seed shares 5000 original observations and 900 graph-intervention records, adding 768 calibration records in two rounds per node, 64 records per round. Only verified direct-target values supervise calibration. Nonlinear calibration uses Eq.~\eqref{eq:short-calibration}'s five bases, 100 cumulative-replay steps per round, Adam learning rate 0.01, and preservation weight 0.01; affine fully fits the same terms. New records serve calibration only; mechanisms use original records and ridge coefficient $10^{-3}$. Inverse calibration returns to original encoder coordinates before original decoder normalization, with maximum round-trip error below $10^{-10}$.

After freezing, each seed uses 64 new factual contexts, six targets, and four commands: 1536 paired CFs at training-label mean $\pm0.75/\pm2$ standard deviations for ID/amplitude shift. Each target adds 64 ID direct-readout evaluations; each population query uses 4096 model and 4096 independent simulator samples. Training labels fix command units without test selection. This split does not replace main-table runs; its two independent visual models give $n=2$, with data and queries shared among calibration configurations within seeds.

\begin{table}[htbp]
\centering\small\setlength{\tabcolsep}{4pt}
\caption{Calibration transfer on Gears-MEC with fixed graphs, seeds 300/301. All errors are multiplied by $10^3$ and reported as mean $\pm$ sample SD across seeds; readouts use independent ID verification. All three configurations share $D$; $n=2$ permits only descriptive comparison.}
\label{tab:gears-paired}
\begin{tabularx}{\linewidth}{Xlcccc}
\toprule
Calibration & Query & \shortstack{ID readout\\MSE} & \shortstack{Non-target effect\\NMSE} & \shortstack{Population mean\\NMSE} & RGB MSE\\
\midrule
Uncalibrated & ID & \datum{tables-gears-paired-tex-c0001} & \datum{tables-gears-paired-tex-c0002} & \datum{tables-gears-paired-tex-c0003} & \datum{tables-gears-paired-tex-c0004}\\
Uncalibrated & Amplitude shift & \datum{tables-gears-paired-tex-c0005} & \datum{tables-gears-paired-tex-c0006} & \datum{tables-gears-paired-tex-c0007} & \datum{tables-gears-paired-tex-c0008}\\
Affine & ID & \datum{tables-gears-paired-tex-c0009} & \datum{tables-gears-paired-tex-c0010} & \datum{tables-gears-paired-tex-c0011} & \datum{tables-gears-paired-tex-c0012}\\
Affine & Amplitude shift & \datum{tables-gears-paired-tex-c0013} & \datum{tables-gears-paired-tex-c0014} & \datum{tables-gears-paired-tex-c0015} & \datum{tables-gears-paired-tex-c0016}\\
Nonlinear & ID & \datum{tables-gears-paired-tex-c0017} & \datum{tables-gears-paired-tex-c0018} & \datum{tables-gears-paired-tex-c0019} & \datum{tables-gears-paired-tex-c0020}\\
Nonlinear & Amplitude shift & \datum{tables-gears-paired-tex-c0021} & \datum{tables-gears-paired-tex-c0022} & \datum{tables-gears-paired-tex-c0023} & \datum{tables-gears-paired-tex-c0024}\\
\bottomrule
\end{tabularx}
\end{table}

Both calibrations slightly lower mean ID target-readout error but raise mean non-target effect, population-mean, and RGB errors. Nonlinear amplitude effect NMSE rises from uncalibrated $1.556\times10^{-3}$ to $2.324\times10^{-3}$: better direct readouts do not guarantee better propagation. Two seeds establish neither general failure conditions nor single-component attribution.

Shared-decoder RGB error is separate from effect error against simulator states. Without an independent Gears image-semantic readout, these results do not verify generated target/descendant/non-descendant semantics. Both prespecified seeds remain; results reselect neither calibration nor test scope.

\FloatBarrier
\section{Decoders and Paired Image Evaluation}
\subsection{Paired decoder diagnostics with fixed causal-state predictions}
\label{app:decoder-only-r40}
This post hoc paired diagnostic changes only the decoder, fixing encoder, calibration, learned graph, and mechanisms to separate RGB quality from causal computation. Test contexts are independent; amplitude shift uses the same command distribution without method selection.

\paragraph{Training and testing.}
For the seven-node domino scene with $64\times64$ RGB images, seeds 5, 6, and 7 each refit one decoder shared by uncalibrated, fully fitted affine, and nonlinear configurations. Deterministic replays of 6000 observation and 896 verification images must match saved encodings within $2\times10^{-4}$. No independent training data, GT CFs, test states, or test images are added. Decoder-only training minimizes ID reconstruction squared error for 40 epochs, batch size 128, Adam learning rate $10^{-4}$, using the final epoch without test selection. Other parameters and caches remain unchanged.

After freezing all modules, each seed uses 64 independent factual contexts, 7 targets, and standardized commands $\{-0.75,0.75,-2,2\}$: 896 paired CFs per seed/split. Factual/GT images share simulator exogenous context; configurations share factual images, commands, GT, and the independent readout. Each configuration has 5376 CFs across three seeds, not 5376 independent seeds. Frozen ordinary joint updating is only a same-query reference, outside decoder-only pairing.

\begin{table}[htbp]
\centering\footnotesize\setlength{\tabcolsep}{3pt}
\caption{Paired decoder diagnostics; errors are multiplied by $10^3$, with mean $\pm$ sample SD across three seeds and 896 paired queries per seed/split. The upper panel compares frozen $D$ and ID-refitted $D'$ on the same images with encoder, calibration, graph, mechanisms, and state predictions fixed. Joint updating changes heads and decoder and is only a reference. Visible-activation MAE uses an independent frozen readout against simulator activations; GT-image readout error is only a measurement reference. The lower panel reports $10^3(\mathrm{MSE}_{D'}-\mathrm{MSE}_{D})$: negative is better, and parentheses count improved seeds for each pair. Its SD is computed from per-seed differences, not subtracted group SDs.}
\label{tab:decoder-only-r40}
\label{tab:decoder-delta-r40}
\begin{tabularx}{\linewidth}{@{}Xrrrr@{}}\toprule
Configuration & \shortstack{ID\\image MSE$\downarrow$} & \shortstack{Amplitude shift\\image MSE$\downarrow$} & \shortstack{Amplitude shift\\target visible\\MAE$\downarrow$} & \shortstack{Amplitude shift\\non-descendant\\effect MAE$\downarrow$}\\\midrule
Uncalibrated, frozen $D$ & \datum{tables-decoder-only-tex-c0001} & \datum{tables-decoder-only-tex-c0002} & \datum{tables-decoder-only-tex-c0003} & \datum{tables-decoder-only-tex-c0004}\\
Uncalibrated, ID-refitted $D$ & \datum{tables-decoder-only-tex-c0005} & \datum{tables-decoder-only-tex-c0006} & \datum{tables-decoder-only-tex-c0007} & \datum{tables-decoder-only-tex-c0008}\\
Fully fitted affine, frozen $D$ & \datum{tables-decoder-only-tex-c0009} & \datum{tables-decoder-only-tex-c0010} & \datum{tables-decoder-only-tex-c0011} & \datum{tables-decoder-only-tex-c0012}\\
Fully fitted affine, ID-refitted $D$ & \datum{tables-decoder-only-tex-c0013} & \datum{tables-decoder-only-tex-c0014} & \datum{tables-decoder-only-tex-c0015} & \datum{tables-decoder-only-tex-c0016}\\
Nonlinear calibration, frozen $D$ & \datum{tables-decoder-only-tex-c0017} & \datum{tables-decoder-only-tex-c0018} & \datum{tables-decoder-only-tex-c0019} & \datum{tables-decoder-only-tex-c0020}\\
Nonlinear calibration, ID-refitted $D$ & \datum{tables-decoder-only-tex-c0021} & \datum{tables-decoder-only-tex-c0022} & \datum{tables-decoder-only-tex-c0023} & \datum{tables-decoder-only-tex-c0024}\\
\midrule
Ordinary joint updating (reference) & \datum{tables-decoder-only-tex-c0025} & \datum{tables-decoder-only-tex-c0026} & \datum{tables-decoder-only-tex-c0027} & \datum{tables-decoder-only-tex-c0028}\\
\bottomrule\end{tabularx}
\par\medskip
\begin{tabularx}{\linewidth}{Xrr}\toprule
\multicolumn{3}{l}{Paired image-MSE difference (improved seeds)}\\
Calibration & ID & Amplitude shift\\\midrule
Uncalibrated & \datum{tables-decoder-only-tex-c0029} (\datum{tables-decoder-only-tex-c0030}) & \datum{tables-decoder-only-tex-c0031} (\datum{tables-decoder-only-tex-c0032})\\
Fully fitted affine & \datum{tables-decoder-only-tex-c0033} (\datum{tables-decoder-only-tex-c0034}) & \datum{tables-decoder-only-tex-c0035} (\datum{tables-decoder-only-tex-c0036})\\
Nonlinear calibration & \datum{tables-decoder-only-tex-c0037} (\datum{tables-decoder-only-tex-c0038}) & \datum{tables-decoder-only-tex-c0039} (\datum{tables-decoder-only-tex-c0040})\\
\bottomrule\end{tabularx}
\end{table}

\FloatBarrier

\paragraph{Image gains with unchanged causal states.}
All three configurations lower RGB error in both splits, with 3/3 seeds improving per pair. Nonlinear ID pixel MSE falls from $2.246\times10^{-3}$ to $1.058\times10^{-3}$ and amplitude MSE from $1.695\times10^{-3}$ to $0.859\times10^{-3}$, mean reductions of 52.90\% and 49.32\%. CF latent variables, physical states, and numerical errors stay unchanged: gains arise from decoder adaptation, not graph or causal-state improvement.

Under amplitude shift, the frozen readout measures target visible-activation MAE falling from 0.06658 to 0.01539 and non-descendant change MAE from 0.15564 to 0.01806, not errors against direct physical-state labels. Its GT-image target MAE, approximately 0.02511, is a measurement reference, neither a theoretical lower bound nor evidence that generated images surpass GT. Affine/uncalibrated refitted decoders remain better on several metrics, and joint updating yields similar images; nonlinear calibration does not uniquely obtain these gains.

\paragraph{Pairing and cost.}
Non-decoder weights are identical; maximum paired CF-latent difference is $1.78\times10^{-15}$, and state/effect NMSE differences are both 0. Queries average within seeds before reporting three-seed mean, sample SD, and paired differences; paired SD is not a difference of group SDs.

Each seed renders 8752 images: 6896 training replays and 1856 factual/test-CF images. Test rendering is outside the training-intervention budget; every configuration pays calibration cost 896 and graph-discovery cost. Computation totals approximately 0.192 GPU-hours.

\subsection{Paired counterfactual images with learned calibration}
This shared-decoder comparison fixes factual index and command. Its environment differs from the main-text physical baseline, precluding cross-figure rankings.
\begin{figure}[htbp]
\centering
\includegraphics[width=.96\linewidth]{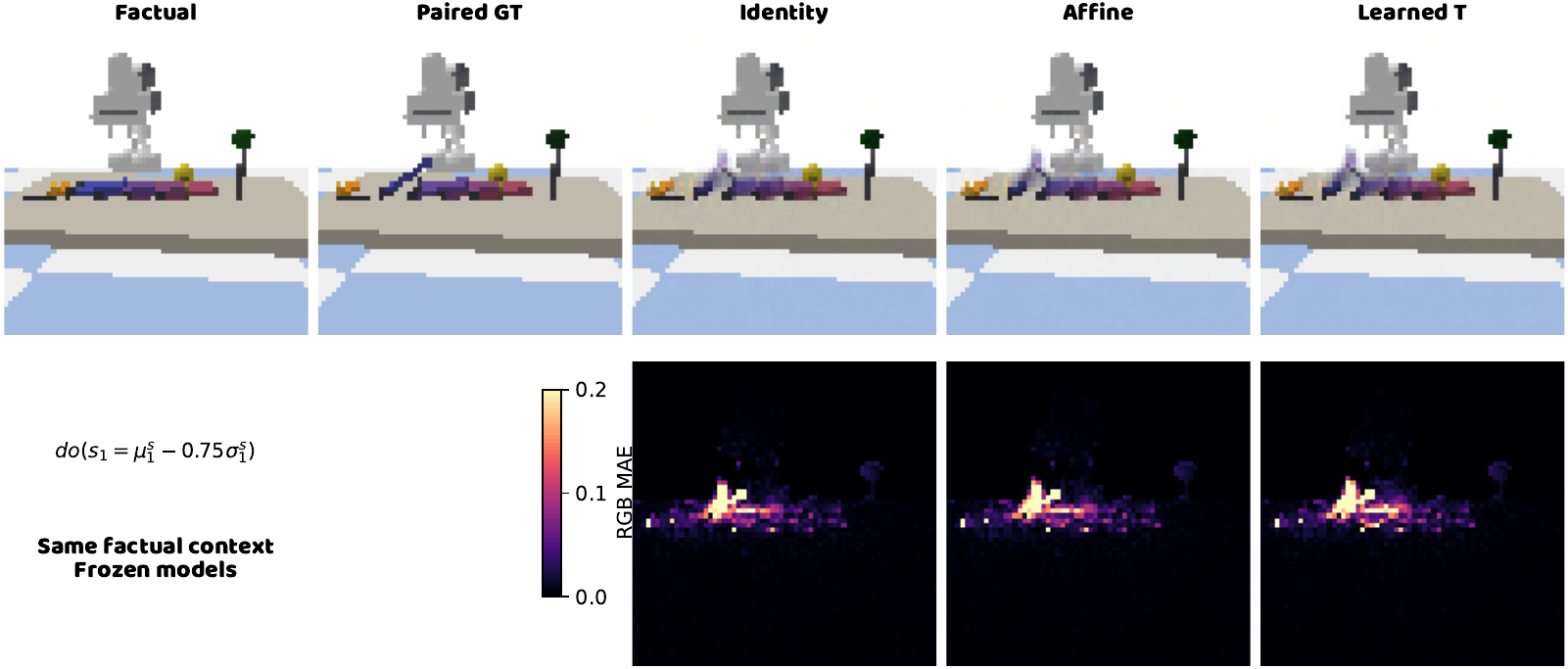}
\caption{Counterfactual queries in the seven-node domino scene with $64\times64$ RGB images, at fixed seed 5, factual index 0, target 1, and ID command $u=\mu_1^s-0.75\sigma_1^s$. All three configurations share factual context/decoder and retain factual residuals in their own coordinates; Learned $T$ is nonlinear calibration. Bottom-row absolute RGB errors share a color scale. This index-prespecified case illustrates an independent frozen diagnostic cohort, not a single-image method ranking.}
\label{fig:full-calibrated-cf}
\end{figure}

\FloatBarrier
\section{Component Ablations and Budget Controls}
\subsection{Component ablations and intervention-budget diagnostics}\label{app:ablation-results}\label{app:ablation-budget-plan}
Ablations on the seven-node domino scene with $64\times64$ RGB images use a frozen backbone and prespecified training seeds 5, 6, and 7 to isolate calibration, descendant propagation, and acquisition strategy.

\subsubsection{What calibration corrects}
With the encoder, decoder, 6000 observations per seed, and 896 target-intervention records fixed, we apply coordinate-wise transformations to the encoded readouts:
\[
 C_0(z)=z,\qquad C_1(z)=1.2z+0.1,\qquad C_2(z)=z+0.08z^3.
\]
Stress conditions were designed post hoc, then formulas, seeds, families, and training schedules were frozen before independent validation, without searching distortion strengths. $C_2$ is strictly monotone but has non-small tails; the finite bounded affine-plus-$\tanh$ family cannot exactly represent its inverse. Each condition compares identity, restricted positive-affine, and monotone nonlinear calibration. All observation/training/validation readouts use the same $C$; commands and labels retain training-anchor units, with supervision only from direct targets.

Each seed adds 448 independent ID target-validation records, 1344 total, excluded from fitting/selection. Nonlinear target-readout error falls by 44.93\%, 88.68\%, and 71.06\% versus identity (Table~\ref{tab:controlled-ablation}A), supporting numerical value/readout alignment, not added structural information. Positive-affine calibration leaves standardized image readouts unchanged up to numerical stabilization, but changes command alignment $q_j(u)$ (Appendix~\ref{app:mechanism-compatibility}). Invertible coordinate-wise maps preserve population conditional independences.

\begin{table}[!htbp]
\centering\small
\setlength{\tabcolsep}{5pt}
\caption{Domino component and budget diagnostics. Errors in A/B are in units of $10^{-3}$, reported as mean $\pm$ sample standard deviation. A reports target-readout MSE and B reports ID counterfactual RGB MSE; values are not compared across panels. Light blue marks nonlinear calibration, and bold marks the lowest error in each row. No standard deviation is reported for a single seed. C gives records used per strategy, including 896 calibration records. $\dagger$ denotes failure to recover after exhausting the budget, not a successful-recovery cost.}
\label{tab:controlled-ablation}
\begin{tabular}{lcrrr}
\toprule
\multicolumn{5}{l}{\textbf{A. Target-calibration error under fixed readout conditions}}\\
Condition & Seeds & Uncalibrated & \shortstack{Restricted positive\\affine $T$} & \cellcolor{OurRowBlue}Nonlinear $T$\\\midrule
No added distortion & \datum{tables-controlled-ablation-tex-c0001} & \datum{tables-controlled-ablation-tex-c0002} & \datum{tables-controlled-ablation-tex-c0003} & \cellcolor{OurRowBlue}$\mathbf{12.852\pm1.690}$\\
Affine distortion & \datum{tables-controlled-ablation-tex-c0004} & \datum{tables-controlled-ablation-tex-c0005} & \datum{tables-controlled-ablation-tex-c0006} & \cellcolor{OurRowBlue}$\mathbf{12.950\pm1.909}$\\
Cubic distortion & \datum{tables-controlled-ablation-tex-c0007} & \datum{tables-controlled-ablation-tex-c0008} & \datum{tables-controlled-ablation-tex-c0009} & \cellcolor{OurRowBlue}$\mathbf{17.781\pm2.835}$\\
\midrule
\multicolumn{5}{l}{\textbf{\shortstack[l]{B. Counterfactual image error within each condition\\Jointly executable seeds}}}\\
Condition & Valid/planned & Uncalibrated & \shortstack{Restricted positive\\affine $T$} & \cellcolor{OurRowBlue}Nonlinear $T$\\\midrule
No added distortion & \datum{tables-controlled-ablation-tex-c0010} & \datum{tables-controlled-ablation-tex-c0011} & \datum{tables-controlled-ablation-tex-c0012} & \cellcolor{OurRowBlue}$2.2683\pm.0320$\\
Affine distortion & \datum{tables-controlled-ablation-tex-c0013} & \datum{tables-controlled-ablation-tex-c0014} & \datum{tables-controlled-ablation-tex-c0015} & \cellcolor{OurRowBlue}$2.2668\pm.0325$\\
Cubic distortion & \datum{tables-controlled-ablation-tex-c0016} & \datum{tables-controlled-ablation-tex-c0017} & \datum{tables-controlled-ablation-tex-c0018} & \cellcolor{OurRowBlue}2.2223\\
\midrule
\multicolumn{5}{l}{\textbf{C. Acquisition strategies with calibration and orientation rules fixed}}\\
Strategy & Seed 5 & Seed 6 & Seed 7 & \shortstack{Complete graph\\recovery}\\\midrule
Coverage targeting & \datum{tables-controlled-ablation-tex-c0019} & \datum{tables-controlled-ablation-tex-c0020} & \datum{tables-controlled-ablation-tex-c0021} & \datum{tables-controlled-ablation-tex-c0022}\\
Random targeting & \datum{tables-controlled-ablation-tex-c0023} & \datum{tables-controlled-ablation-tex-c0024} & \datum{tables-controlled-ablation-tex-c0025} & \datum{tables-controlled-ablation-tex-c0026}\\
Fixed round-robin & \datum{tables-controlled-ablation-tex-c0027} & \datum{tables-controlled-ablation-tex-c0028} & \datum{tables-controlled-ablation-tex-c0029} & \datum{tables-controlled-ablation-tex-c0030}\\
\bottomrule
\end{tabular}
\end{table}

\FloatBarrier

\subsubsection{From readout gains to counterfactual images}
All conditions/$T$ variants share each seed's coverage-trajectory encoding records, then learn their own graphs and mechanisms. Inputs are $C(E(x))$ with shared coordinate-wrapped decoder $D'(z')=D(C^{-1}(z'))$; known $C^{-1}$ does not repair identity-group commands. Factual and CF GT images share simulator exogenous context, not model reconstructions.

Each executable seed uses 64 fixed factual contexts and two ID commands for each of 7 targets at training-anchor mean $\pm0.75$ standard deviations. The first 8 contexts also check implementation without error-based condition selection; the remaining 56 join them for scoring. Only ID queries are tested. Insufficiently oriented units are blocked without substitute predictions or zero errors.

For all three $T$ mappings, seeds 5 and 7 pass under $C_0/C_1$, and only 5 under $C_2$: rates 2/3, 2/3, and 1/3. No direction is wrong; blocked graphs retain one unoriented edge. Of 27 planned units, 15 complete image evaluation and 12 are blocked; only seed 5 is common to all conditions/variants. Panel B therefore supports paired comparisons \emph{within each row's condition}, not across unequal valid denominators.

Nonlinear pixel MSE falls by only 0.71\%, 3.01\%, and 1.98\% versus identity, far below readout gains; positive-affine point estimates are slightly lower throughout. This supports neither general nonlinear superiority nor improved graph recovery. With $C_0$ nonlinear representation, graph, and mechanisms fixed, removing only descendant propagation raises ID CF MSE from $2.2683\pm0.0320$ to $2.9261\pm0.0060$ ($\times10^{-3}$, common seeds 5 and 7), separating propagation from calibration complexity.

\subsubsection{Budget cost and acquisition strategy}
Panel C shares backbone, frozen nonlinear $T$, initial graph, ID command distribution, and graph criteria across three strategies. Each pays $B_{\rm cal}=896$, then starts graph acquisition from an empty intervention pool with 1800 records per batch, at most 6 batches: $B_{\max}=896+10800=11696$. All use \emph{batch} consistency checks, unlike Algorithm 1's sequential commitment.

This three-policy control uses the graph-sampling stream with $N=2600000$ defined in Appendix~\ref{app:query-evidence-r38}. The equal-record control there and the equal-total-budget control in Appendix~\ref{app:ablation-complete-r33} reuse this control's frozen backbone, $T$, and initial graph, but not its graph samples. For seed 5, the first target-1 batch gives reverse-response score 0.10640 on $0$--$1$, within the one-sided margin 0.05 of threshold 0.15. A target-0 batch is therefore needed, giving 4496 total calls. The equal-total-budget control's first-batch score is 0.05252, which already satisfies the one-sided criterion, giving 2696 total calls. This difference reflects the sampled evidence under the specified streams, not omitted calibration cost.

Coverage exhausts the budget with one unoriented edge on seed 6; 11696 is not successful-recovery cost. Round-robin recovers every seed with 2696 records. This shows no coverage-budget advantage, nor does a fixed-numbered seven-node chain establish general round-robin optimality. Terminal accuracy alone hides cost and failure.

Panel C counts \emph{records actually used by each method}, separately from physical new calls reduced by cached target-specific random streams. The strategies acquire 28800 new graph records, without recounting cached replay. Mapping controls hold data fixed; Table~\ref{tab:budget-complete-r33} instead fixes total budget and reallocates the uncalibrated configuration's saved calibration budget to graph acquisition. These controls are summarized separately.

\paragraph{Scope and cost.}
This ID readout-distortion diagnostic uses 1344 independent validation, 1792 CF GT, and 128 factual records. Fitting/graph learning reuse training records without new training calls; thresholds, distortion strengths, and seeds are fixed in advance.

\subsection{Module controls for preservation, replay, propagation, and mechanism fitting}\label{app:module-ablation-r28}
Preservation, replay, orientation propagation, factual residuals, and fitting-record controls use the seven-node domino scene with $64\times64$ RGB images. They share a frozen backbone and planned seeds 5, 6, and 7. Configurations are fixed before evaluation, without OOD selection.

\begin{table}[htbp]
\centering\small\setlength{\tabcolsep}{5pt}
\caption{Module controls with a fixed visual backbone. Values are mean $\pm$ sample SD across training seeds. Bold marks only the lowest point estimate in each error column; light blue marks the full configuration. A uses independent ID validation, while B/C reuse fixed evidence and queries.}
\label{tab:module-ablation-r28}
\begin{tabularx}{\linewidth}{Xrr}
\toprule
\multicolumn{3}{l}{\textbf{\shortstack[l]{A. Calibration objective and replay\\3/3 seeds, all values multiplied by $10^3$}}}\\
Configuration & Target-readout MSE$\downarrow$ & \shortstack{Paired difference\\(ablation $-$ full)}\\
\midrule
\rowcolor{OurRowBlue}
Full configuration & \datum{tables-module-ablation-tex-c0001} & \datum{tables-module-ablation-tex-c0002}\\
Without observation preservation & \datum{tables-module-ablation-tex-c0003} & \datum{tables-module-ablation-tex-c0004}\\
Without cumulative sample replay & \datum{tables-module-ablation-tex-c0005} & \datum{tables-module-ablation-tex-c0006}\\
\midrule
\multicolumn{3}{l}{\textbf{\shortstack[l]{B. Post-intervention orientation propagation\\3/3 seeds, identical fixed evidence}}}\\
Configuration & SHD$\downarrow$ & Complete valid graphs\\
\midrule
\rowcolor{OurRowBlue}
Full configuration & \datum{tables-module-ablation-tex-c0007} & \datum{tables-module-ablation-tex-c0008}\\
Without post-intervention Meek propagation & \datum{tables-module-ablation-tex-c0009} & \datum{tables-module-ablation-tex-c0010}\\
\midrule
\multicolumn{3}{l}{\textbf{\shortstack[l]{C. Residuals and mechanism fitting\\Jointly valid 2/3 seeds, all values $\times10^3$}}}\\
Configuration & CF pixel MSE$\downarrow$ & Non-target encoding NMSE$\downarrow$\\
\midrule
\rowcolor{OurRowBlue}
Full configuration & \datum{tables-module-ablation-tex-c0011} & \datum{tables-module-ablation-tex-c0012}\\
Without descendant factual residuals & \datum{tables-module-ablation-tex-c0013} & \datum{tables-module-ablation-tex-c0014}\\
Non-root mechanisms fitted on all mixed records & \datum{tables-module-ablation-tex-c0015} & \datum{tables-module-ablation-tex-c0016}\\
Non-root mechanisms fitted on observations only & \datum{tables-module-ablation-tex-c0017} & \datum{tables-module-ablation-tex-c0018}\\
\bottomrule
\end{tabularx}
\end{table}

\FloatBarrier

\paragraph{A: Preservation term and cumulative replay.}
With encoder, 896 target records, and 1400 Adam steps fixed, either preservation weight changes from $0.01$ to 0 or a single acquisition batch replaces cumulative evidence while retaining model/optimizer states. After all 9 mappings freeze, shared validation adds 448 independent ID records per seed, 1344 total. Removing replay raises mean target error; removing preservation slightly lowers it but increases training-observation drift. Neither a preservation benefit for target accuracy nor downstream improvement from drift is established.

\paragraph{B: Orientation propagation.}
Both configurations share initial PC graph, calibration, and full coverage evidence prefixes. Disabling only post-intervention Meek additions retains direct evidence and acyclicity/collider/consistent-extension checks. Final SHDs change from $0,1,0$ to $4,5,4$, without incorrect directions; complete valid graphs fall from 2/3 to 0/3. This isolates fixed-evidence propagation, not online budget advantage under resampling.

\paragraph{C: Factual residuals and mechanism fitting.}
With graph, representation, root pool, and decoder fixed, controls separately zero updated descendants' factual residuals, fit non-roots from a mixed pool including self-interventions, or fit observations only. Valid seeds share 64 contexts, 7 targets, and two commands from a fixed ID cache: a retrospective paired control. Seed 6 is insufficiently oriented in all four configurations and blocked without zero imputation; means/SDs use common valid seeds 5 and 7.

Removing factual residuals raises pixel and encoded-readout errors. The mixed pool slightly lowers mean pixel error but raises encoding error, with inconsistent per-seed pixel differences; observation-only fitting stays close to full. Full is therefore not best on every metric. Encoding NMSE is frozen-coordinate error, not independent physical-state error. The mixed pool intervenes only on nodes 0 and 1, so self-target exclusion changes non-root fitting only for node 1. Observation-only fitting also reduces sample size, preventing single-rule attribution.

\paragraph{Cost and interpretation scope.}
No training-environment calls are added; 1344 independent ID validation records are counted separately. B/C reuse evidence, with $B_{\rm cal}=896$ charged per configuration. B's final graph-record counts are $3600,10800,3600$.

\subsection{Component ablations, fitting tradeoffs, and intervention budgets}\label{app:ablation-complete-r33}
These independent paired controls examine affine constraints, nonlinear calibration, and acquisition budgets with separately defined seeds, graph-commitment rules, and evaluation data. They are not pooled as one end-to-end ablation; statistics are descriptive, not significance tests.

\paragraph{Experimental conditions and data access.}
Appendices~\ref{app:query-evidence-r38}, \ref{app:interface-r39}, and \ref{app:module-ablation-r28} retain graph-stage, interface, and replay/Meek/residual controls. Below, frozen-decoder calibration and equal-budget acquisition controls include calibration cost but do not cover every environment/query in the eight-scenario main tables.

\subsubsection{Restricted affine, fully fitted affine, and nonlinear calibration}
Restricted affine gains lie in $[0.75,1.25]$; fully fitted affine retains only $g_j\ge10^{-6}$ and solves coordinate-wise weighted least squares with the same target squared error and $\lambda=0.01$ preservation loss. Positivity bounds are inactive in all seeds. Both affine variants and nonlinear calibration share 896 records; uncalibrated does not fit $T$. Each configuration learns its graph/mechanisms using its own observation normalization, then freezes before independent test generation.

\begin{table}[htbp]
\centering\small\setlength{\tabcolsep}{3pt}
\caption{Calibration-mapping controls with a frozen decoder; errors are multiplied by $10^3$. The upper panel reports mean $\pm$ sample SD and SHD in seed order 5/6/7; the lower panel gives every seed value. State estimation uses 3 seeds, CF the jointly valid 2/3. Seed 6 retains one unoriented edge in every configuration: CF is blocked (--), not zero.}
\label{tab:full-calibration-r33}
\begin{tabularx}{\linewidth}{Xrrc}
\toprule
Configuration & Target MSE & CF image MSE & SHD\\\midrule
Uncalibrated & \datum{appendix-ablation-complete-tex-c0001} & \datum{appendix-ablation-complete-tex-c0002} & \datum{appendix-ablation-complete-tex-c0003}\\
Restricted positive affine & \datum{appendix-ablation-complete-tex-c0004} & \datum{appendix-ablation-complete-tex-c0005} & \datum{appendix-ablation-complete-tex-c0006}\\
Fully fitted positive affine & \datum{appendix-ablation-complete-tex-c0007} & \datum{appendix-ablation-complete-tex-c0008} & \datum{appendix-ablation-complete-tex-c0009}\\
\rowcolor{OurRowBlue}
Nonlinear calibration & \datum{appendix-ablation-complete-tex-c0010} & \datum{appendix-ablation-complete-tex-c0011} & \datum{appendix-ablation-complete-tex-c0012}\\
\bottomrule
\end{tabularx}
\par\medskip
\begin{tabularx}{\linewidth}{Xrrrrrr}
\toprule
& \multicolumn{2}{c}{Seed \datum{appendix-ablation-complete-tex-c0013}} & \multicolumn{2}{c}{Seed \datum{appendix-ablation-complete-tex-c0025}} & \multicolumn{2}{c}{Seed \datum{appendix-ablation-complete-tex-c0033}}\\
\cmidrule(lr){2-3}\cmidrule(lr){4-5}\cmidrule(lr){6-7}
Configuration & \shortstack{Target\\MSE} & \shortstack{CF image\\MSE} & \shortstack{Target\\MSE} & \shortstack{CF image\\MSE} & \shortstack{Target\\MSE} & \shortstack{CF image\\MSE}\\\midrule
Uncalibrated & \datum{appendix-ablation-complete-tex-c0014} & \datum{appendix-ablation-complete-tex-c0015} & \datum{appendix-ablation-complete-tex-c0026} & -- & \datum{appendix-ablation-complete-tex-c0034} & \datum{appendix-ablation-complete-tex-c0035}\\
Restricted affine & \datum{appendix-ablation-complete-tex-c0017} & \datum{appendix-ablation-complete-tex-c0018} & \datum{appendix-ablation-complete-tex-c0028} & -- & \datum{appendix-ablation-complete-tex-c0037} & \datum{appendix-ablation-complete-tex-c0038}\\
Fully fitted affine & \datum{appendix-ablation-complete-tex-c0020} & \datum{appendix-ablation-complete-tex-c0021} & \datum{appendix-ablation-complete-tex-c0030} & -- & \datum{appendix-ablation-complete-tex-c0040} & \datum{appendix-ablation-complete-tex-c0041}\\
\rowcolor{OurRowBlue}
Nonlinear calibration & \datum{appendix-ablation-complete-tex-c0023} & \datum{appendix-ablation-complete-tex-c0024} & \datum{appendix-ablation-complete-tex-c0032} & -- & \datum{appendix-ablation-complete-tex-c0043} & \datum{appendix-ablation-complete-tex-c0044}\\
\bottomrule
\end{tabularx}
\end{table}

Nonlinear mean target error falls by 43.94\% versus no calibration and 13.10\% versus fully fitted affine; these are ratios of means, with improvement in all three seeds, not identical per-seed reductions. Yet all four graphs are identical. Mean CF error is only 0.76\% below uncalibrated and 0.34\% above fully fitted affine, which is lower in both valid seeds. Better state estimates therefore establish neither graph nor image superiority. Similar restricted/fully fitted affine results indicate that the gain interval is not the main constraint here.

This tests direct-target readout correction, not separate image target execution, descendant outcomes, or non-descendant preservation. Affine/nonlinear maps share verified-value supervision; comparison with uncalibrated $T$ does not fully isolate extra supervision from standard updating. Nonlinear changes to local relations/residuals may create tradeoffs with linear mechanisms, but this explanation is not isolated. Cross-source readouts appear in Appendix~\ref{app:query-evidence-r38}; standard updating and image semantics in Appendix~\ref{app:interface-r39}.

Each seed adds 64 independent ID validation records for each of 7 targets, 1344 total. CF uses valid seeds 5 and 7, 64 factual contexts, 7 targets, and two commands per target at $\pm0.75$ training standard deviations: 1792 GT records and 128 factual images. These 3264 test returns are outside training budgets; predictors receive no GT states. Seed 6 stays in the planned denominator without true-graph completion. Appendix~\ref{app:calibration-chain-r30}'s independent ID validation yields 44.35\%/14.54\% reductions and is summarized separately.

\subsubsection{Acquisition efficiency including calibration cost}
Seeds 5, 6, and 7 have budget $B=11696$ per configuration. Three calibrated configurations pay $B_{\rm cal}=896$, leaving 10800 graph records; uncalibrated allocates all 11696 to graph acquisition. Independent random streams pair batches for shared targets; cached reuse still counts against each information budget. Batches contain at most 1800 records with batch consistency checks. Target-value distribution, PC/orientation thresholds are frozen; stopping uses only estimated graph, budget, and available targets, never graph truth.

This equal-total-budget control uses graph-sampling stream $N=12600000$, defined alongside the cross-control frozen-file correspondence in Appendix~\ref{app:query-evidence-r38}. Its nonlinear configurations share training seeds, backbone, $T$, initial graphs, and batch-orientation code with Table~\ref{tab:controlled-ablation}C, but use different graph samples; Appendix~\ref{app:ablation-results} explains the different first-batch decisions for seed 5. Total calls for the nonlinear configurations include 896 calibration records, whereas Table~\ref{tab:r38-acquisition} reports graph records. Comparing the tables requires matching both the sampling stream and the counting convention.

\begin{table}[htbp]
\centering\small\setlength{\tabcolsep}{4pt}
\caption{Equal-total-budget controls. Call counts include calibration. An asterisk denotes exhaustion of the budget without full orientation, not the cost of successful recovery. Seed order is 5/6/7 throughout.}
\label{tab:budget-complete-r33}
\begin{tabularx}{\linewidth}{Xccc}
\toprule
Configuration & \shortstack{Complete true-graph\\recovery} & Actual calls & SHD\\\midrule
Nonlinear calibration + coverage targeting & \datum{appendix-ablation-complete-tex-c0045} & \datum{appendix-ablation-complete-tex-c0046} & \datum{appendix-ablation-complete-tex-c0047}\\
Nonlinear calibration + random targeting & \datum{appendix-ablation-complete-tex-c0048} & \datum{appendix-ablation-complete-tex-c0049} & \datum{appendix-ablation-complete-tex-c0050}\\
Nonlinear calibration + round-robin targeting & \datum{appendix-ablation-complete-tex-c0051} & \datum{appendix-ablation-complete-tex-c0052} & \datum{appendix-ablation-complete-tex-c0053}\\
Uncalibrated + coverage targeting & \datum{appendix-ablation-complete-tex-c0054} & \datum{appendix-ablation-complete-tex-c0055} & \datum{appendix-ablation-complete-tex-c0056}\\
\bottomrule
\end{tabularx}
\end{table}

Coverage has no budget advantage; round-robin recovers all three seeds with 2696 total calls. On seed 6, coverage samples node 1 then repeatedly node 0; final reverse-response score 0.1523 exceeds fixed threshold 0.15, so the two-sided rule retains $0$--$1$. These are zero-based implementation indices. Round-robin samples node 0 first, completing orientation under one-sided evidence. Trajectory, one-/two-sided criteria, and stopping jointly explain this difference, not general inferiority of active design.

Appendix~\ref{app:module-ablation-r28} retains positive and negative preservation, replay, Meek, and fitting results. Divergent target-MSE, observation-drift, and RGB changes do not imply that every module benefits every metric.
\FloatBarrier

\FloatBarrier
\section{Cross-Environment Results and Model Information Access}
\subsection{Configurations and statistics for supplementary comparisons}\label{app:unified-evidence}
Appendices~\ref{app:run-index} and~\ref{app:comparison-scope} give the main tables' reported configurations, run links, and statistical conventions. This section reports the MEC mechanism, representation-model, and rigid-body controls separately. \texttt{--} denotes non-participation in a protocol and exclusion from ranking. Calibration cohorts are scored under their own graph-update rules: frozen-decoder, batch-orientation CF uses common valid seeds 5 and 7; sequential orientation with same-record controls and independent tests completes 3/3 seeds for all four configurations (Appendix~\ref{app:interface-r39}).

\paragraph{Physical-counterfactual ILCM.}
This cell uses seed 42, 1280 training image pairs, and 256 validation image pairs. Four targets, two binary commands, and 32 fixed contexts give 256 queries. Raw floating-point decoded outputs yield pixel MSE $5.707387\times10^{-3}$; clipping to $[0,1]$ for display gives $5.696037\times10^{-3}$, without replacing the prespecified raw metric. Paired GT arrays match archived seed-42 queries for our method and CausalVAE, but the other two main-table methods retain their own summaries; ILCM represents one seed. Its paired-data permissions, label usage, and simulation budget differ, and its decoded command response is weak. Results therefore characterize this adaptation, not ILCM's general capability limit (Appendix~\ref{app:ilcm-physical-demo}).

\paragraph{MEC query groups and training permissions.}
The MEC mechanism-comparison group contains our method with learned $T$ and the no-graph ablation; both share the frozen visual and calibration interface. Structure learning uses 8 seeds per scene, with a development configuration for Balance. Image and OOD queries instead use seeds 300/301: PnP/Gears complete 2/2 and Balance 1/2, with its second seed blocked by an unoriented graph. Means and SDs use executable seeds only, retaining the failure denominator. The MEC representation-model comparison group uses CausalVAE and ILCM at seed 300 in each scene, with RGB64, 2400 training contexts, and 256 validation contexts. CausalVAE uses observational images and concept labels; ILCM trains on paired images and uses training labels for semantic matching and unit calibration during evaluation. Encoder training, resolution, permitted inputs, and normalized readouts are not fully matched across these groups, which are reported separately and not pooled in main-table rankings.

\paragraph{Graph and counterfactual interfaces.}
The MEC representation-model group seals graph scores before accessing the true graph, with discretization threshold fixed at 0.1. CausalVAE clamps target values at the input/output of a general $K$-concept Mask Layer and calls its own decoder, differing from the official four-slot elementwise query. ILCM uses its own conditional implicit mechanisms to convert the set causal-latent coordinate to conditional noise, preserves other noise values, and calls its own decoder; graph extraction passes ID forward/inverse round-trip validation. Each scene uses 32 fixed factual contexts and two ID commands per target: Domino 448 pairs, PnP 384, Balance 320, and Gears 384. GT uses same-exogenous-noise replay; RGB scoring uses the fixed $[0,1]$ scale without per-image normalization.

\paragraph{OOD units.}
The MEC representation-model group normalizes physical readouts by true training-concept standard deviations, using 512 reference contexts and 512 independent simulator contexts. Two ID and two OOD commands per target give $4K$ commands. The mechanism group uses frozen training coordinates and scales, with 2048 outcomes per command. Both exclude the directly set target, but their NMSE readouts and protocols differ; multiplication by $10^3$ is only a display conversion. Test distributions appear in Appendix~\ref{app:ood-definitions-r30}.

\paragraph{ILCM quality diagnostics in the representation-model comparison group.}
PnP ILCM in this group produces nearly identical representations for different ID images, with semantic-readout $R^2$ near 0 and reconstruction worse than the training-mean image. A two-stage frontend control also fails ID quality requirements and therefore receives no OOD evaluation. Actual scores are retained; collapse in this single-seed adaptation does not characterize the main-table method or other scenes.

\paragraph{Rigid-body conditional-mean propagation control.}
The paired rigid-body conditional-mean control's training protocol and checkpoints differ from those of the four-scene MEC representation group. Conditional-mean propagation completes 2/3 seeds and CausalVAE 3/3. Over common seeds 42/314, our result is $0.880\pm0.016$ and CausalVAE's is $1.287\pm0.039$ ($\times10^{-3}$). They share 32 initial states and 256 paired queries, but CausalVAE receives more training concept labels. These statistics belong only to this paired control; they are not per-seed evidence for either the uncalibrated physical aggregate in Table~\ref{tab:run-index} or the main-table learned-calibration result. Appendix~\ref{app:reported-protocol} provides this cohort's full settings and the uncalibrated RGB64/128 visual controls.

\begin{table}[!htbp]
\centering\small
\caption{Disaggregated statistics for the MEC mechanism-comparison group, with errors multiplied by $10^3$. Two-seed results give mean $\pm$ sample standard deviation; parentheses show executable/planned seed counts.}
\begin{tabular}{llrrr}\toprule
Task & Method & PnP (2/2) & Balance (1/2) & Gears (2/2)\\\midrule
CF-image MSE & Our method & \datum{appendix-unified-evidence-tex-c0001} & \datum{appendix-unified-evidence-tex-c0002} & \datum{appendix-unified-evidence-tex-c0003}\\
 & No-graph ablation & \datum{appendix-unified-evidence-tex-c0004} & \datum{appendix-unified-evidence-tex-c0005} & \datum{appendix-unified-evidence-tex-c0006}\\
ID NMSE & Our method & \datum{appendix-unified-evidence-tex-c0007} & \datum{appendix-unified-evidence-tex-c0008} & \datum{appendix-unified-evidence-tex-c0009}\\
 & No-graph ablation & \datum{appendix-unified-evidence-tex-c0010} & \datum{appendix-unified-evidence-tex-c0011} & \datum{appendix-unified-evidence-tex-c0012}\\
OOD NMSE & Our method & \datum{appendix-unified-evidence-tex-c0013} & \datum{appendix-unified-evidence-tex-c0014} & \datum{appendix-unified-evidence-tex-c0015}\\
 & No-graph ablation & \datum{appendix-unified-evidence-tex-c0016} & \datum{appendix-unified-evidence-tex-c0017} & \datum{appendix-unified-evidence-tex-c0018}\\\bottomrule
\end{tabular}
\end{table}

\begin{table}[!htbp]
\centering\small
\caption{Supplementary results for dominoes (RGB images) with our method; errors are multiplied by $10^3$. Neither original-coordinate MSE nor mean-image MSE substitutes for physical-readout NMSE in the main text.}
\begin{tabular}{lrr}\toprule
Metric & Dominoes (64, 8/10) & Dominoes (128, 9/10)\\\midrule
Counterfactual-image MSE & \datum{appendix-unified-evidence-tex-c0019} & \datum{appendix-unified-evidence-tex-c0020}\\
ID original-coordinate MSE & \datum{appendix-unified-evidence-tex-c0021} & \datum{appendix-unified-evidence-tex-c0022}\\
OOD original-coordinate MSE & \datum{appendix-unified-evidence-tex-c0023} & \datum{appendix-unified-evidence-tex-c0024}\\
ID mean-image MSE & \datum{appendix-unified-evidence-tex-c0025} & \datum{appendix-unified-evidence-tex-c0026}\\
OOD mean-image MSE & \datum{appendix-unified-evidence-tex-c0027} & \datum{appendix-unified-evidence-tex-c0028}\\\bottomrule
\end{tabular}
\end{table}

\subsection{Baseline ID diagnostics and model selection}\label{app:baseline-health}
System-error differences must be interpreted together with frontend quality and query responses. This section links evaluation checkpoints in the MEC representation-model comparison group to their ID training records, distinguishing training objectives, state readouts, and image responses. Native validation objectives have different units across models and are not used as cross-model RGB reconstruction comparisons.

\begin{table}[htbp]
\centering\small\setlength{\tabcolsep}{4pt}
\caption{Available ID validation and model-selection records for the MEC representation-model comparison group; each row uses seed 300, $n=1$. $R^2$ is the range across nodes, not a seed-level confidence interval. Selection uses the native ID validation objective over the full mechanism stage. Diagnostics apply only to these checkpoints; coinciding values do not justify extending them to other main-table runs.}
\label{tab:baseline-health}
\begin{tabularx}{\linewidth}{Xlcc}\toprule
Scene & Method & ID node-$R^2$ range & Selected/trained epochs\\\midrule
Domino-MEC & CausalVAE & \datum{tables-baseline-health-tex-c0001} & \datum{tables-baseline-health-tex-c0002}\\
Domino-MEC & ILCM & \datum{tables-baseline-health-tex-c0003} & \datum{tables-baseline-health-tex-c0004}\\
PnP-MEC & CausalVAE & \datum{tables-baseline-health-tex-c0005} & \datum{tables-baseline-health-tex-c0006}\\
PnP-MEC & ILCM & \datum{tables-baseline-health-tex-c0007} & \datum{tables-baseline-health-tex-c0008}\\
Balance-MEC & CausalVAE & \datum{tables-baseline-health-tex-c0009} & \datum{tables-baseline-health-tex-c0010}\\
Balance-MEC & ILCM & \datum{tables-baseline-health-tex-c0011} & \datum{tables-baseline-health-tex-c0012}\\
Gears-MEC & CausalVAE & \datum{tables-baseline-health-tex-c0013} & \datum{tables-baseline-health-tex-c0014}\\
Gears-MEC & ILCM & \datum{tables-baseline-health-tex-c0015} & \datum{tables-baseline-health-tex-c0016}\\
\bottomrule\end{tabularx}
\end{table}

\paragraph{Model selection and executability.}
CausalVAE uses an observation-only training set, while ILCM uses paired images. Semantic matching and unit calibration use only ID training labels. Evaluation checkpoints match the SHA of their corresponding training records and are selected at the epoch with the lowest ID validation objective over the full mechanism stage. All four ILCM minima occur at epoch 120; this identifies the selected models but does not establish adequate optimization convergence. Passing no-op and algebraic target-setting checks in each scene also does not mean that generated images correctly respond to commands.

\paragraph{Available ID reconstruction and command responses.}
PnP ILCM has ID reconstruction MSE 0.007011, worse than the training-mean image's 0.004991, and node-readout $R^2$ near 0. Representation quality in this single-seed adaptation is inadequate; its error cannot be interpreted as an upper limit on ILCM's general capabilities. For the remaining MEC models, the training records do not separately provide same-protocol ID reconstruction MSE and visible command responses. Native validation objectives or algebraic checks do not substitute for these quantities.

Physical ILCM uses the seed-42 checkpoint corresponding to Figure~\ref{fig:rigidbody-causalvae}B, selected at epoch 120 by the full-mechanism-stage ID objective. Across 32 fixed contexts, changing the first domino's command from 0 to 1 yields mean absolute change $2.28\times10^{-5}$ in unclipped decoded outputs, compared with $9.91\times10^{-3}$ in paired physical GT. Executable target setting remains clearly distinct from effective visible response. This physical adaptation has different budgets and supervision permissions; complete settings appear in Appendix~\ref{app:ilcm-physical-demo}.

\paragraph{Scope of the comparison.}
These diagnostics apply only to the listed MEC and physical runs; they do not automatically cover CAUSAL3D, CausalWorld, or other models with the same names. ILCM already uses interventional paired data, so the performance gap cannot be attributed to the baseline lacking interventions. Our interface comparisons examine how known verified intervention values are used for calibration. The effects of data permissions, representation quality, and update scope on system differences are interpreted separately.

\FloatBarrier
\section{Assumptions and Scope of Related Methods}
\subsection{Related-method assumptions and scope}\label{sec:positioning-details}\label{sec:related-details}
Tables~\ref{tab:positioning-assumptions} and~\ref{tab:positioning} expand the settings in Figure~\ref{fig:positioning-venn}: identifying unknown factors, calibrating known variables, and selecting experiments are distinct tasks. The tables describe original inputs, supervision, modules, and graph outputs, not equal-budget performance or proof of priority.

\paragraph{Visual variables and intervention information.}
Causal representation learning targets causally meaningful generative factors rather than reconstruction alone~\citep{scholkopf2021toward}. CausalVAE uses concept supervision~\citep{yang2021causalvae}; CITRIS (Causal Identifiability from Temporal Intervened Sequences) uses temporal images and known targets, and iCITRIS includes instantaneous effects~\citep{lippe2022citris,lippe2023icitris}. BISCUIT (Causal Representation Learning from Binary Interactions) learns variable--interaction correspondence~\citep{lippe2023biscuit}. The implicit latent causal model (ILCM) uses paired observations around unknown random interventions and permits subsequent graph extraction~\citep{brehmer2022weakly}. Identification from interventional-distribution geometry can require neither target indices nor actual values in the corresponding settings~\citep{ahuja2023interventional}. These methods use different graph-learning stages; later extraction is not absence of graph capability. Our known target identities and actual values are stronger supervision, used to calibrate an existing visual correspondence while preserving decoder coordinates.

\paragraph{Control, acquisition, and structure.}
Independently controllable factors (ICF) jointly learns pixel factors and policies using selectivity feedback, without estimating their adjacency graph~\citep{thomas2018disentangling}; Causal Curiosity learns behaviors that distinguish environment factors~\citep{sontakke2021causal}. Action-controllable factorization (ACF) instead estimates policies to correct given transition-data distributions, not to acquire new experiments~\citep{rodriguez2025pixels}. The visual causal discovery network (V-CDN) learns keypoints and structure from given videos~\citep{li2020causalvideos}. For observed variables, active strategies~\citep{hauser2014two}, active intervention targeting (AIT)~\citep{scherrer2021active}, causal Bayesian experimental design (CBED)~\citep{tigas2022interventions}, and causal amortized active structure learning (CAASL)~\citep{annadani2024amortized} choose experiments as detailed in the tables. Our calibration acquisition is fixed round-robin; only subsequent graph discovery selects targets from current ambiguity. Values follow a fixed distribution.

\paragraph{Extrapolation and task-specific structure.}
Rep4Ex uses action--latent invariance for intervention extrapolation~\citep{saengkyongam2024identifying}; action-informed representations therefore have precedent. Causality-Seeking Agent learns context-relevant subgraphs from structured state--action trajectories~\citep{zhao2025curious}, reward-oriented causal representation learning recovers task-relevant variables/structure through adaptive exploration~\citep{yan2025reward}, and Active Causal Experimentalist (ACE) studies mechanism estimation with a known graph~\citep{cooper2026active}. Their specialized outputs should not be interpreted as general capability deficits.

\paragraph{Our scope.}
We calibrate the magnitude of pixel-derived readouts with known semantics, then connect target settings, graph-conditioned consequences, and decoding. Constraint-based and interventional-equivalence-class methods supply graph-learning tools~\citep{colombo2014order,hauser2012characterization}; invertible coordinate-wise calibration adds no identifiability guarantee. Empirical gains require matched-data and matched-budget controls. Image counterfactuals and OOD outcomes have distinct query/distribution definitions; a checkmark denotes neither a theorem nor overall superiority.

\begingroup\footnotesize\setlength{\tabcolsep}{4pt}\renewcommand{\arraystretch}{1.05}
\begin{longtable}{@{}L{3.15cm}L{2.7cm}L{3.55cm}>{\centering\arraybackslash}p{\dimexpr\linewidth-9.4cm-24pt\relax}@{}}
\caption{Assumptions and information available to related methods. A ``concept anchor'' is supervision that associates a visual coordinate with a known physical concept during training. More information does not imply a stronger method; these conditions determine whether a comparison is fair.}
\label{tab:positioning-assumptions}\\
\toprule
Work & Input/samples & Node or variable supervision & Intervention information and key conditions\\
\midrule
\endfirsthead
\caption[]{Assumptions and information available to related methods (continued).}\\
\toprule
Work & Input/samples & Node or variable supervision & Intervention information and key conditions\\
\midrule
\endhead
CausalVAE (2021) & Images & Concept labels & Causal layer and latent-space editing; no active environment acquisition\\
ICF (2018) & Pixel interaction trajectories & Independently controllable factors & Selectivity feedback jointly trains representations and policies\\
Causal Curiosity (2021) & Interaction trajectories & Learned environment factors & Executable experimental behaviors; not standard node-setting supervision\\
CITRIS (2022) & Temporal images & Learned latent variables & Known target labels; actual values not required\\
iCITRIS (2023) & Temporal images & Learned latent variables & Known targets, including instantaneous effects\\
BISCUIT (2023) & Interactive temporal images & Learned variable/interaction correspondence & Binary interaction mechanisms\\
ILCM (2022) & Paired pre/post-intervention observations & Weakly supervised variable learning & Pairing with shared exogenous conditions; unknown targets\\
Ahuja et al. (2023) & Sets of observational/interventional distributions & Representation identification & Targets and actual values need not be known in the corresponding settings\\
ACF (2025, preprint) & Image-transition triplets & Controllable-factor identification & Given data; policy estimation corrects the acquisition distribution\\
V-CDN (2020) & Video images & Keypoint and structure learning & Given data from different configurations; no active experiment selection\\
Rep4Ex (2024) & High-dimensional features, actions, outcomes & Representation identification & Assumptions including linear structure from actions to latent variables\\
\midrule
Hauser--B\"uhlmann (2014) & Observed variables & Given nodes & Experimental design over equivalence classes of candidate graphs\\
AIT (2021) & Observed variables & Given nodes & Given intervention-target space and generator\\
CBED (2022) & Observed variables & Given nodes & Causal-model posterior and executable interventions; active value selection\\
CAASL (2024) & Variables/experiment history & Given nodes & Simulator pretraining with true graphs in the reward; active value selection\\
\midrule
Causality-Seeking Agent (2025) & Structured states--actions & Given state variables & Learned context and its relevant subgraph\\
Reward-oriented CRL (2025) & High-dimensional mixed observations/rewards & Learned task-relevant variables & Linear mixing, causal, and reward assumptions\\
ACE (2026, preprint) & Variables/experiment history & Given graph and nodes & Learned mechanisms and experimental policy; active value selection\\
\midrule
\textbf{This work} & \textbf{RGB images and environment experiments} & \textbf{Known, concept-anchored nodes} & \textbf{Known hard-intervention targets and setting values; frozen after calibration}\\
\bottomrule
\end{longtable}
\endgroup
\par\smallskip
\begin{minipage}{\linewidth}\footnotesize
\textbf{Interpretation:} Using intervention data does not imply identical supervision. Our assumptions of known nodes and setting values are stronger than those of several weakly supervised representation methods, but do not require their particular temporal or factual--interventional pairing formats. High-dimensional inputs in Rep4Ex and reward-oriented methods are not equated here with image experiments; context-dependent subgraphs are also distinct from our static post-settlement graph. Sources for all rows appear in Appendix~\ref{sec:related-details}.
\end{minipage}
\begin{table}[!htbp]
\centering
\caption{Capabilities and representation constraints of related methods. $\checkmark$: explicitly included in the original method; $\triangle$: a different form or a subsequent module; ---: not a component of the method, without ruling out extensions. The last column states what is constrained, rather than treating different tasks as a single type of calibration.}
\label{tab:positioning}
\begingroup\small\setlength{\tabcolsep}{3pt}\renewcommand{\arraystretch}{1.12}
\begin{tabularx}{\linewidth}{@{}L{3.05cm}>{\centering\arraybackslash}p{.85cm}>{\centering\arraybackslash}p{.85cm}L{3.0cm}Y@{}}
\toprule
Work & \shortstack{Repr.\\learn.} & \shortstack{Active\\exp.} & Graph output/task & Representation constraints\\
\midrule
CausalVAE & $\checkmark$ & --- & Static causal graph & Concept supervision and causal layer\\
ICF & $\triangle$ & $\checkmark$ & No explicit graph-recovery task & Independent controllability and intervention selectivity\\
Causal Curiosity & $\triangle$ & $\checkmark$ & No general explicit graph recovery & Experiments distinguish environment factors\\
CITRIS & $\checkmark$ & --- & $\triangle$ Subsequent temporal graph & Intervention-target constraints\\
iCITRIS & $\checkmark$ & --- & Instantaneous and temporal graphs & Intervention-target constraints\\
BISCUIT & $\checkmark$ & --- & $\triangle$ Subsequent temporal graph & Binary interaction-mechanism correspondence\\
ILCM & $\checkmark$ & --- & $\triangle$ Post-training graph extraction & Pre/post-intervention pairing\\
Ahuja et al. & $\checkmark$ & --- & Not the main output & Geometry/constancy of interventional distributions\\
ACF & $\checkmark$ & --- & No explicit graph-recovery task & Action effects relative to natural transitions\\
V-CDN & $\checkmark$ & --- & Structure among keypoints & Structure and dynamics in given videos\\
Rep4Ex & $\checkmark$ & --- & Not a latent-graph discovery task & Action--representation invariance\\
\midrule
Hauser--B\"uhlmann & --- & $\checkmark$ & Graph equivalence classes/orientation & ---\\
AIT & --- & $\checkmark$ & Interfaces with graph-learning backends & ---\\
CBED & --- & $\checkmark$ & Graph and mechanism posterior & ---\\
CAASL & --- & $\checkmark$ & Interfaces with graph-learning backends & ---\\
\midrule
Causality-Seeking Agent & $\triangle$ & $\checkmark$ & Context-dependent subgraph & Latent context state\\
Reward-oriented CRL & $\checkmark$ & $\checkmark$ & Task-relevant coarse structure & Reward relevance\\
ACE & --- & $\checkmark$ & Mechanism estimation for a known graph & ---\\
\midrule
\textbf{This work} & $\triangle$ & $\checkmark$ & \textbf{Static graph estimates/unoriented edges} & \textbf{Invertible coordinate-wise calibration to actual setting values}\\
\bottomrule
\end{tabularx}
\endgroup
\par\smallskip
\begin{minipage}{\linewidth}\footnotesize
\textbf{Scope:} Active value selection means selecting or optimizing the numerical hard-intervention value of a known node using feedback; it cannot be inferred merely from a continuous-action policy. Our method actively selects only targets, sampling values from a prespecified distribution without optimizing them. Our $\triangle$ denotes coordinate learning and readout calibration with existing concept identities, not disentanglement of unknown factors. Active experiments occur only during graph discovery; calibration acquisition uses fixed round-robin. Factor learning in ICF and Causal Curiosity is not a general latent-structure identification guarantee, and the latent context learned by the Causality-Seeking Agent is not node recovery from pixels. Subsequent graph extraction does not imply an absence of graph-discovery capability. Checkmarks imply neither theoretical guarantees nor empirical superiority under equal budgets.
\end{minipage}
\end{table}


\begin{thebibliography}{26}
\providecommand{\natexlab}[1]{#1}
\providecommand{\url}[1]{\texttt{#1}}
\expandafter\ifx\csname urlstyle\endcsname\relax
  \providecommand{\doi}[1]{doi: #1}\else
  \providecommand{\doi}{doi: \begingroup \urlstyle{rm}\Url}\fi

\bibitem[Ahmed et~al.(2021)Ahmed, Tr{\"a}uble, Goyal, Neitz, Bengio, Sch{\"o}lkopf, W{\"u}thrich, and Bauer]{ahmed2021causalworld}
Ossama Ahmed, Frederik Tr{\"a}uble, Anirudh Goyal, Alexander Neitz, Yoshua Bengio, Bernhard Sch{\"o}lkopf, Manuel W{\"u}thrich, and Stefan Bauer.
\newblock {CausalWorld}: A robotic manipulation benchmark for causal structure and transfer learning.
\newblock In \emph{International Conference on Learning Representations}, 2021.
\newblock URL \url{https://openreview.net/forum?id=SK7A5pdrgov}.

\bibitem[Ahuja et~al.(2023)Ahuja, Mahajan, Wang, and Bengio]{ahuja2023interventional}
Kartik Ahuja, Divyat Mahajan, Yixin Wang, and Yoshua Bengio.
\newblock Interventional causal representation learning.
\newblock In \emph{Proceedings of the 40th International Conference on Machine Learning}, volume 202, pp.\  372--407. PMLR, 2023.
\newblock URL \url{https://proceedings.mlr.press/v202/ahuja23a.html}.

\bibitem[Annadani et~al.(2024)Annadani, Tigas, Bauer, and Foster]{annadani2024amortized}
Yashas Annadani, Panagiotis Tigas, Stefan Bauer, and Adam Foster.
\newblock Amortized active causal induction with deep reinforcement learning.
\newblock In \emph{Advances in Neural Information Processing Systems}, volume~37, 2024.
\newblock \doi{10.52202/079017-1403}.
\newblock URL \url{https://papers.neurips.cc/paper_files/paper/2024/hash/4e2c6423d724370b36c3a7612f25b78c-Abstract-Conference.html}.

\bibitem[Brehmer et~al.(2022)Brehmer, de~Haan, Lippe, and Cohen]{brehmer2022weakly}
Johann Brehmer, Pim de~Haan, Phillip Lippe, and Taco~S. Cohen.
\newblock Weakly supervised causal representation learning.
\newblock In \emph{Advances in Neural Information Processing Systems}, volume~35, 2022.
\newblock URL \url{https://papers.nips.cc/paper_files/paper/2022/hash/fa567e2b2c870f8f09a87b6e73370869-Abstract-Conference.html}.

\bibitem[Colombo \& Maathuis(2014)Colombo and Maathuis]{colombo2014order}
Diego Colombo and Marloes~H. Maathuis.
\newblock Order-independent constraint-based causal structure learning.
\newblock \emph{Journal of Machine Learning Research}, 15\penalty0 (116):\penalty0 3921--3962, 2014.
\newblock URL \url{https://www.jmlr.org/papers/v15/colombo14a.html}.

\bibitem[Cooper \& Velasquez(2026)Cooper and Velasquez]{cooper2026active}
Patrick Cooper and Alvaro Velasquez.
\newblock Active causal experimentalist ({ACE}): Learning intervention strategies via direct preference optimization.
\newblock arXiv:2602.02451, 2026.
\newblock URL \url{https://arxiv.org/abs/2602.02451v2}.
\newblock Version 2, June 2026; preprint.

\bibitem[Ding et~al.(2026)Ding, Lai, Chen, Zhang, and Chen]{ding2026causalvae}
Ziyi Ding, Xianxin Lai, Weiyu Chen, Xiao-Ping Zhang, and Jiayu Chen.
\newblock {CausalVAE} as a plug-in for world models: Towards reliable counterfactual dynamics.
\newblock In \emph{Computer Vision -- ECCV 2026}, volume 17017 of \emph{Lecture Notes in Computer Science}, pp.\  278--293. Springer, 2026.
\newblock \doi{10.1007/978-3-032-37252-9_16}.
\newblock URL \url{https://link.springer.com/chapter/10.1007/978-3-032-37252-9_16}.

\bibitem[Hauser \& B{\"u}hlmann(2012)Hauser and B{\"u}hlmann]{hauser2012characterization}
Alain Hauser and Peter B{\"u}hlmann.
\newblock Characterization and greedy learning of interventional {Markov} equivalence classes of directed acyclic graphs.
\newblock \emph{Journal of Machine Learning Research}, 13:\penalty0 2409--2464, 2012.
\newblock URL \url{https://www.jmlr.org/papers/v13/hauser12a.html}.

\bibitem[Hauser \& B{\"u}hlmann(2014)Hauser and B{\"u}hlmann]{hauser2014two}
Alain Hauser and Peter B{\"u}hlmann.
\newblock Two optimal strategies for active learning of causal models from interventional data.
\newblock \emph{International Journal of Approximate Reasoning}, 55\penalty0 (4):\penalty0 926--939, 2014.

\bibitem[Komanduri et~al.(2024)Komanduri, Wu, Chen, and Wu]{komanduri2024icmvae}
Aneesh Komanduri, Yongkai Wu, Feng Chen, and Xintao Wu.
\newblock Learning causally disentangled representations via the principle of independent causal mechanisms.
\newblock In \emph{Proceedings of the Thirty-Third International Joint Conference on Artificial Intelligence}, pp.\  4308--4316, 2024.
\newblock \doi{10.24963/ijcai.2024/476}.
\newblock URL \url{https://www.ijcai.org/proceedings/2024/476}.

\bibitem[Li et~al.(2020)Li, Torralba, Anandkumar, Fox, and Garg]{li2020causalvideos}
Yunzhu Li, Antonio Torralba, Anima Anandkumar, Dieter Fox, and Animesh Garg.
\newblock Causal discovery in physical systems from videos.
\newblock In \emph{Advances in Neural Information Processing Systems}, volume~33, 2020.
\newblock URL \url{https://proceedings.neurips.cc/paper_files/paper/2020/hash/6822951732be44edf818dc5a97d32ca6-Abstract.html}.

\bibitem[Lippe et~al.(2022)Lippe, Magliacane, L{\"o}we, Asano, Cohen, and Gavves]{lippe2022citris}
Phillip Lippe, Sara Magliacane, Sindy L{\"o}we, Yuki~M Asano, Taco Cohen, and Stratis Gavves.
\newblock {CITRIS}: Causal identifiability from temporal intervened sequences.
\newblock In \emph{Proceedings of the 39th International Conference on Machine Learning}, volume 162, pp.\  13557--13603. PMLR, 2022.
\newblock URL \url{https://proceedings.mlr.press/v162/lippe22a.html}.

\bibitem[Lippe et~al.(2023{\natexlab{a}})Lippe, Magliacane, L{\"o}we, Asano, Cohen, and Gavves]{lippe2023biscuit}
Phillip Lippe, Sara Magliacane, Sindy L{\"o}we, Yuki~M Asano, Taco Cohen, and Efstratios Gavves.
\newblock {BISCUIT}: Causal representation learning from binary interactions.
\newblock In \emph{Proceedings of the Thirty-Ninth Conference on Uncertainty in Artificial Intelligence}, volume 216, pp.\  1263--1273. PMLR, 2023{\natexlab{a}}.
\newblock URL \url{https://proceedings.mlr.press/v216/lippe23a.html}.

\bibitem[Lippe et~al.(2023{\natexlab{b}})Lippe, Magliacane, L{\"o}we, Asano, Cohen, and Gavves]{lippe2023icitris}
Phillip Lippe, Sara Magliacane, Sindy L{\"o}we, Yuki~M. Asano, Taco Cohen, and Efstratios Gavves.
\newblock Causal representation learning for instantaneous and temporal effects in interactive systems.
\newblock In \emph{International Conference on Learning Representations}, 2023{\natexlab{b}}.
\newblock URL \url{https://arxiv.org/abs/2206.06169}.

\bibitem[Liu et~al.(2025)Liu, Qiao, Liu, Lu, Zhou, Liang, Yin, and Ma]{liu2025causal3d}
Disheng Liu, Yiran Qiao, Wuche Liu, Yiren Lu, Yunlai Zhou, Tuo Liang, Yu~Yin, and Jing Ma.
\newblock {CAUSAL3D}: A comprehensive benchmark for causal learning from visual data.
\newblock \emph{arXiv preprint arXiv:2503.04852}, 2025.
\newblock URL \url{https://arxiv.org/abs/2503.04852}.

\bibitem[Meek(1995)]{meek1995}
Christopher Meek.
\newblock Causal inference and causal explanation with background knowledge.
\newblock In \emph{Proceedings of the Eleventh Conference on Uncertainty in Artificial Intelligence}, pp.\  403--410, 1995.
\newblock URL \url{https://arxiv.org/abs/1302.4972}.

\bibitem[Rodriguez-Sanchez et~al.(2025)Rodriguez-Sanchez, Allen, and Konidaris]{rodriguez2025pixels}
Rafael Rodriguez-Sanchez, Cameron Allen, and George Konidaris.
\newblock From pixels to factors: Learning independently controllable state variables for reinforcement learning.
\newblock arXiv:2510.02484, 2025.
\newblock URL \url{https://arxiv.org/abs/2510.02484}.
\newblock Preprint.

\bibitem[Saengkyongam et~al.(2024)Saengkyongam, Rosenfeld, Ravikumar, Pfister, and Peters]{saengkyongam2024identifying}
Sorawit Saengkyongam, Elan Rosenfeld, Pradeep Ravikumar, Niklas Pfister, and Jonas Peters.
\newblock Identifying representations for intervention extrapolation.
\newblock In \emph{International Conference on Learning Representations}, 2024.
\newblock URL \url{https://proceedings.iclr.cc/paper_files/paper/2024/hash/aa9508b2577e0d9c824da9e2e27fe9d6-Abstract-Conference.html}.

\bibitem[Scherrer et~al.(2021)Scherrer, Bilaniuk, Annadani, Goyal, Schwab, Sch{\"o}lkopf, Mozer, Bengio, Bauer, and Ke]{scherrer2021active}
Nino Scherrer, Olexa Bilaniuk, Yashas Annadani, Anirudh Goyal, Patrick Schwab, Bernhard Sch{\"o}lkopf, Michael~C. Mozer, Yoshua Bengio, Stefan Bauer, and Nan~Rosemary Ke.
\newblock Learning neural causal models with active interventions.
\newblock arXiv:2109.02429, 2021.
\newblock URL \url{https://arxiv.org/abs/2109.02429}.
\newblock Revised March 2022.

\bibitem[Sch{\"o}lkopf et~al.(2021)Sch{\"o}lkopf, Locatello, Bauer, Ke, Kalchbrenner, Goyal, and Bengio]{scholkopf2021toward}
Bernhard Sch{\"o}lkopf, Francesco Locatello, Stefan Bauer, Nan~Rosemary Ke, Nal Kalchbrenner, Anirudh Goyal, and Yoshua Bengio.
\newblock Toward causal representation learning.
\newblock \emph{Proceedings of the IEEE}, 109\penalty0 (5):\penalty0 612--634, 2021.
\newblock URL \url{https://arxiv.org/abs/2102.11107}.

\bibitem[Sontakke et~al.(2021)Sontakke, Mehrjou, Itti, and Sch{\"o}lkopf]{sontakke2021causal}
Sumedh~A Sontakke, Arash Mehrjou, Laurent Itti, and Bernhard Sch{\"o}lkopf.
\newblock Causal curiosity: {RL} agents discovering self-supervised experiments for causal representation learning.
\newblock In \emph{Proceedings of the 38th International Conference on Machine Learning}, volume 139, pp.\  9848--9858. PMLR, 2021.
\newblock URL \url{https://proceedings.mlr.press/v139/sontakke21a.html}.

\bibitem[Thomas et~al.(2018)Thomas, Bengio, Fedus, Pondard, Beaudoin, Larochelle, Pineau, Precup, and Bengio]{thomas2018disentangling}
Valentin Thomas, Emmanuel Bengio, William Fedus, Jules Pondard, Philippe Beaudoin, Hugo Larochelle, Joelle Pineau, Doina Precup, and Yoshua Bengio.
\newblock Disentangling the independently controllable factors of variation by interacting with the world.
\newblock arXiv:1802.09484, 2018.
\newblock URL \url{https://arxiv.org/abs/1802.09484}.

\bibitem[Tigas et~al.(2022)Tigas, Annadani, Jesson, Sch{\"o}lkopf, Gal, and Bauer]{tigas2022interventions}
Panagiotis Tigas, Yashas Annadani, Andrew Jesson, Bernhard Sch{\"o}lkopf, Yarin Gal, and Stefan Bauer.
\newblock Interventions, where and how? experimental design for causal models at scale.
\newblock In \emph{Advances in Neural Information Processing Systems}, volume~35, 2022.
\newblock URL \url{https://arxiv.org/abs/2203.02016}.

\bibitem[Yan et~al.(2025)Yan, Acart{\"u}rk, and Tajer]{yan2025reward}
Zirui Yan, Emre Acart{\"u}rk, and Ali Tajer.
\newblock Reward-oriented causal representation learning.
\newblock In \emph{Advances in Neural Information Processing Systems}, volume~38, 2025.
\newblock \doi{10.52202/085713-1381}.
\newblock URL \url{https://papers.neurips.cc/paper_files/paper/2025/hash/3b04bba0f95caacbd9aa9033e2c913c9-Abstract-Conference.html}.

\bibitem[Yang et~al.(2021)Yang, Liu, Chen, Shen, Hao, and Wang]{yang2021causalvae}
Mengyue Yang, Furui Liu, Zhitang Chen, Xinwei Shen, Jianye Hao, and Jun Wang.
\newblock {CausalVAE}: Disentangled representation learning via neural structural causal models.
\newblock In \emph{IEEE/CVF Conference on Computer Vision and Pattern Recognition}, pp.\  9593--9602, 2021.

\bibitem[Zhao et~al.(2025)Zhao, Li, Zhang, Wang, Faccio, Schmidhuber, and Yang]{zhao2025curious}
Zhiyu Zhao, Haoxuan Li, Haifeng Zhang, Jun Wang, Francesco Faccio, J{\"u}rgen Schmidhuber, and Mengyue Yang.
\newblock Curious causality-seeking agents in open-ended worlds.
\newblock In \emph{Advances in Neural Information Processing Systems}, volume~38, 2025.
\newblock \doi{10.52202/085713-5145}.
\newblock URL \url{https://papers.neurips.cc/paper_files/paper/2025/hash/e21c92a1661709c3caf21fdd958bac99-Abstract-Conference.html}.

\end{thebibliography}
\end{document}